%% file: main_final.tex
\documentclass{amia}
\usepackage{graphicx}
\usepackage[labelfont=bf]{caption}
\usepackage[superscript,nomove]{cite}
\usepackage{color}
\usepackage{longtable}
\usepackage{multirow}
\usepackage{amsmath,amssymb}
\usepackage{mathtools}
\usepackage{subfigure}
\usepackage{subfloat}
\usepackage{caption}
\usepackage{todonotes}
\usepackage{lscape}
\usepackage{url}
\usepackage{comment}
\usepackage{booktabs}
\usepackage{setspace}
\usepackage{bibspacing}
\titleformat*{\section}{\LARGE\bfseries}
\titleformat*{\subsection}{\Large\bfseries}
\titleformat*{\subsubsection}{\large\bfseries}
\renewcommand*{\thefootnote}{\fnsymbol{footnote}}

\newcommand\blfootnote[1]{%
  \begingroup
  \renewcommand\thefootnote{}\footnote{#1}%
  \addtocounter{footnote}{-1}%
  \endgroup
}

\begin{document}

\title{Generating Electronic Health Records with Multiple Data Types and Constraints}

\author{Chao Yan*, MS$^{1}$, Ziqi Zhang*, BS$^{1}$, Steve Nyemba, MS$^{2}$, Bradley A. Malin, PhD$^{1,2}$}
\blfootnote{* Equal contribution}

\institutes{
    $^1$Vanderbilt University, Nashville, TN; $^2$Vanderbilt University Medical Center, Nashville, TN\\
}

\maketitle

\noindent{\bf Abstract}
\textit{Sharing electronic health records (EHRs) on a large scale may lead to privacy intrusions. Recent research has shown that risks may be mitigated by simulating EHRs through generative adversarial network (GAN) frameworks. Yet the methods developed to date are limited because they 1) focus on generating data of a single type (e.g., diagnosis codes), neglecting other data types (e.g., demographics, procedures or vital signs) and 2) do not represent constraints between features. In this paper, we introduce a method to simulate EHRs composed of multiple data types by 1) refining the GAN model, 2) accounting for feature constraints, and 3) incorporating key utility measures for such generation tasks. Our analysis with over $770,000$ EHRs from Vanderbilt University Medical Center demonstrates that the new model achieves higher performance in terms of retaining basic statistics, cross-feature correlations, latent structural properties, feature constraints and associated patterns from real data, without sacrificing privacy.}


\captionsetup[table]{skip=1pt}
\input{intro_final.tex}

\input{relatedwork_final.tex}

\input{methods_final.tex}

\input{result_final.tex}

\input{conclusion_final.tex}


\makeatletter
\renewcommand{\@biblabel}[1]{\hfill #1.}
\makeatother
\bibliographystyle{vancouver}

\begin{spacing}{0.1}
\begin{small}
\bibliography{references}
\end{small}
\end{spacing}

\end{document}

%% file: intro_final.tex
\section*{Introduction}


Electronic health records (EHRs) have been widely adopted by healthcare organizations (HCO) to provide more accurate, timely and safer care for patients. Though initially designed to support healthcare, EHRs have shown great promise in secondary scenarios, including facilitating software systems development, enabling training in clinical settings, and boosting biomedical research. \cite{jensen2012mining,casey2016using} As a result, HCOs are incentivized to share EHR data, but  are concerned that doing so could lead to patient privacy intrusions and loss in trust. \cite{filkins2016privacy,mcguire2008confidentiality} Various computational approaches have been proposed to maintain patient anonymity and confidentiality \cite{fung2010privacy}, but providing raw data, at any degree of specificity, leads to a tradeoff between privacy and data utility. \cite{dwork2013toward} Moreover, as the amount of amendments made to EHR data grows, one realizes greater privacy protection at the cost of lower utility.


Recent advances in machine learning may make it possible to alleviate such tensions by enabling the generation of synthetic EHR data with the look and feel of real data. In particular, deep learning approaches based on generative adversarial networks (GANs)  \cite{goodfellow2014generative,goodfellow2016deep} have demonstrated an uncanny ability for simulating realistic-looking data instances with high statistical generalizability, scalability and limited reliance upon knowledge drawn from domain experts. \cite{zhang2020ensuring,choi2017generating,baowaly2019synthesizing} This is because GAN-based models are trained in an adversarial environment, where a generator produces increasingly realistic instances, such that an evolving discriminator cannot distinguish them from real data.

However, there are several gaps between current GAN-based EHR simulation approaches and the practical needs of simulation and evaluation. First, current models are designed to simulate one type of data, such as International Classification of Diseases (ICD) codes. Yet EHRs are an amalgamation of a multitude of types, such as demographics (e.g., age, gender, and race), procedures, medications, laboratory test results, and vital signs. Beyond differences in semantics, the data has different syntax (being composed of binary, categorical and continuous values), which induces higher dimensionality and greater sparsity. Second, there are various record-level constraints that exist between features. For example, the results of a blood pressure test should indicate that systolic pressure is greater than diastolic pressure. However, such constraints are not directly embedded in a generative model, leading to situations where the corresponding semantics are violated in the generated dataset, which weakens the overall utility of synthetic EHRs.

Moreover, the current array of data utility measures for synthetic EHRs are deficient in several respects. First, they do not assess record-wise consistency, which is the extent to which frequently associated conditions or events in real EHRs are sufficiently maintained by the generative model in record level (e.g., correlated diseases and common diagnosis-procedure pairings). This is a concern because poor record-wise consistency can invalidate record-level studies in the synthetic data. Second, to simulate EHRs with multiple data types, we need to measure if the conditional distribution of one data type is well-represented with respect to another. For instance, the correlation between the distribution of blood pressure with respect to a certain diagnosis (e.g., hypertension) should be similar in the real and synthetic data.

In this paper, we address the aforementioned challenges by refining GAN-based generative models and enriching the current set of data utility measures. Specifically, we 1) incorporate a penalization into the GAN learning process such that, if a violation transpires during training, then the generator will be penalized and, thus, forced to output records with more desirable feature constraints; 2) refine the deep neural networks of both the generator and discriminator to make signal sparser and, thus, more efficient, 3) introduce measures for constraint violations, feature associations, and conditional distributions to assess EHR simulation models. We then demonstrate the effectiveness of the new model (as well as the privacy risks) and measures, using a dataset based on over $770,000$ EHRs from Vanderbilt University Medical Center (VUMC). 


%% file: relatedwork_final.tex
\section*{Related Work}

In this section, we review recent GAN-based developments in EHR simulation. Choi et al. first customized a GAN framework to generate a set of structured and discrete features, in the form of ICD-9 codes. \cite{choi2017generating} Recognizing that the original GAN framework \cite{goodfellow2014generative} could not generate discrete values, an autoencoder was incorporated to learn the distribution of discrete data and a trained decoder was applied to generate data in a discrete space. A limitation of this model was that it could suffer from a mode collapse (that is, the generator maps different inputs to the same output) and mode drop (that is, the generator only captures certain regions of the underlying distribution of the real data). To mitigate these problems,  Baowaly et  al. introduced an approach that used Wasserstein divergence  to more effectively measure difference in real and synthetic data distribution. \cite{baowaly2019synthesizing} Zhang et al. enhanced the billing code simulation model by removing the auto-encoder and incorporating utility measures to evaluate structural properties of the data. \cite{zhang2020ensuring} Still, all of these techniques focused on generating only a single feature type. Recently, Chin-Cheong et al. explored the use of off-the-shelf models \cite{arjovsky2017wasserstein,gulrajani2017improved} to generate EHRs with more than one data type. \cite{chin2019generation} However, their work did not improve the learning model, nor did it address feature constraints. They also did not evaluate data with respect to consistency between features. By contrast, in this paper, we focus on model refinement and evaluation in more challenging scenarios than previous simulation settings.

%% file: methods_final.tex
\section*{Data Overview}

The data in this study was derived from the VUMC Synthetic Derivative (SD), a de-identified warehouse of over $2.2$ million EHRs. We collected the EHRs of patients with at least one recorded visit during a ten-year period. For each EHR, we extracted several types of data: 1) age (at the latest time in the resource), 2) gender, 3) ICD-9 codes, 4) Current Procedural Terminology-Fourth Version (CPT-4) codes, 5) body mass index (BMI), and 5) systolic and diastolic blood pressures. We restricted our analysis to EHRs with at least one documented BMI and blood pressure reading. We note that we use BMI and blood pressure due to the fact that they are frequently populated in EHRs and are common covariates in biomedical research. We rolled up 1) ICD codes to their subcategories (by removing the portion of the codes to the right of the ``.'') and 2) CPT codes to the minor categories (including $115$ distinct categories) \cite{medprice2017monkey}. As such, through the remainder of this paper, ICD and CPT codes refer to their rolled-up versions. We refer to this dataset, which includes $928,089$ records, as the SD.

To mitigate noise in the data, we further refined the set of records. First, we ranked all ICD codes based on their prevalence and removed those with a rate less than 1/1000 (which corresponded to $928$ patients). The same process was applied to CPT codes. Second, we removed records that were composed of less than $5$ distinct ICD and CPT codes. Third, we removed obviously-errored test results, which specifically corresponded to cases where 1) systolic pressure was less than diastolic pressure in the same observation, 2) systolic pressure was greater than $300$, and 3) BMI was less than $5$. We refer to the refined dataset as CSD (C stands for clean). It contains $770,231$ records, $693$ distinct ICD codes and $65$ CPT codes. Summary statistics for both datasets are provided in Table \ref{data}.

\begin{table}[ht]
       \fontsize{6.5}{6}\selectfont
	\centering
	\caption{Summary statistics of the EHR datasets.}\label{data}
	\begin{tabular}{ccccccccccc}
		\toprule
		\textbf{Dataset}  &  \textbf{Patients} & \textbf{Gender}   & \textbf{ICD} & \textbf{ICD Codes} & \textbf{Patients Per} & \textbf{CPT}   & \textbf{CPT Codes}  & \textbf{Patients Per} & \textbf{BMI Instances} & \textbf{Blood Pressure}\\[0.1em]
		  &   &   & \textbf{Codes}    & \textbf{Per Patient} & \textbf{ICD Codes} & \textbf{Codes}  & \textbf{Per Patient} & \textbf{CPT Codes}  & \textbf{Per Patient} & \textbf{Instances Per Patient} \\[0.3em] \hline
		  \rule{0pt}{1.0\normalbaselineskip}
		 \textbf{SD}  & $928,089$ & M:43\% F:57\% & $926$ & $12.03$ & $15,208$ & $88$ & $7.02$ & $104,044$ & $10.32$ & $33.60$ \\[0.3em]
		 \textbf{CSD} & $770, 231$ & M:44\%: F:56\%  &  $693$ & $13.88$ & $20,221$ & $65$ & $8.00$ & $140,815$ & $12.05$ & $40.05$ \\[0.1em]
		\bottomrule  
	\end{tabular}
	\vspace{-2mm}
\end{table}

For reference, Table \ref{format} illustrates the format of a record, where the second row indicates the length of the sub-vector needed to represent each feature space for the CSD dataset. As can be seen, the CSD contains multiple types of features and two syntactic types, which are binary and continuous. To incorporate a statistical summary of the vital signs of each record, we computed the $minimum$, $median$ and $maximum$ values of BMI, systolic, and diastolic pressure, respectively. Each patient's record can then be represented as a vector over age, gender, ICD codes, CPT codes, minimum BMI (systolic and diastolic), median BMI (systolic and diastolic) and maximum BMI (systolic and diastolic). Note that categorical features can be embedded by applying a one-hot encoding strategy, which leads to a binary representation.

\begin{table}[h]
       \fontsize{6.5}{6}\selectfont
	\centering
	\caption{Data representation of the CSD dataset. (\emph{B}: binary; \emph{C}: continuous).}\label{format}
	\begin{tabular}{@{}cccccccccccccc@{}}
	   \toprule
	 \multirow{2}{*}{\textbf{Data Type}}  &  \multirow{2}{*}{\textbf{Age}}  & \multirow{2}{*}{\textbf{Gender}} & \multirow{2}{*}{\textbf{ICD Codes}} & \multirow{2}{*}{\textbf{CPT Codes}}  & \multicolumn{3}{c}{\textbf{BMI}}   & \multicolumn{3}{c}{\textbf{Systolic}} & \multicolumn{3}{c}{\textbf{Diastolic}} \\[0.0em] \cmidrule(l){6-8} \cmidrule(l){9-11} \cmidrule(l){12-14}  
		  &   &   &  &  & \textbf{Min} & \textbf{Median} & \textbf{Max} & \textbf{Min} & \textbf{Median} & \textbf{Max} & \textbf{Min} & \textbf{Median} & \textbf{Max}  \\[0.3em] 
		  \hline
		  \rule{0pt}{1.0\normalbaselineskip}
		 \textbf{Variable Type} & B & B & B & B & C &C & C & C & C & C & C & C & C   \\[0.3em]
		 \textbf{Length} & $100$ & $1$ & $693$ & $65$ & $1$ & $1$ & $1$ & $1$ & $1$ & $1$ & $1$ & $1$ & $1$   \\[0.1em]
		\bottomrule  
	\end{tabular}
	\vspace{-2mm}
\end{table}

\textbf{Constraints.} In addition to the heterogeneity in data types, there are multiple constraints that can exist between features in practical data synthesis tasks. Consider the CSD dataset. In each record, the minimum value of a continuous feature (i.e., BMI, systolic and diastolic pressure) should be no greater than its corresponding median value. Similarly, the median value of a continuous feature should be no greater than its maximum value. The generative model needs to capture these constraints so that it can simulate synthetic EHRs without violating them.

 
\section*{Method}

In this section, we introduce the GAN-based generative model for the simulation of EHR data with heterogenous data types and feature constraints. We then define new utility measures to assess the quality of synthetic data.


\emph{\textbf{Generative Model}}

\emph{\textbf{Preliminary of GAN in EHR simulation.}} There are two core components in GANs for EHR data simulation: 1) a generator neural network  $G(\mathbf{z}, \mathbf{\theta}_g)$ which accepts random noise $\mathbf{z} \in \mathbb{R}^r$ and 2) a discriminator neural network $D(\mathbf{x, \mathbf{\theta}}_d)$ which accepts EHR data represented as vector $\mathbf{x}$. These two components are updated iteratively and alternatingly through competition with one another. 
The generator $G(\mathbf{z}, \mathbf{\theta}_g)$ is trained to minimize the statistical divergence between the distribution of real data $\mathbb{P}_r$ and the distribution of generated data $\mathbb{P}_g$; i.e., 
$\min_{G}Div(\mathbb{P}_r, \mathbb{P}_g)$. Ideally, $G$ is able to learn $\mathbb{P}_r$ and generate synthetic records that are indistinguishable to the discriminator $D$. By contrast, the discriminator $D$ is trained to distinguish synthetic data generated by $G$ from real data. In EHR simulation, the state-of-the-art models \cite{zhang2020ensuring} adopt Wasserstein Divergence \cite{arjovsky2017wasserstein} to characterize the earth mover distance between two distributions. When combined with a gradient penalty technique \cite{gulrajani2017improved}, such a divergence measure can effectively mitigate the problem of poor convergence, which is the main cause of mode collapses and mode drops.

\begin{figure}[h]
	\centering
	\includegraphics[scale=0.35]{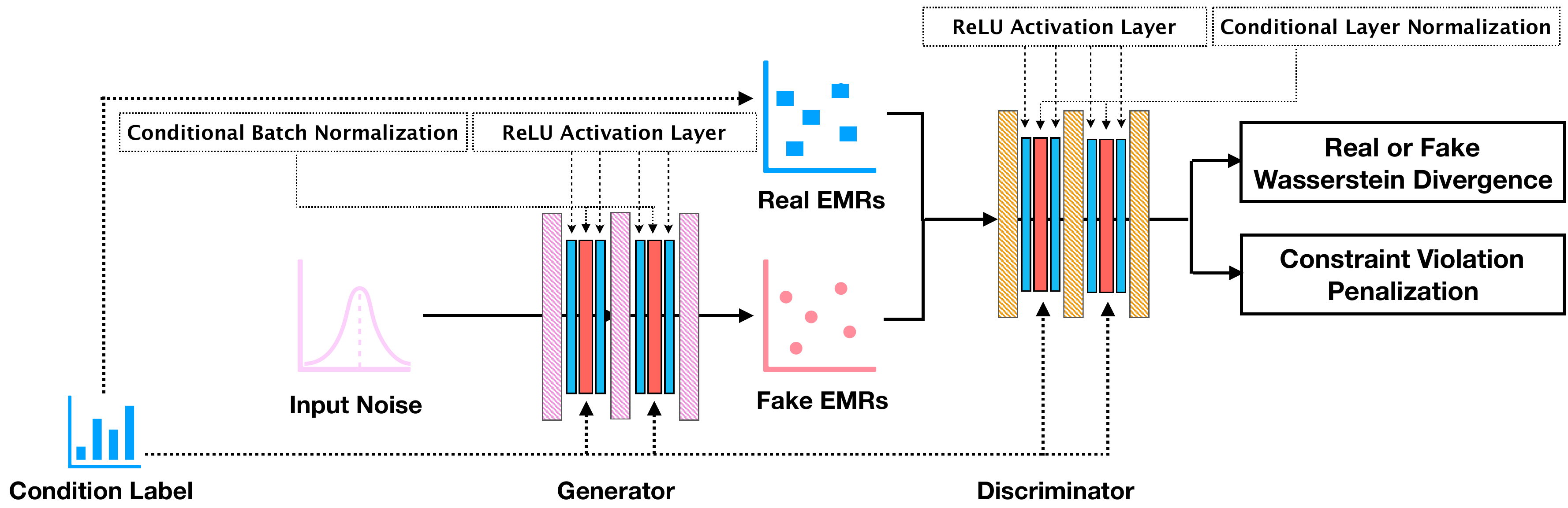} \vspace*{-10mm}
	\caption{Architecture of HGAN.}
	\label{framework}
\end{figure}

\emph{\textbf{Heterogeneous GAN.}} We build our generative model, namely \emph{Heterogeneous GAN} (HGAN), based on the basic structure of the state-of-the-art model, \emph{EMR-CWGAN}.   
The architecture of HGAN is shown in Figure \ref{framework}. 
First, we apply a conditional generation framework, where the conditional batch normalization \cite{dumoulin2016learned,de2017modulating} (in $G$) and conditional layer normalization (in $D$) are leveraged to control the specific generation and discrimination. In this work, we use age and gender as the conditioning features to simulate EHR data. Specifically, we build one distributed representation vector (i.e., embedding) for each integer age and one embedding vector for each gender. We train HGAN using a batch of records from CSD in each training step and then use their conditioning features to extract the associated embedding representations to build the normalization layers. 

Second, we reorder the neural networks for both the generator and discriminator. Instead of applying \emph{conditional normalization} $\rightarrow$ \emph{rectified linear unit (ReLU) layer} between fully connected layers (as adopted by EMR-CWGAN \cite{zhang2020ensuring}), we use \emph{ReLU}  $\rightarrow$ \emph{conditional normalization} $\rightarrow$ \emph{ReLU} to filter signals. Though the former implementation has been widely utilized in the deep learning community \cite{ioffe2015batch}, we anticipated that our design can make data sparser, which helps disentangle the signals and, thus, make the representation robust and efficient. \cite{glorot2011deep} We confirm this expectation in the experimental results.



Third, to encourage simulated EHRs to adhere to the record-level feature constraints, we incorporate a penalization term as part of the loss function of the GAN model. Formally, the \emph{minmax} objective with gradient penalty and constraint violation penalty becomes:
\begin{equation}
\begin{aligned}
 \min_G \max_D  \underbrace{\mathbb{E}_{\mathbf{x}\sim \mathbb{P}_r}[D(\mathbf{x})] - \mathbb{E}_{\mathbf{\widetilde{x}}\sim \mathbb{P}_g}[D(\mathbf{\widetilde{x}})]}_{\text{Original Discriminator Loss}}  - &\underbrace{\lambda \mathbb{E}_{\mathbf{\widehat{x}} \sim \mathbb{P}_{\mathbf{\widehat{x}}}, \mathbf{\delta} \sim N_d(\mathbf{0},aI))} [(||\bigtriangledown_{\mathbf{\widehat{x}}} D(\mathbf{\widehat{x}}+\mathbf{\delta})||_2 - \mathbf{1} )^2]}_{\text{Gradient Penalty}} \\
&+ \underbrace{\beta \sum_{c\in \cal{C}} \mathbb{E}_{\mathbf{\widetilde{x}}\sim \mathbb{P}_g} [P(\widetilde{\mathbf{x}}, {\cal{F}}_c(\cdot))]}_{\text{Constraint Violation Penalty}},
\end{aligned}
\end{equation}
where the first term denotes the classification loss of $D$ and the second term penalizes the situation where the gradient of $D$ is far from $1$. In the third term, $c$ denotes a feature constraint from set $\cal{C}$ and $P(\widetilde{\mathbf{x}}, {\cal{F}}_c(\cdot))$ represents the quantity of penalty applied to the synthetic record $\widetilde{\mathbf{x}}$ when using the constraint-specific penalization function ${\cal{F}}_c(\cdot)$.
The coefficient $\lambda$ and $\beta$ control the weights of the two penalty terms during optimization. Recall that in CSD there is one type of constraint within multiple continuous feature pairs due to their ordinal relationship in semantics (e.g., for each type of vital signs, the \emph{min} value is no greater than the \emph{median} value, and the \emph{median} value is no greater than the \emph{max} value). We define the constraint violation penalty function for such a constraint as:
\begin{equation}
{\cal{F}}_c(x) = \max\{f_1(x)-f_2(x),0\},
\end{equation}
where $f_1(\cdot)$ and $f_2(\cdot)$ denote the feature extractor of EHR data, which correspond to \emph{min} and \emph{median} values for the former case and \emph{median} and \emph{max} values for the latter case, respectively.
Another constraint is the mutual exclusiveness between two binary features.  In this case, the constraint violation penalty function can be defined with the format ${\cal{F}}_c(x) = \omega [f_1(x) \cdot f_2(x)]$. Consider an example that a synthetic EHR of a man cannot own the CPT code ``$59400$ vaginal delivery". In this case, $f_1(x)$ and $f_2(x)$ represent the gender ($1$ for male) and the CPT code $59400$ ($1$ for presence), respectively. As long as both values are close to $1$ in the training process, the generator will be heavily penalized. The existence of feature constraints depends on the feature space for EHR simulation tasks and ${\cal{F}}(\cdot)$ can be flexibly defined based on specific constraints.

\emph{\textbf{Utility Measures}}

Various utility measures for EHR simulation have been proposed \cite{choi2017generating,zhang2020ensuring}, including dimension-wise statistics (DWS), dimension-wise prediction (DWP), latent space representation (LSR), and first-order proximity (FOP). DWS investigates the degree to which the distribution of each binary feature (e.g., ICD or CPT code) in the synthetic EHR records is similar to real data. By contrast, DWP evaluates the degree to which a generative model captures the cross-dimensional relationships of real data. LSR and FOP measure the ability of a generative model to maintain the structural properties of real data in the latent and original space, respectively. 
Due to space limitations, we refer the reader to the supplemental materials of a recent paper \cite{zhang2020ensuring} for the implementation details of these measures. In this section, we introduce three new utility measures to provide greater intuition into the quality of simulated EHR data in the general simulation scenarios.


\textbf{Constraint Violation Test (CVT).} Given that there may be known constraints between features in EHR data, we need to determine the extent to which the synthetic data satisfy these constraints. Though there may be a large number of constraints among the features in an EHR dataset, in this paper we focus on solving the ordinal relationship between continuous features. To assess if the minimum value is no greater than the median value, we compute the difference between the values in median and minimum columns of each vital sign (including BMI, systolic and diastolic pressure) in each synthetic record, and then investigate whether this difference is positive. Similarly, we perform this evaluation for the max versus median case for each vital sign. As a baseline, for each setting we provide the corresponding distribution of difference obtained from the real data. An ideal generative model can simulate EHR data that obeys the record-level feature constraints and has a similar distribution with the real data.

\textbf{Frequent Association Rules (FAR).} This utility measure investigates the extent to which the record-level medical condition associations in real EHRs maintained in the synthetic ones. This measure functions over a set of categorical features, such as a mixture of ICD and CPT codes. 
The two key criteria in association rule mining are \emph{support} and \emph{confidence}. Support is an indication of how frequently the condition set appears in the dataset, whereas confidence is an indication of how often a condition rule is true.  With respect to rule mining in EHR, the support of condition set $X$ (e.g., a set of diseases and procedures) with respect to record dataset $T$ is defined as the proportion of records in $T$ that contain $X$.  
The confidence of a condition rule, $X\Rightarrow Y$, with respect to $T$, is the proportion of  records that contain $X$ that also contain $Y$.  
We first obtain all  frequent condition sets, FCS for abbreviation, (forming a set  $\cal{S}$) with frequency larger than a threshold $min_s$ such that any subset of any FCS is not in $\cal{S}$.  For each FCS $f \in \cal{S}$, we then determine the set of association rules $R: f' \Rightarrow f - f'$, where each rule satisfies that the number of records which have $f'$ also have $f$ is greater than a threshold $min_c$.  
By applying such a process to both real and synthetic EHR data, we measure the proportion of the association rules that are from the synthetic data that are in the real
data  and vice versa, which we refer to as recall and precision, respectively. We use the well-known association rule mining technique--\emph{Apriori} \cite{han2011data} to learn FCSs and the association rules from the real and synthetic EHRs. 
It is notable that FAR can be regarded as an expansion on the structural measure FOP. This is because FAR does not limit the number of features to consider and, thus, consider deeper and broader dependencies between features. By contrast, FOP focuses on the condition sets containing only two features. As a consequence, in this paper we report the FAR results, instead of FOP.


\textbf{Cross-type Conditional Distribution (CCD).} This utility measure evaluates the ability of a generative model to maintain the distribution of one data type conditioned on another. Since the correlation between ICD and CPT codes (binary one-hot representation) can be extracted and evaluated by FAR, we focus on the correlation between continuous and binary features. Specifically, we investigate the distribution of vital sign features conditioned on ICD and CPT codes. We compare the mean and standard deviation of each conditional distribution for real and synthetic EHR data. Similarly, we investigate the distributions of the conditioning labels of GAN model (i.e., age and gender in this study) on each ICD and CPT codes. This indicates the degree to which the demographic distributions for each health condition and procedure are similar in the real and synthetic data.


\emph{\textbf{Privacy Measures}}

We investigate the extent to which the synthetic data is susceptible to \emph{membership} and \emph{attribute inference} attacks. \cite{choi2017generating}
For the membership inference, the attacker is assumed to have the entire record of a set of real patients and attempts to infer which patients are in the training dataset of the generative model. This is achieved by calculating the Hamming distance between each compromised and synthetic record. We then apply a threshold such that one compromised record has a distance (to any synthetic record) less than the threshold is considered in the training dataset. 
For the attribute inference, an attacker is assumed to possess a subset of features of certain real EHRs and aims to infer the value of a missing feature.  $k$-nearest neighbors algorithm is adopted to determine the missing values.
As these two approaches were designed for the situation where all features are represented in a binary form, we uniformly discretize all continuous features into categories and then use one-hot distributed representations.

%% file: result_final.tex
\section*{Results}

In this section, we compare the utility of synthetic EHR data generated by our model and several alternative models. The first alternative is the state-of-the-art model, \emph{EMR-CWGAN}, which generates one data type, and, thus, will be relied upon to investigate whether generating a mixture of data types decreases the data utility on the binary features (i.e., ICD and CPT codes). To investigate how the new filter \emph{ReLU}  $\rightarrow$ \emph{conditional normalization} $\rightarrow$ \emph{ReLU} between fully connected layers influences data generation, we set the second alternative model by applying the filter in EMR-CWGAN to HGAN, which is \emph{conditional normalization} $\rightarrow$ \emph{ReLU}. We refer to this model as \emph{HGAN-U}. It should be noted that, for evaluation purposes, we first train HGAN and HGAN-U and use them to generate heterogeneous records with the feature space depicted in Table \ref{format}. We then extract the features related to specific data utility measure to evaluate the performance.
In addition, we build a baseline by randomly partitioning the real dataset into two equal sized datasets to construct a real \emph{vs} real setting. This allows us to derive an upper bound on what an ideal generative model can achieve with respect to each utility measure. Note that, we evaluate ICD and CPT as one set of features as both are one-hot encoded.


\emph{\textbf{Experimental Setup}} 

To compare EHR data simulation methods, we fixed hyperparameter settings for all models in experiments. Specifically, we structured the generator and discriminator with a network formation of  ($128$, $256$, $256$, $512$, $512$, $512$, $512$, $767$) and ($767$, $512$, $384$, $256$, $256$, $128$, $128$, $1$), respectively. To construct the conditional normalization layers, we set the length of the embedding vectors for age and gender to $96$ and $32$, respectively. All generative models were trained over $1,000$ epochs. We applied the Adam optimizer \cite{kingma2014adam} with a learning rate of $4 \cdot 10^{-6}$ and $2 \cdot 10^{-5}$ for the generator and discriminator, respectively. 

\begin{figure}[ht]%
\centering
\subfigure[Real]{%
\label{dws_1}%
\includegraphics[width=2.3cm]{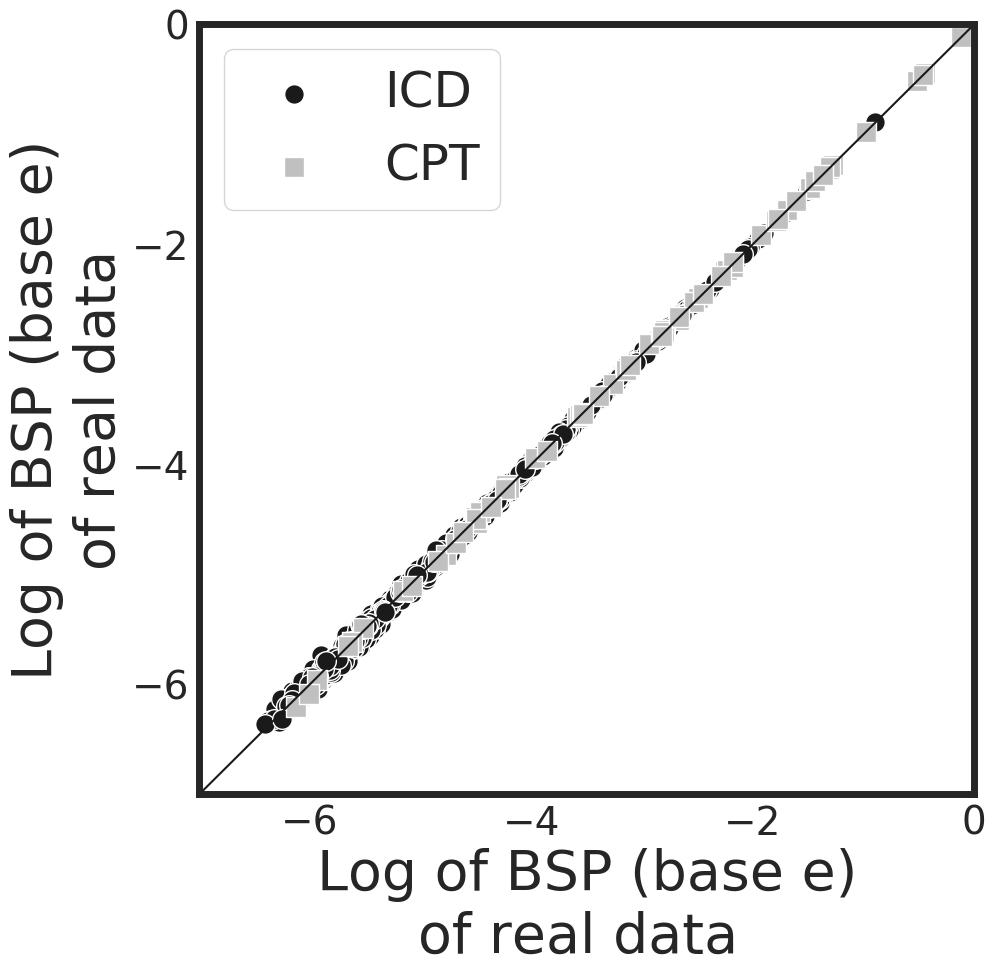}} \hspace{7mm}
\subfigure[EMR-CWGAN]{%
\label{dws_2}%
\includegraphics[width=2.3cm]{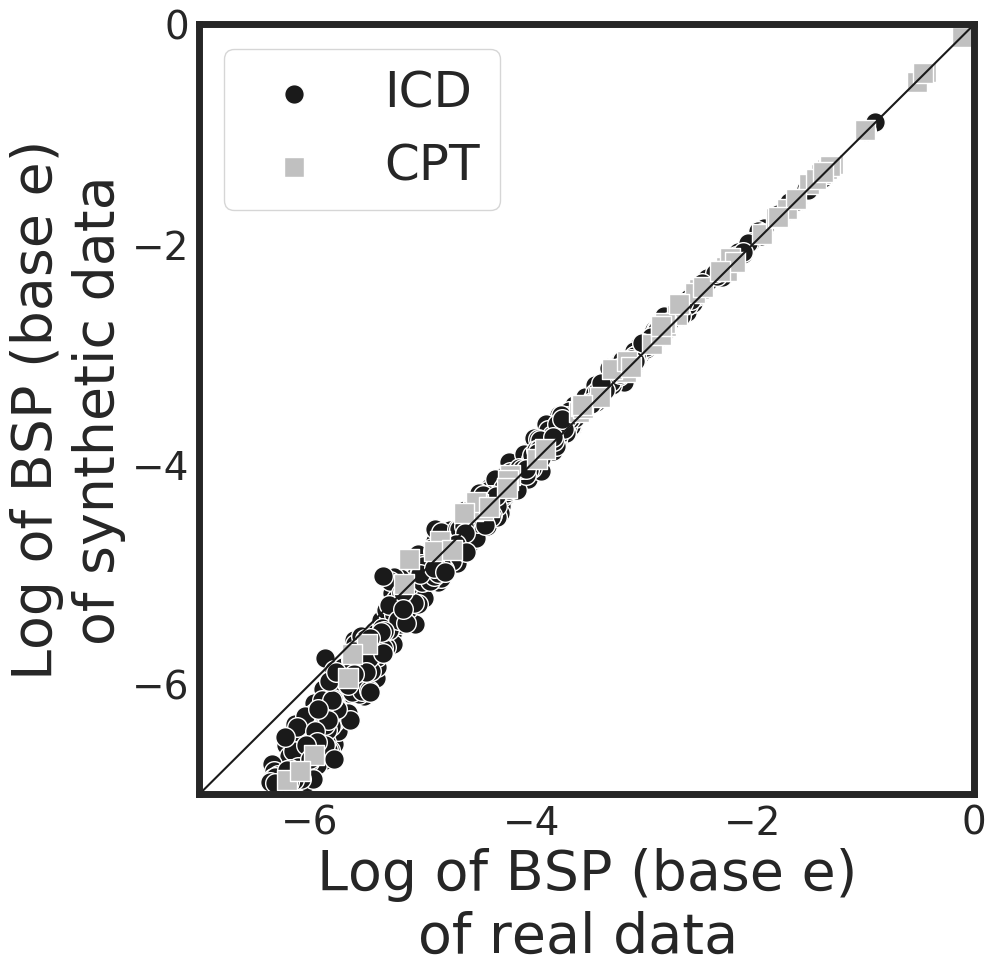}}\hspace{7mm}
\subfigure[HGAN-U]{%
\label{dws_3}%
\includegraphics[width=2.3cm]{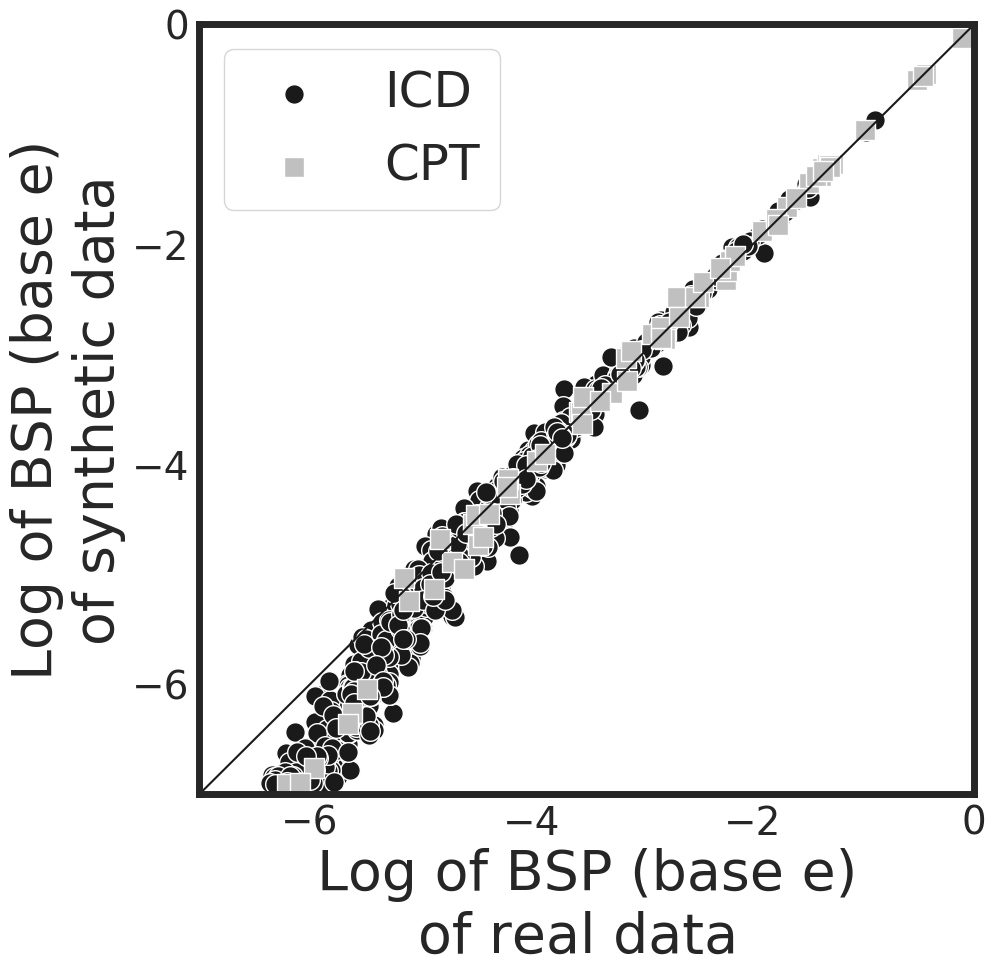}}\hspace{7mm}
\subfigure[HGAN]{%
\label{dws_4}%
\includegraphics[width=2.3cm]{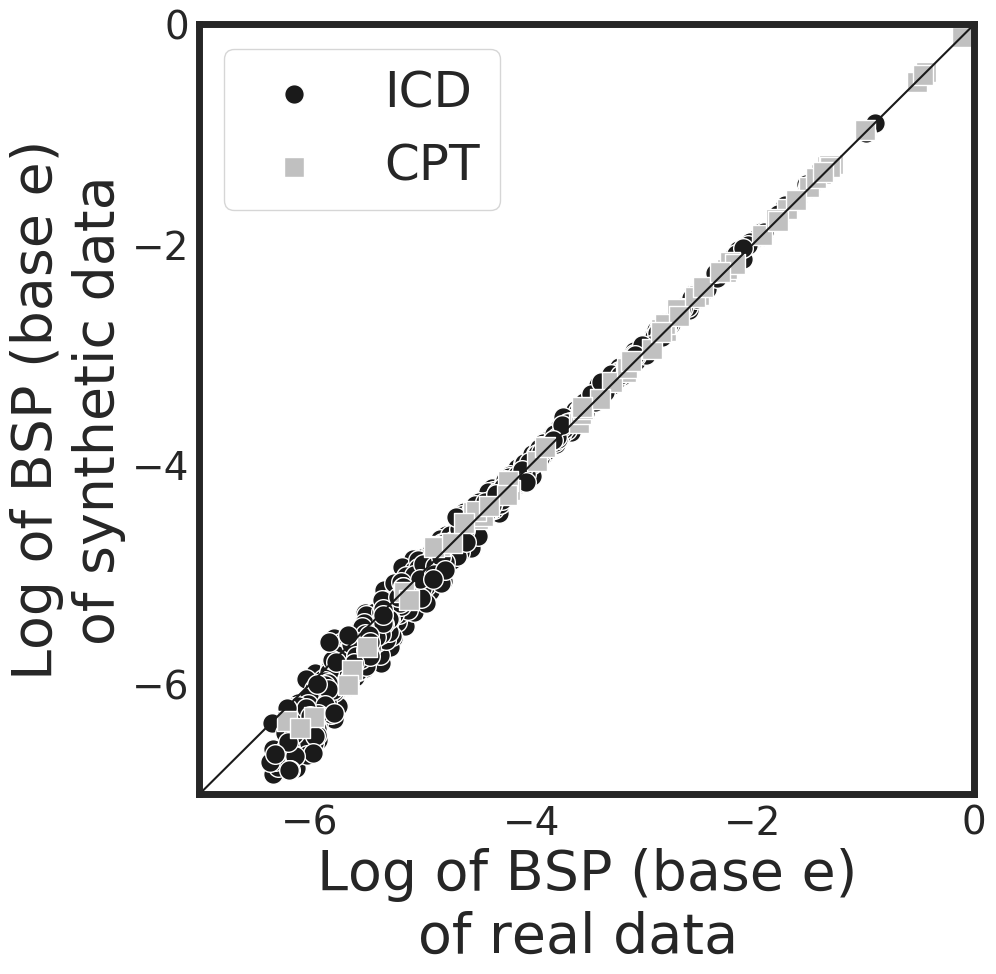}}%
\vspace*{-10mm}
\caption{Dimension-wise statistics. Bernoulli success probabilities (BSP) in logarithmic scale for ICD and CPT codes.}\label{dws}
\end{figure}

\textbf{Dimension-wise statistics (DWS).}  
The DWS results are shown in Figures \ref{dws_1}-\ref{dws_4}, where Figure \ref{dws_1} describes the real \emph{vs} real setting. In all other subfigures, the $x$-axis corresponds to the Bernoulli success probability (or incidence rate) of one (ICD or CPT) code in the CSD dataset and the $y$-axis corresponds to this probability in the synthetic data. The $45$ degree diagonal line corresponds to what a perfect correlation would indicate. There are several findings to highlight. First, the incidence rates of codes in the real \emph{vs} real setting are closely distributed along the diagonal line, which indicates highest stability of basic statistics. 
Second, by comparing Figures \ref{dws_2} and \ref{dws_4}, it can be seen that HGAN achieves very similar performance to EMR-CWGAN for codes with high prevalence ($x>-4$) and induces less bias with respect to codes with low prevalence ($x<-5$).
Third, by comparing Figures \ref{dws_3} and \ref{dws_4}, similar observations can be made in that HGAN-U exhibits a biased and less stable result for low prevalence codes ($x<-5$) than HGAN. This evidence suggests that HGAN provides a more faithful representation of the basic statistics in real EHR data.
Fourth, it can be seen from all Figure \ref{dws} subfigures that the difference in the distributions between the two code types (i.e., ICD and CPTs) is negligible, which implies that the learning is unbiased between them.

\begin{figure}[ht]%
\centering
\subfigure[Real]{%
\label{dwp_1}%
\includegraphics[width=2.3cm]{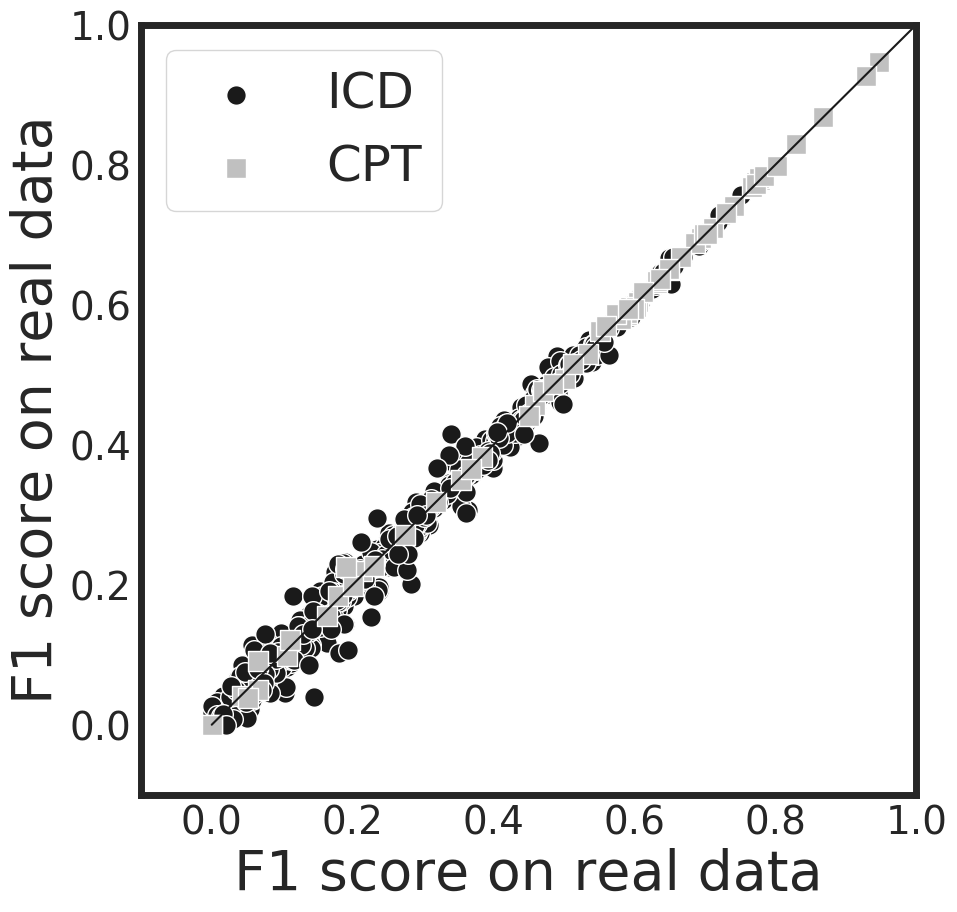}}\hspace{7mm}
\subfigure[EMR-CWGAN]{%
\label{dwp_2}%
\includegraphics[width=2.3cm]{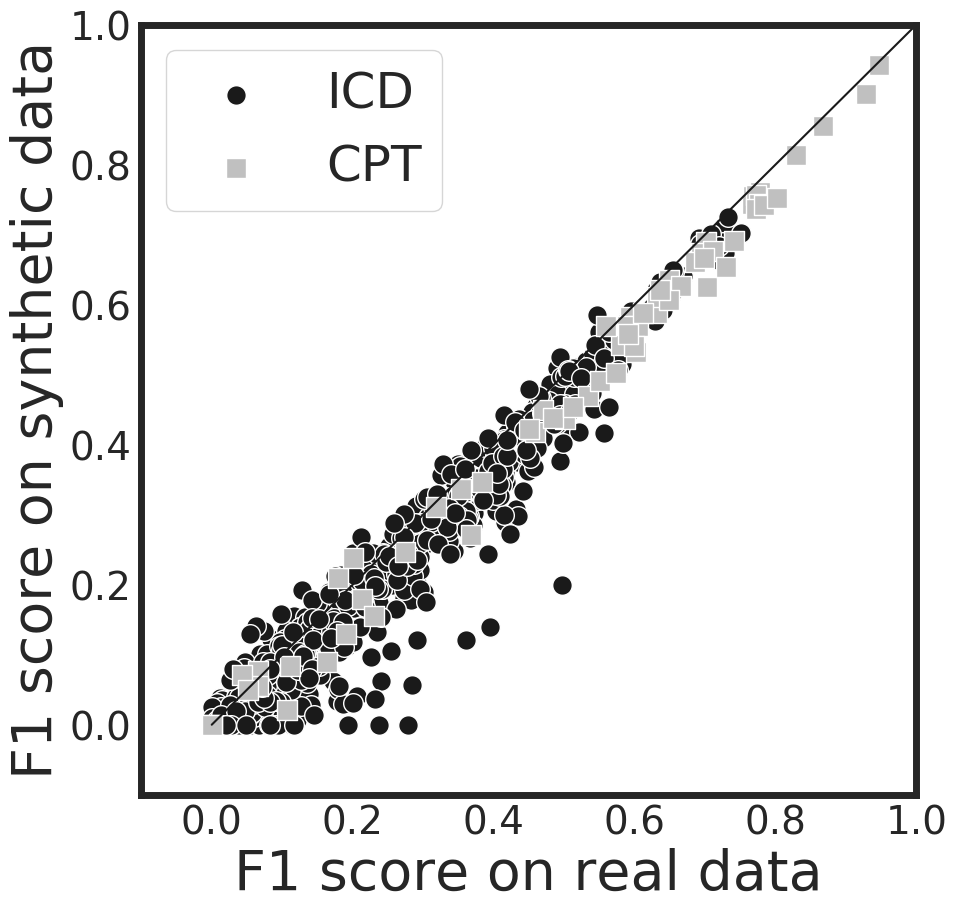}}\hspace{7mm}
\subfigure[HGAN-U]{%
\label{dwp_3}%
\includegraphics[width=2.3cm]{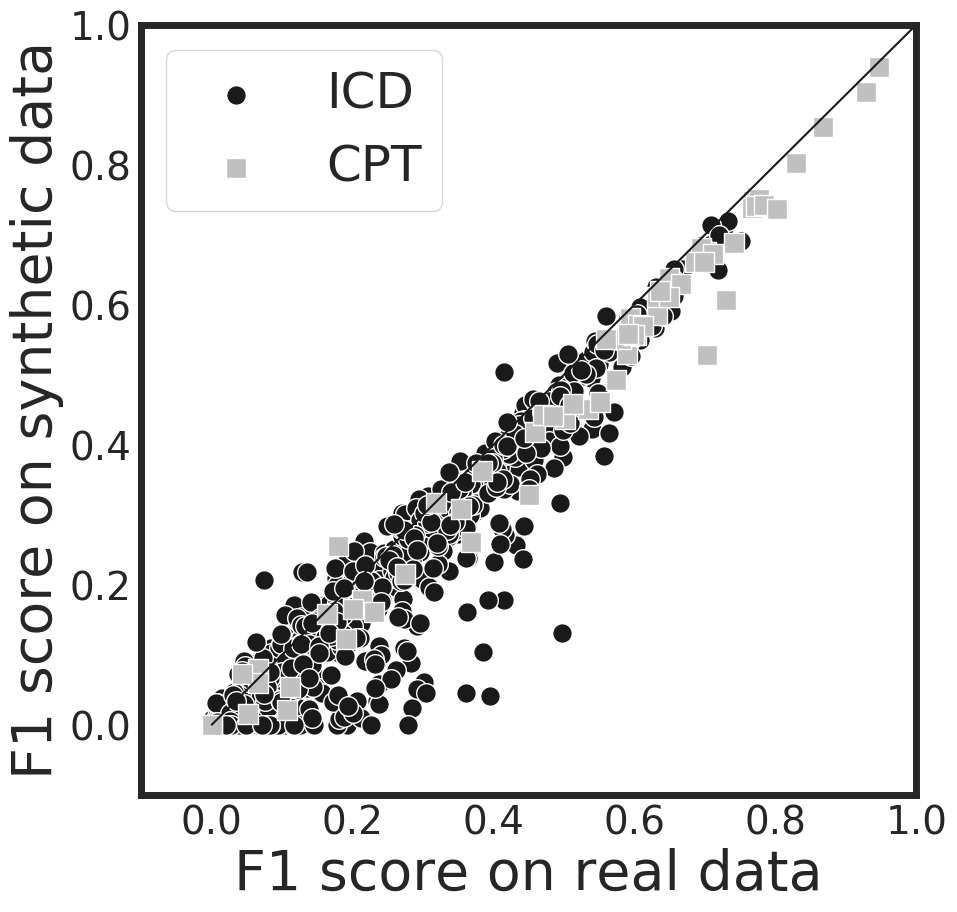}}\hspace{7mm}
\subfigure[HGAN]{%
\label{dwp_4}%
\includegraphics[width=2.3cm]{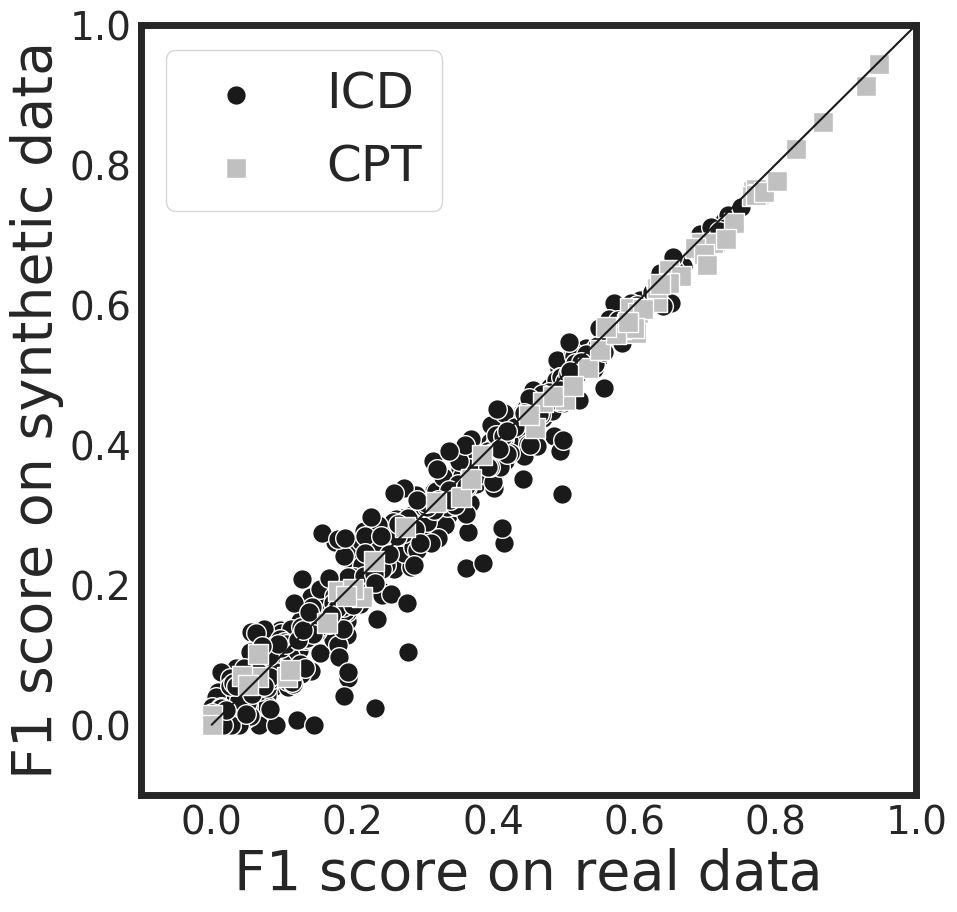}}%

\subfigure[Distribution of dot-to-diagonal distance of (a)]{
\label{dwp_5}%
\includegraphics[width=2.2cm]{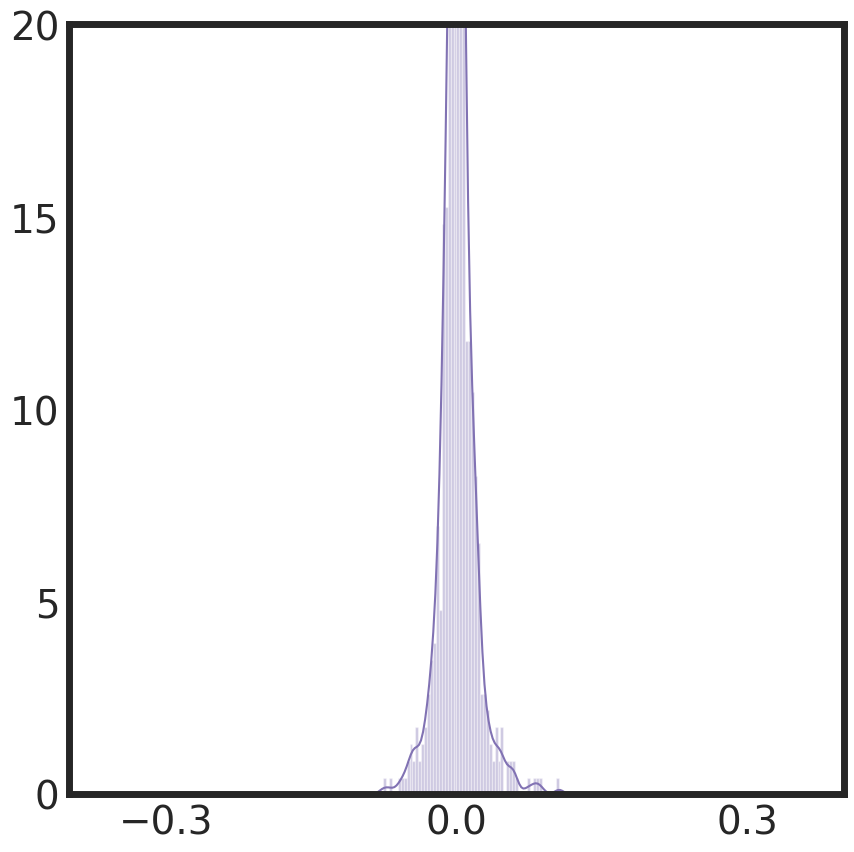}}\hspace{9mm}
\subfigure[Distribution of dot-to-diagonal distance of (b)]{%
\label{dwp_6}%
\includegraphics[width=2.2cm]{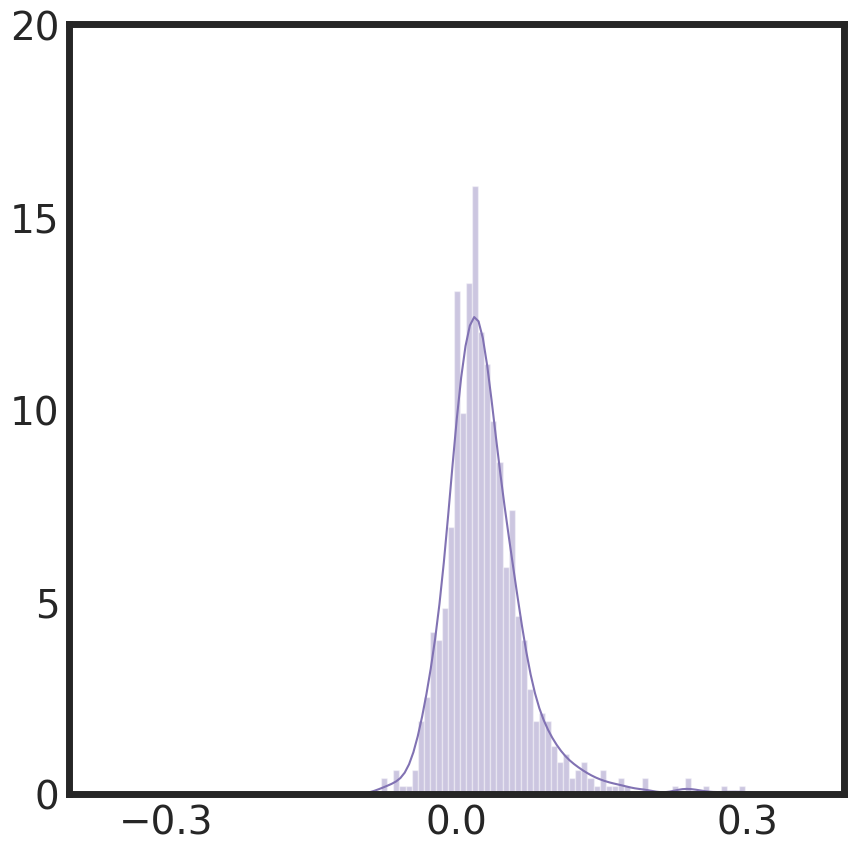}}\hspace{9mm}
\subfigure[Distribution of dot-to-diagonal distance of (c)]{%
\label{dwp_7}%
\includegraphics[width=2.2cm]{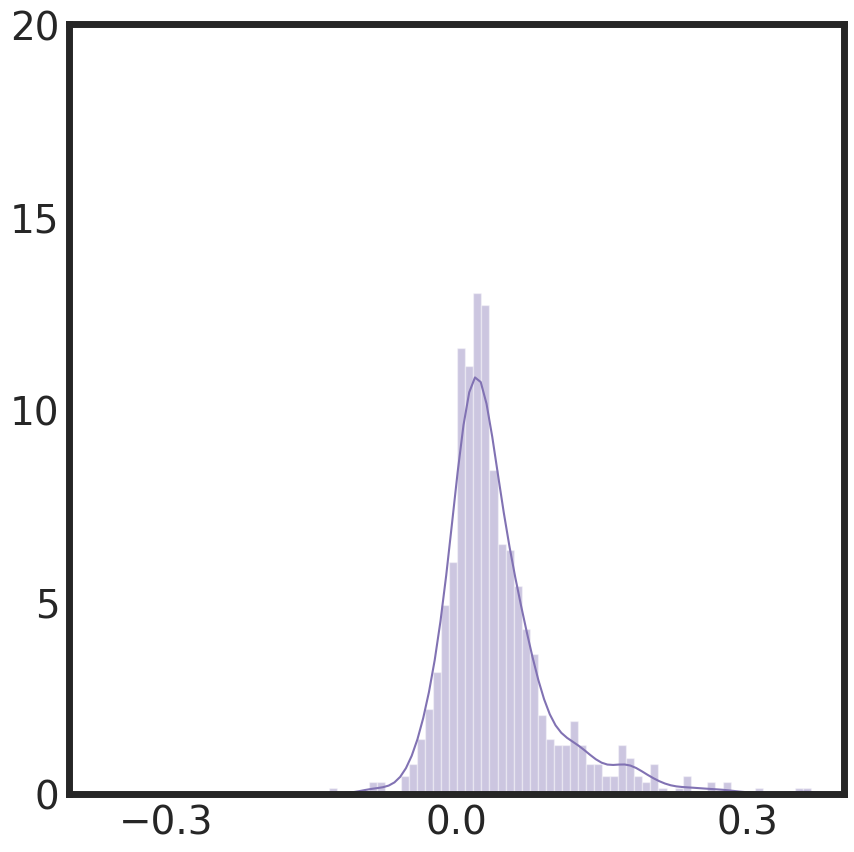}}\hspace{9mm}
\subfigure[Distribution of dot-to-diagonal distance of (d)]{%
\label{dwp_8}%
\includegraphics[width=2.2cm]{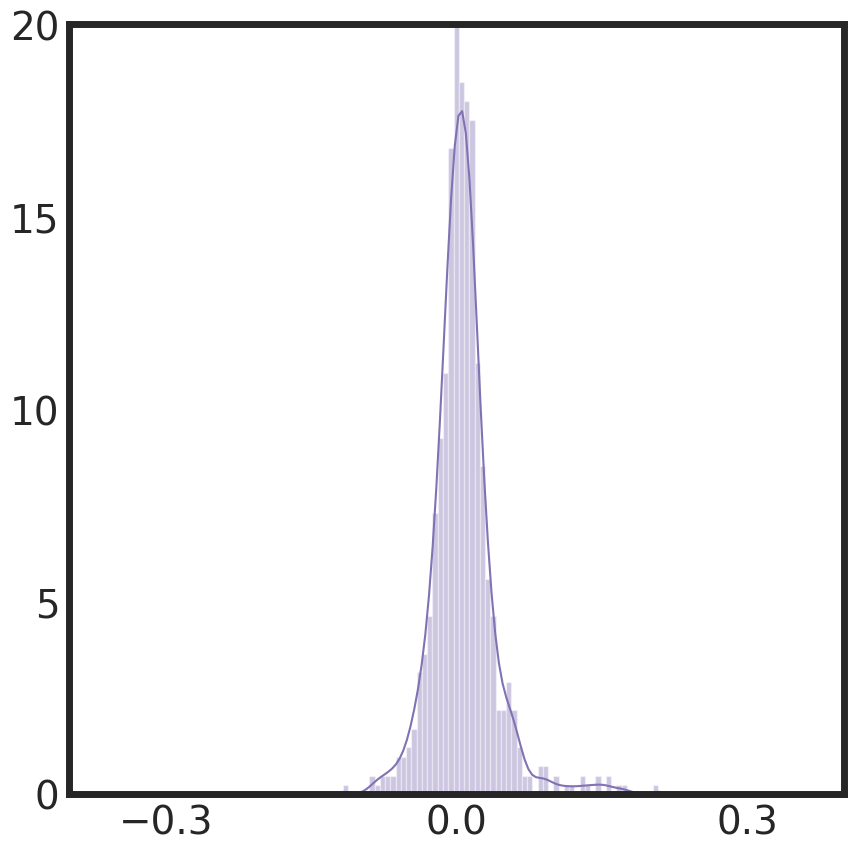}}%
\vspace*{-10mm}
\caption{Dimension-wise prediction. (a) F1 scores of logistic regression classifiers in the real \emph{vs} real setting. (b, c, d) Results of real \emph{vs} synthetic setting for the generative models. (e, f, g, h) Distributions of shortest distances from dots to the diagonal line for panels (a), (b), (c), and (d), respectively.}\label{dwp}
\end{figure}

\textbf{Dimension-wise prediction (DWP).}  
For each (ICD or CPT) code, two logistic regression classifiers were trained--one on the real data and one on synthetic data (with the same number of records as the real data). The binary status of a code served as the dependent variable while all remaining codes served as the independent variables. These classifiers were tested on a test dataset of the real EHRs (20\% of CSD). In the first row of Figure \ref{dwp}, each point denotes an ICD or CPT code, whose $x$ and $y$ value are the \emph{F1} score of the model trained on real and synthetic data, respectively (except \ref{dwp_1}--the real \emph{vs} real setting). There are several notable findings. First, Figure \ref{dwp_1} shows that the F1 scores in the real \emph{vs} real setting are closely distributed along the diagonal line without obvious skew. As such, the corresponding distribution of dot-to-diagonal distances, as shown in \ref{dwp_5}, is roughly symmetric, which indicates the stability of the cross-dimensional relationship in the original system. Second, the dot-to-diagonal distribution shown in Figure \ref{dwp_8} is less skewed than the one in Figure \ref{dwp_6}, which indicates that HGAN is better at representing the cross-dimensional relationships than EMR-CWGAN. Third, as depicted in Figure \ref{dwp_7}, it can be seen that there is skew towards the real data, suggesting that HGAN-U reduces the ability to learn relationships between features. Fourth, it can be seen from Figure \ref{dwp_4} that both types of codes behave similarly in the real \emph{vs} HGAN setting, which indicates that there is high similarity in the cross-dimensional performance between two types of codes.


\begin{figure}[ht]%
\centering
\subfigure{%
\includegraphics[width=2.0cm]{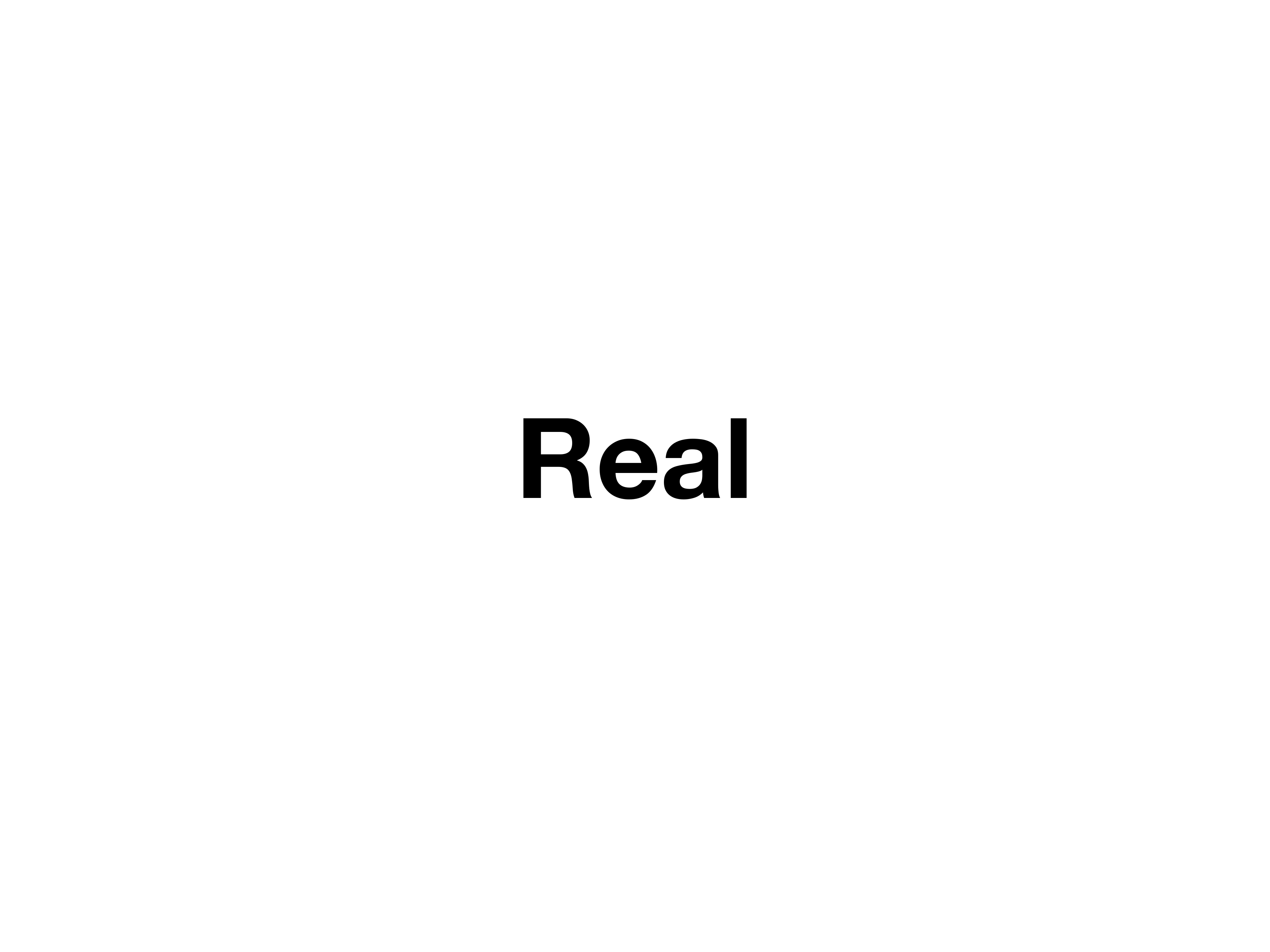}}
\subfigure{%
\includegraphics[width=2.0cm]{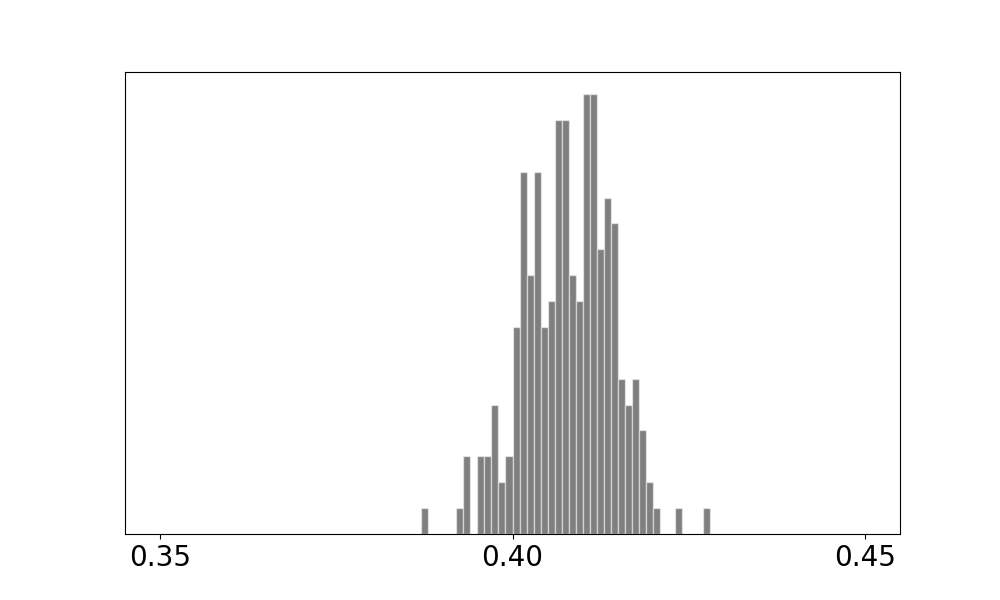}}
\subfigure{%
\includegraphics[width=2.0cm]{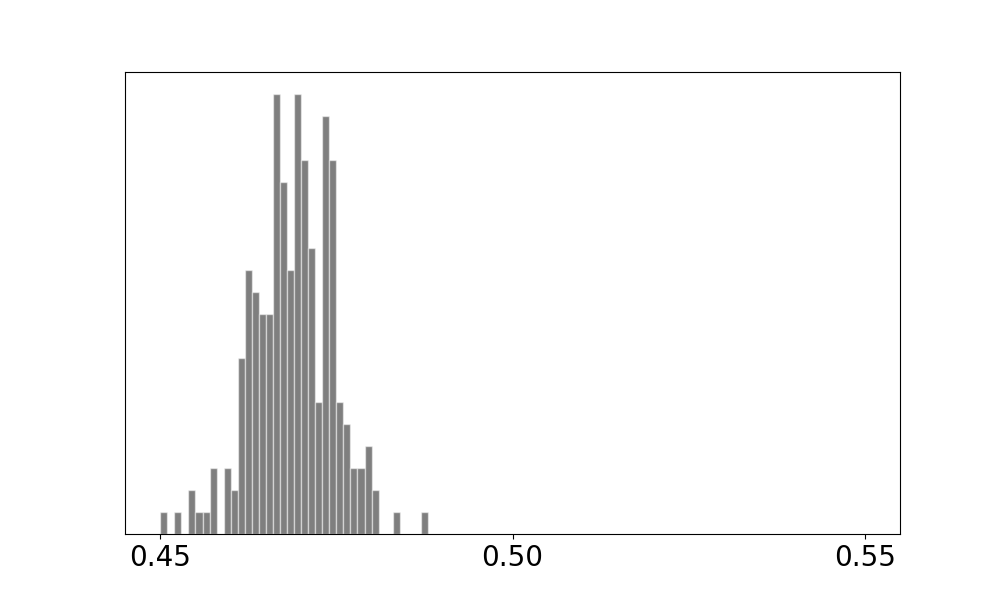}}
\subfigure{%
\includegraphics[width=2.0cm]{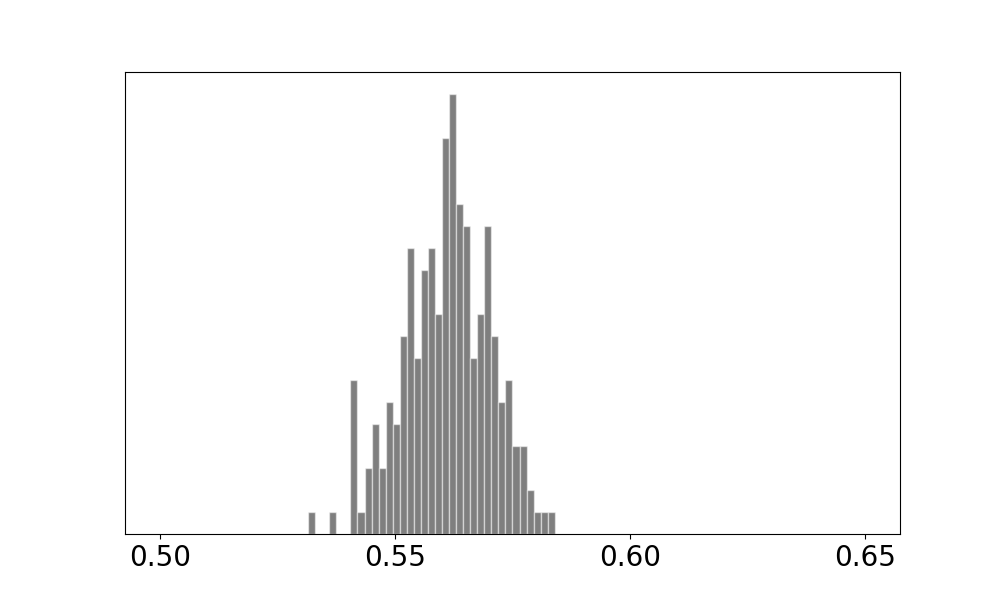}}%
\subfigure{%
\includegraphics[width=2.0cm]{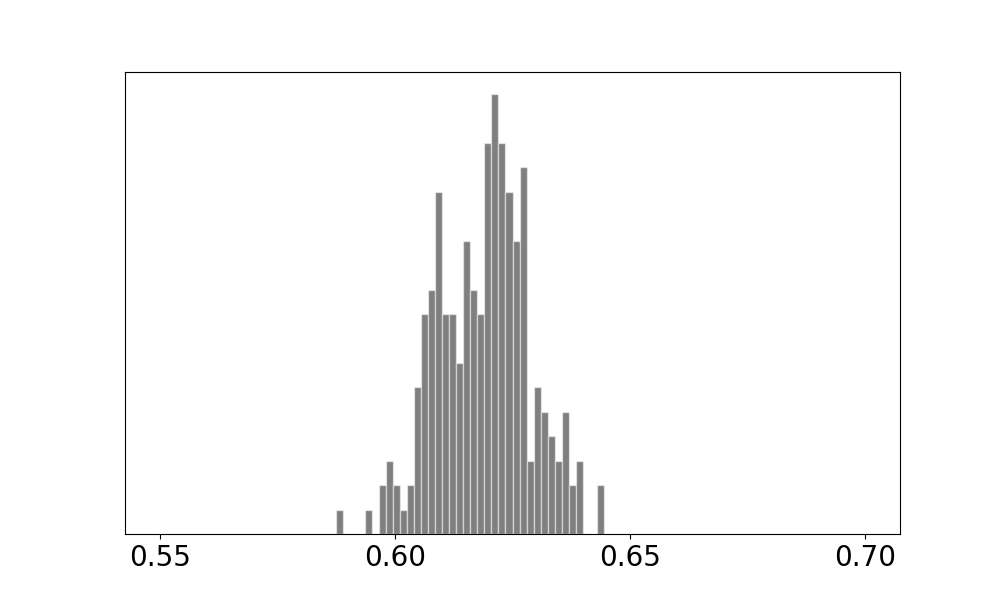}}%
\vspace*{-7mm}

\subfigure{%
\includegraphics[width=2.0cm]{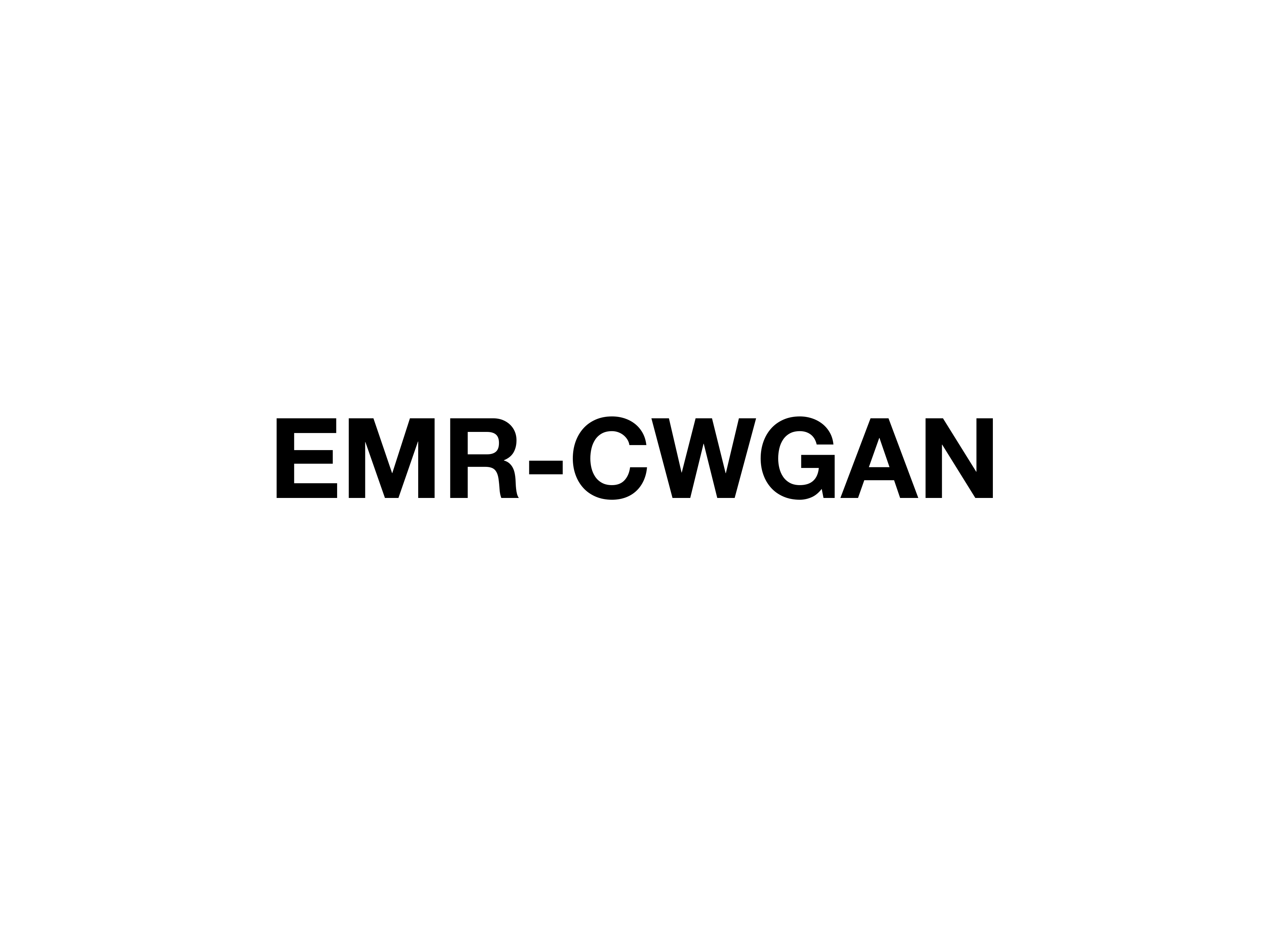}}
\subfigure{%
\includegraphics[width=2.0cm]{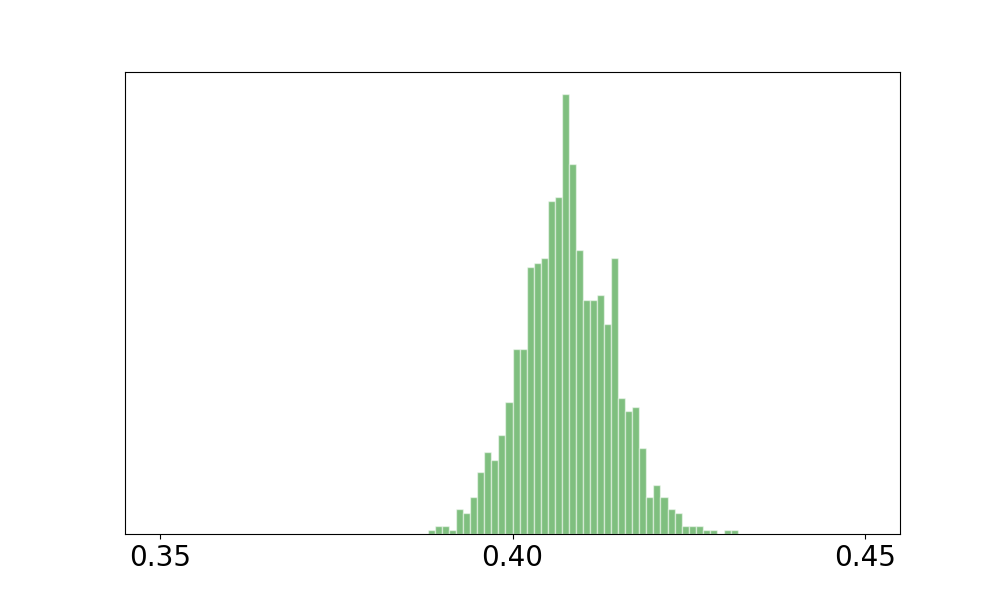}}
\subfigure{%
\includegraphics[width=2.0cm]{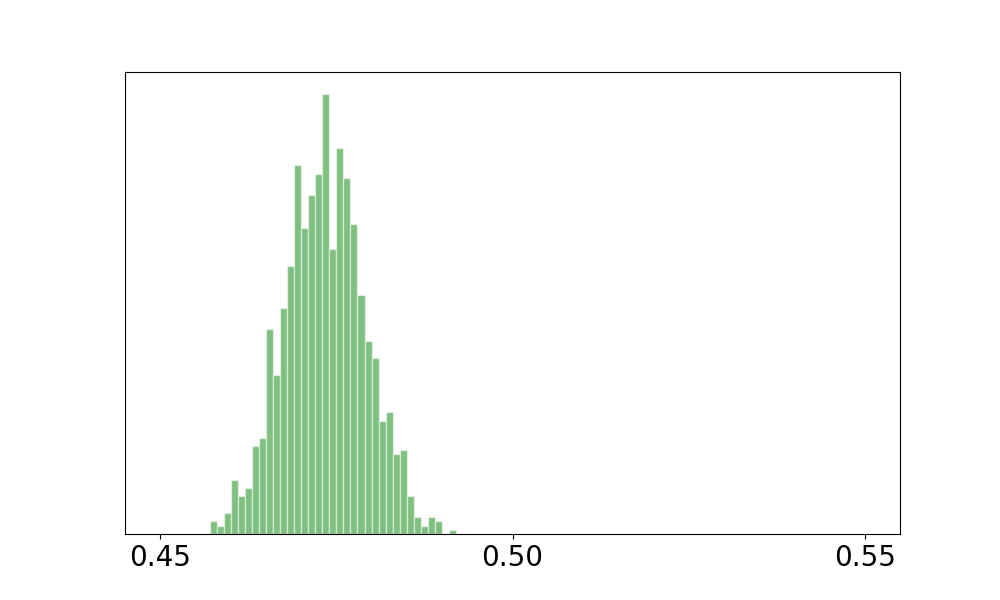}}
\subfigure{%
\includegraphics[width=2.0cm]{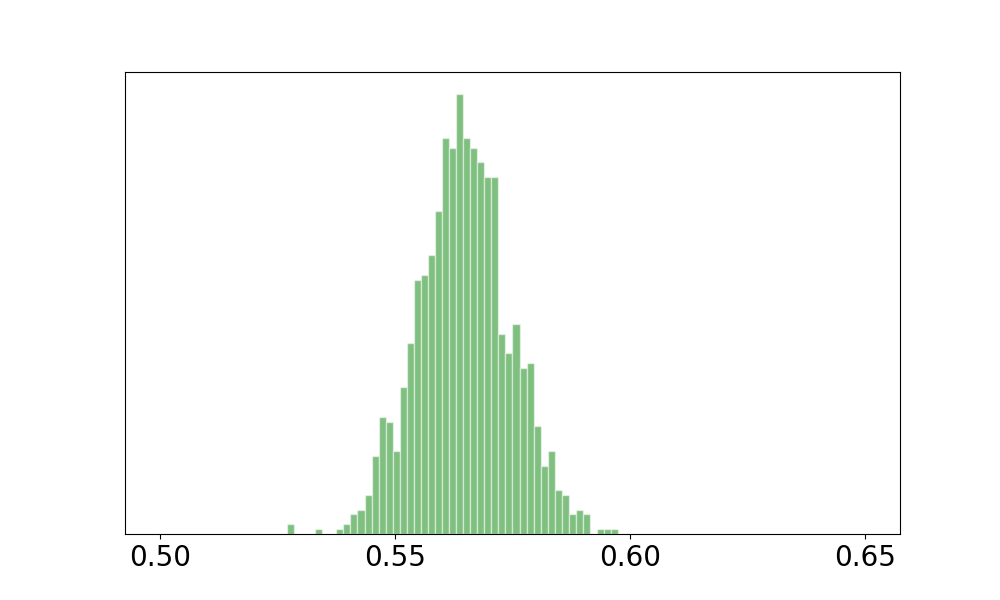}}%
\subfigure{%
\includegraphics[width=2.0cm]{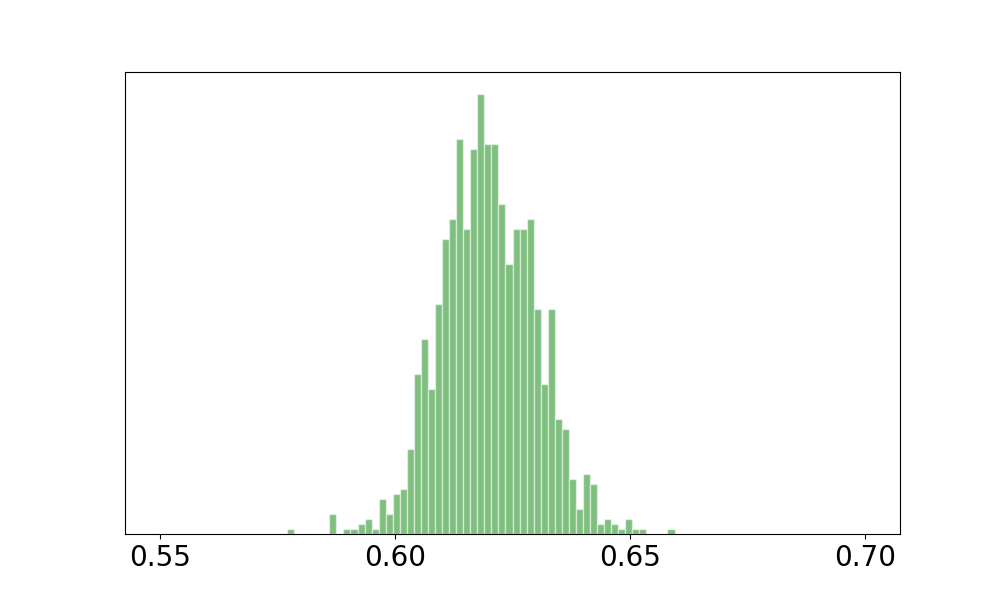}}%
\vspace*{-7mm}

\subfigure{%
\includegraphics[width=2.0cm]{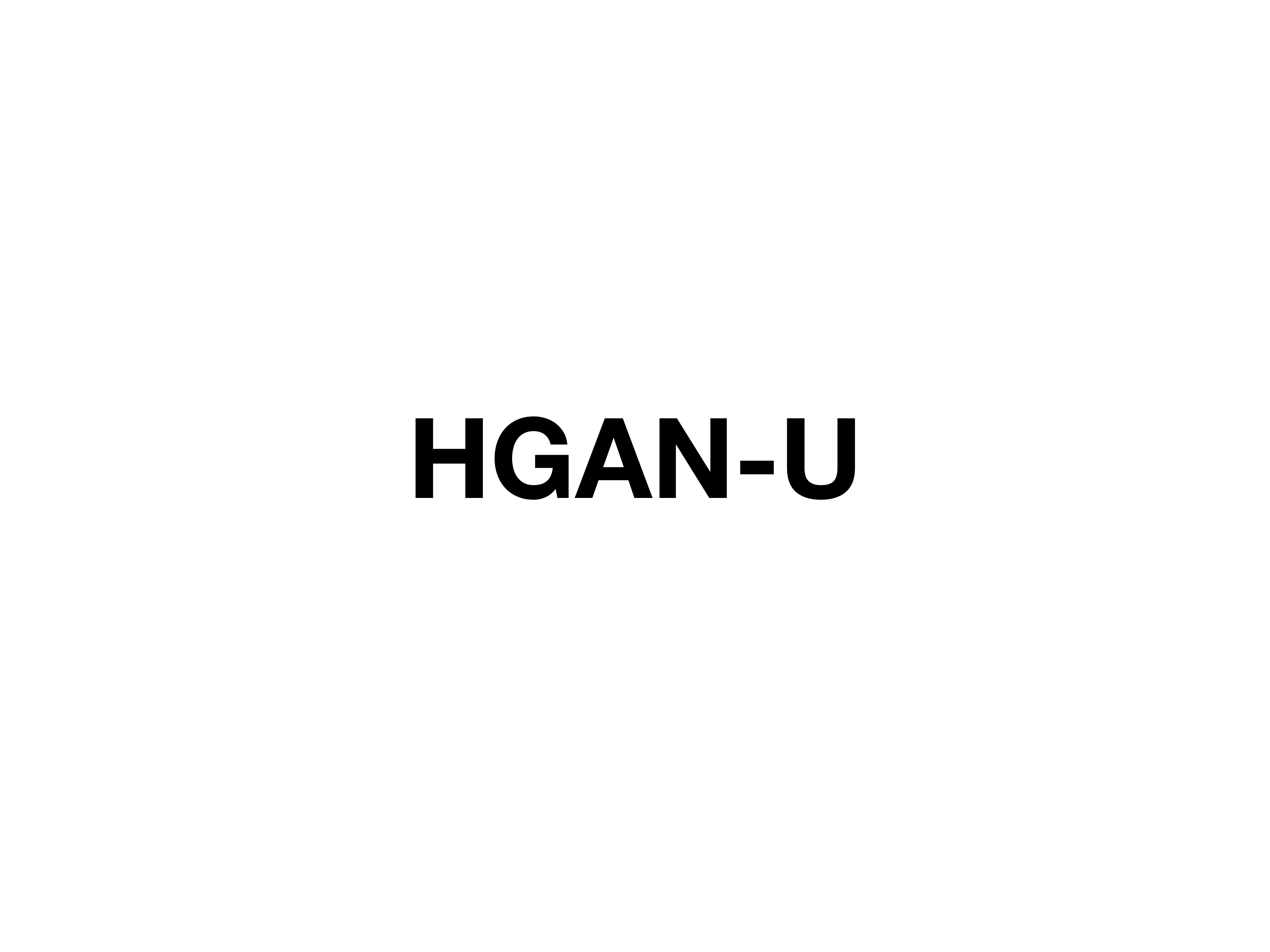}}
\subfigure{%
\includegraphics[width=2.0cm]{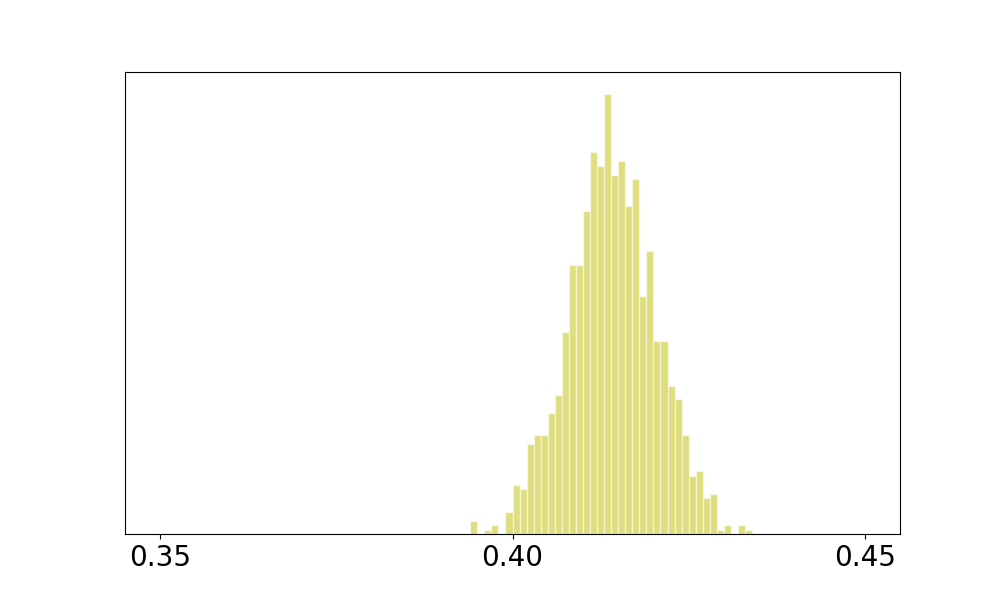}}
\subfigure{%
\includegraphics[width=2.0cm]{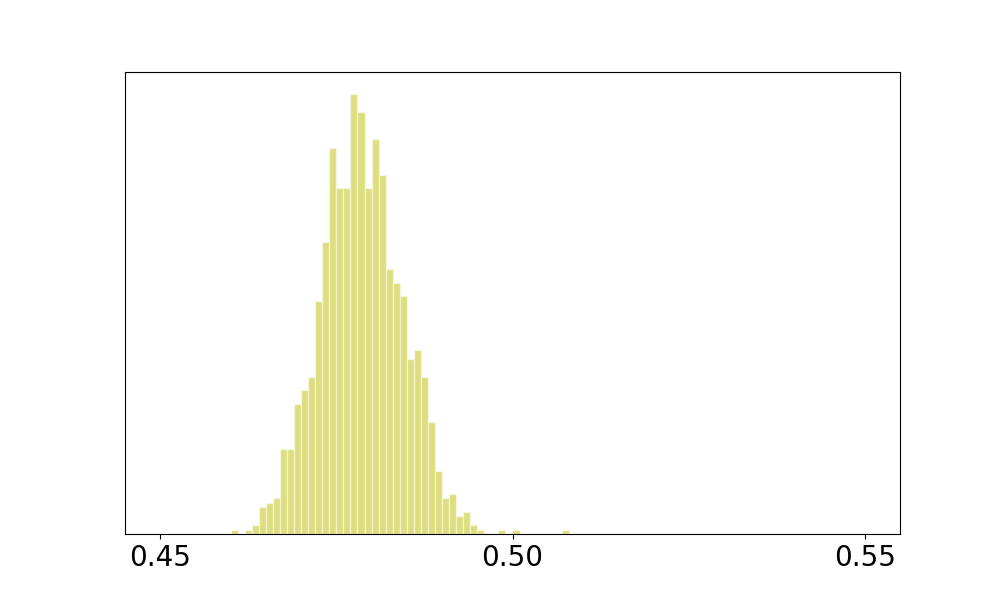}}
\subfigure{%
\includegraphics[width=2.0cm]{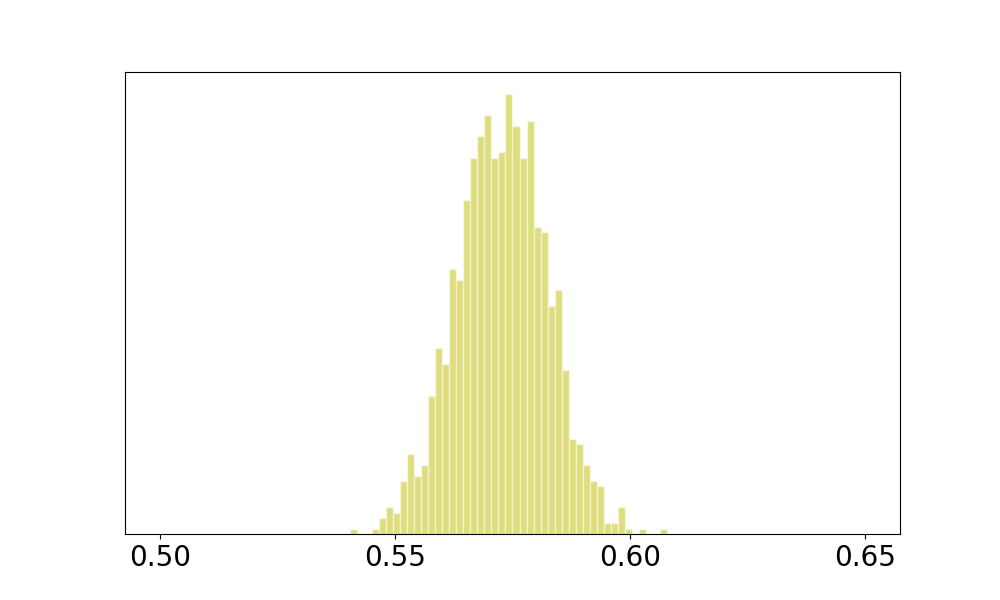}}%
\subfigure{%
\includegraphics[width=2.0cm]{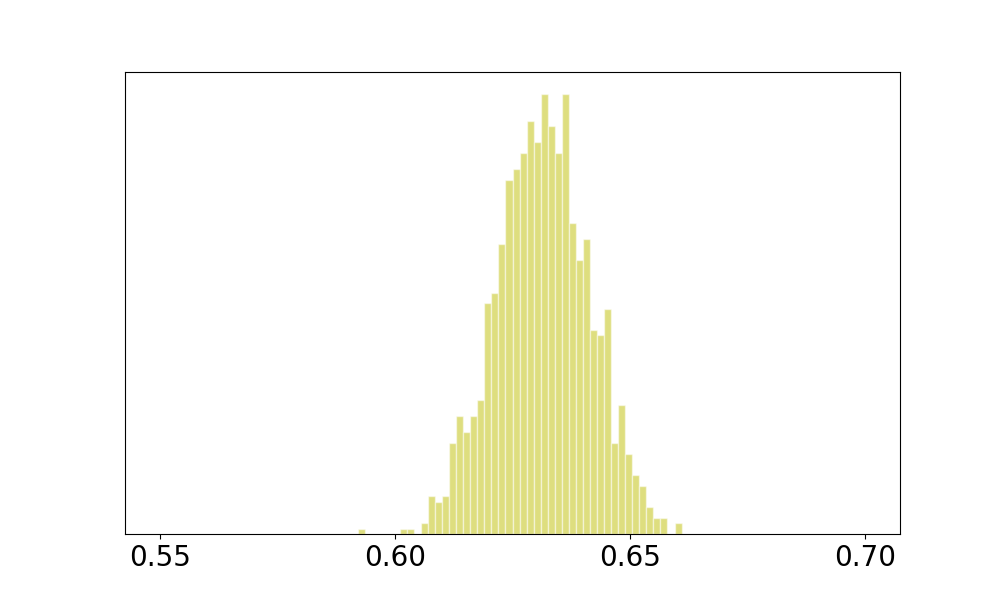}}%
\vspace*{-7mm}

\subfigure{%
\includegraphics[width=2.0cm]{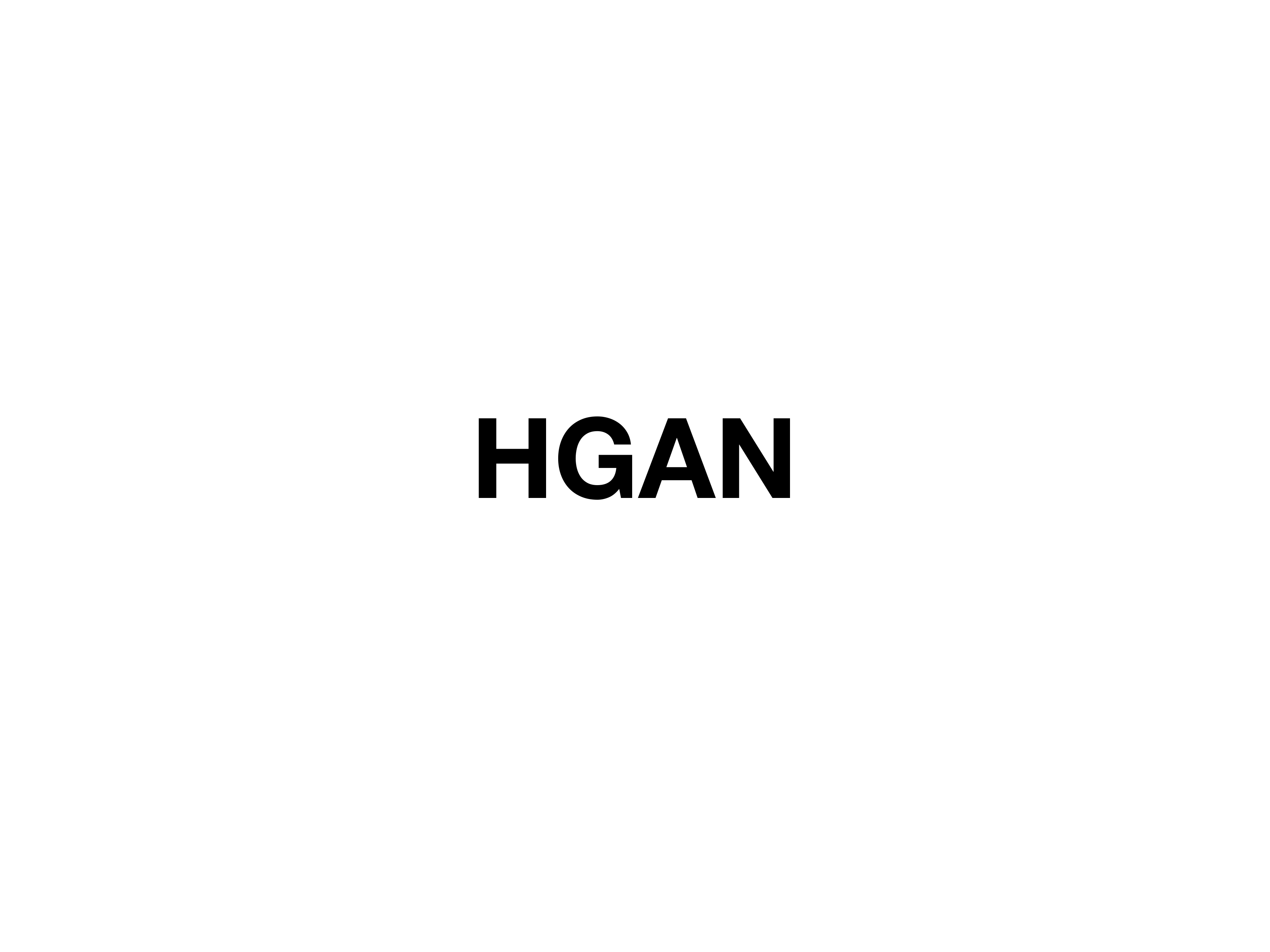}}
\subfigure{%
\includegraphics[width=2.0cm]{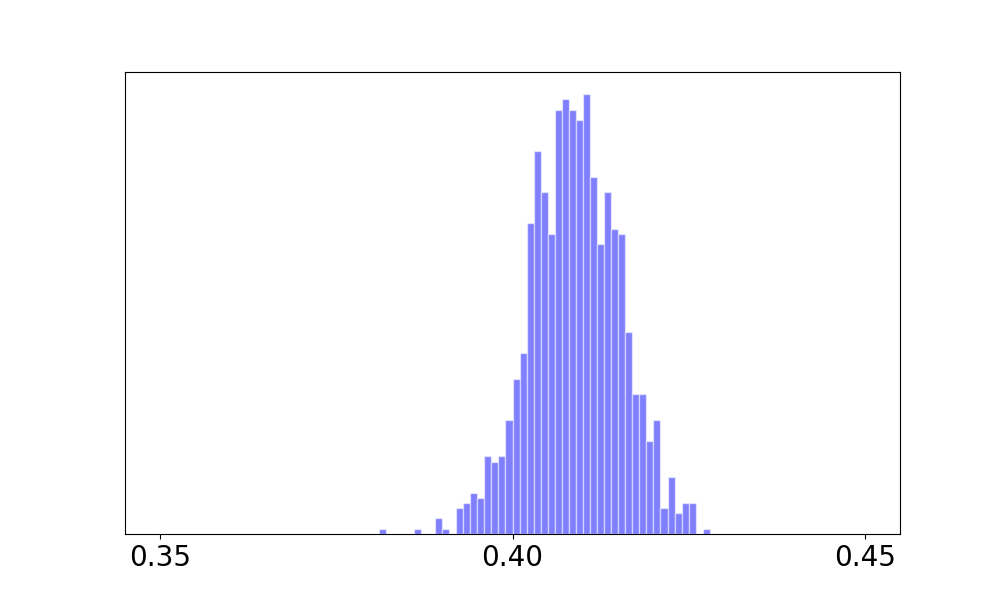}}
\subfigure{%
\includegraphics[width=2.0cm]{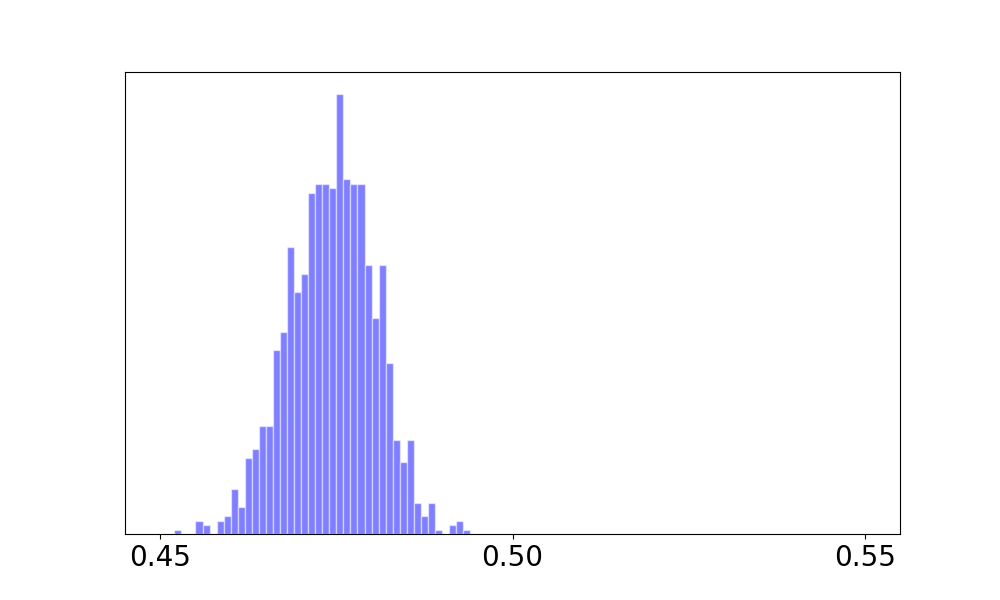}}
\subfigure{%
\includegraphics[width=2.0cm]{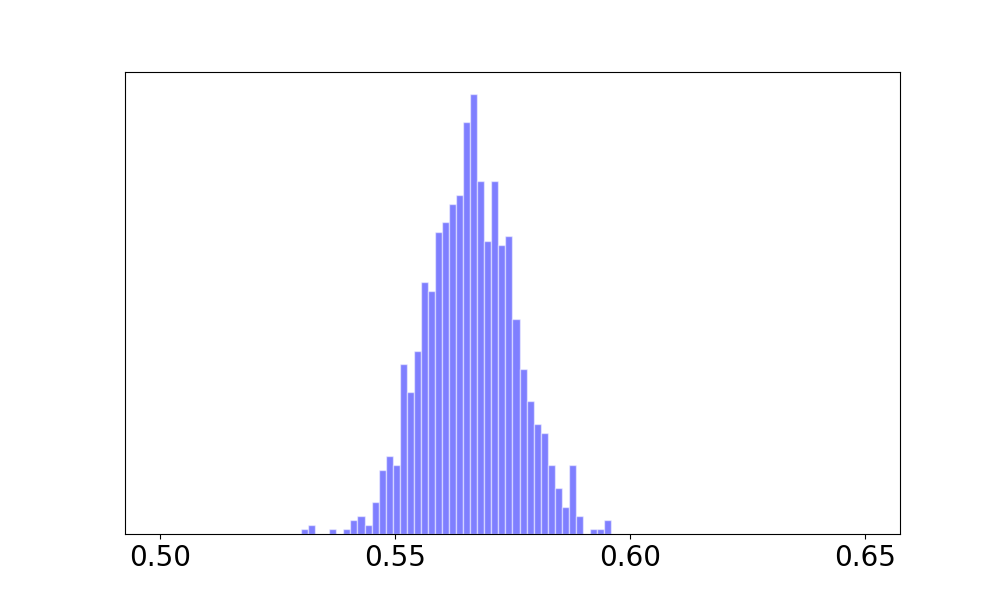}}%
\subfigure{%
\includegraphics[width=2.0cm]{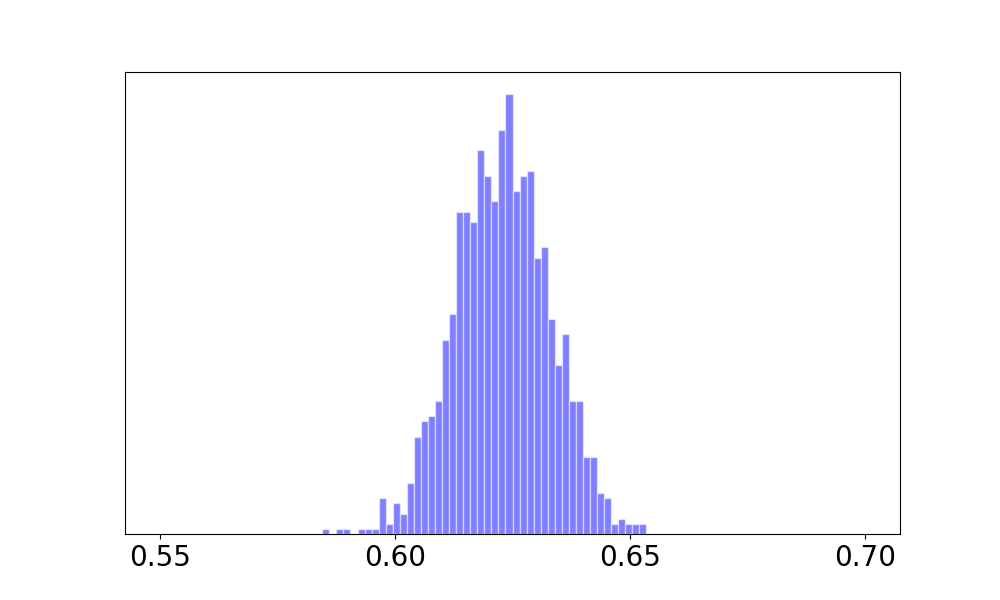}}%
\vspace*{-1mm}

\subfigure{%
\includegraphics[width=2.0cm]{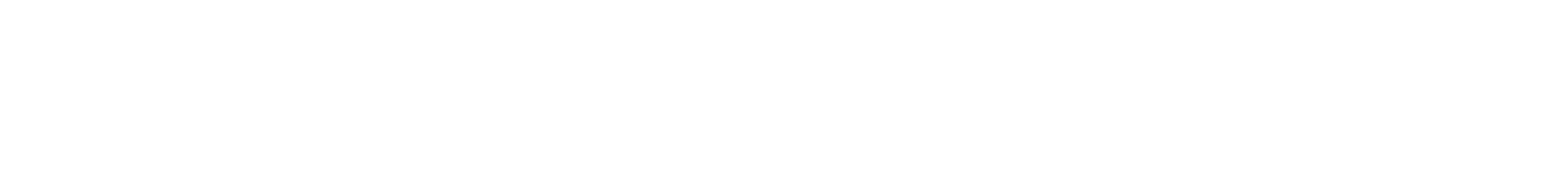}}
\subfigure{%
\includegraphics[width=2.0cm]{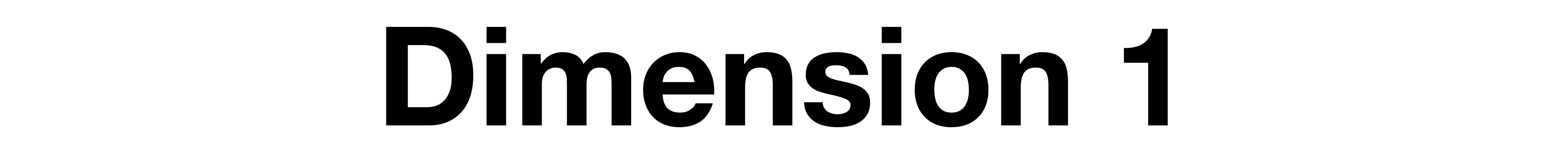}}
\subfigure{%
\includegraphics[width=2.0cm]{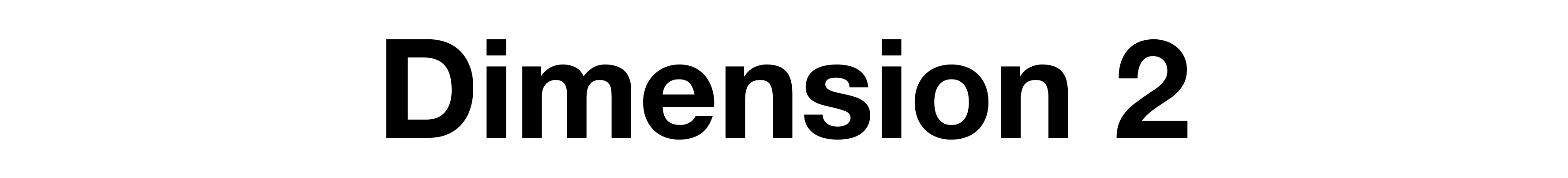}}
\subfigure{%
\includegraphics[width=2.0cm]{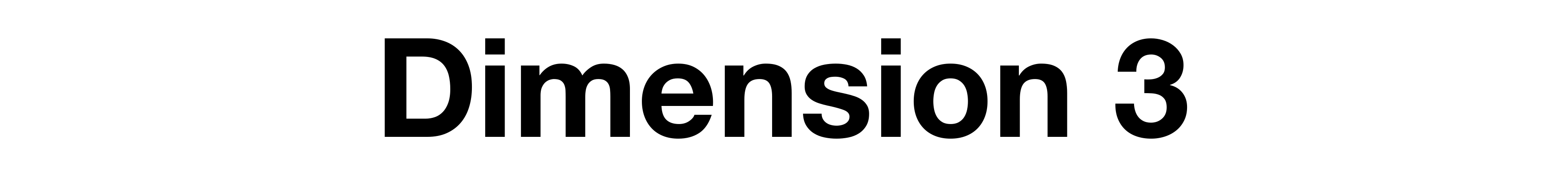}}%
\subfigure{%
\includegraphics[width=2.0cm]{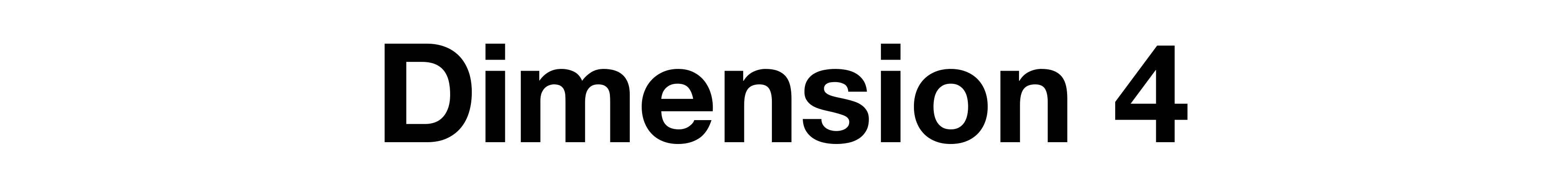}}%
\vspace*{-10mm}
\caption{Latent space representation. Each subfigure illustrates the distribution of the variances in one latent dimension. The topmost row corresponds to the real EHR data. Each subsequent row corresponds to a synthetic data generation model.}\label{lsr}
\end{figure}

\textbf{Latent space representation (LSR).}
At a high level, the investigation into LSR has three main steps: 1) use real data to train a $\beta$-Variational Auto-Encoder ($\beta$-VAE) \cite{Higgins2017betaVAELB}, a tool to disentangle the latent factors in data by forgetting less important latent dimensions, to discover their efficient latent dimensions for data reconstruction, as well as the corresponding distribution of variances, 2) input the generated data into the trained $\beta$-VAE model and assess their variance distributions on the latent dimensions, and 3) for each efficient latent dimension, assess whether the distribution of variances obtained from 2) is pushed closer to $1$, which is an indicator of losing important structural properties in real data. 
Figure \ref{lsr} shows the results for LSR, where we selected all efficient latent dimensions with the mean of the variance distribution in the real data less than  $0.70$. Both EMR-CWGAN and HGAN are highly similar to the real data, suggesting that both retain the latent structural properties. By contrast, HGAN-U exhibits a large shift in the variance distributions in all dimensions, which suggests it is less capable of representing key structures in the latent space.

\captionsetup[subfigure]{labelformat=empty}

\begin{figure}[ht]%
\centering
\subfigure{%
\label{far1_1}%
\includegraphics[width=1.7cm]{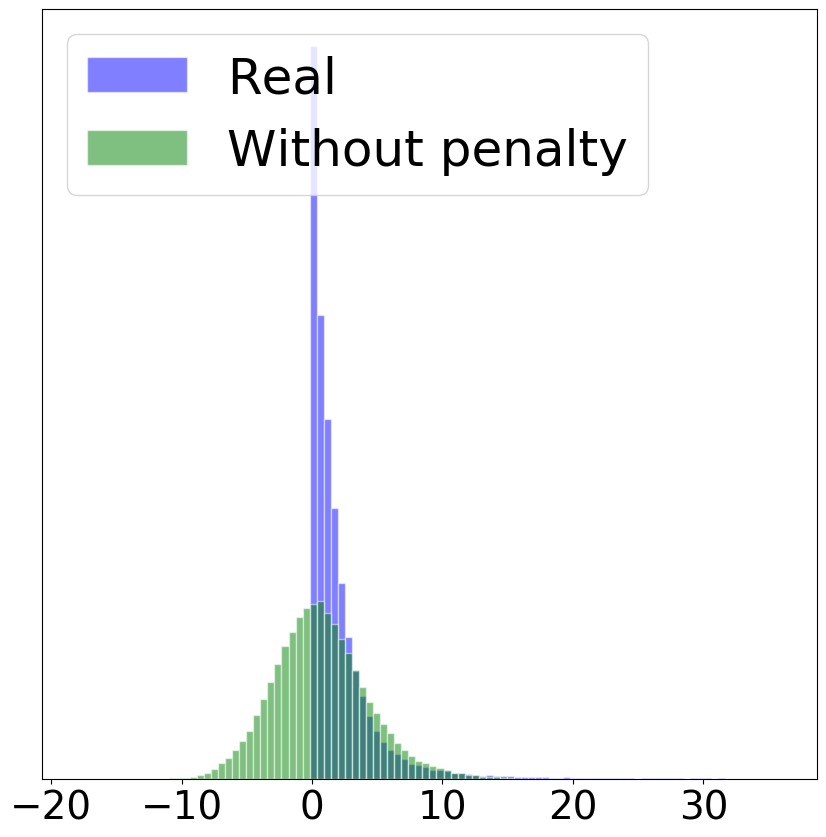}}
\subfigure{%
\label{far1_2}%
\includegraphics[width=1.7cm]{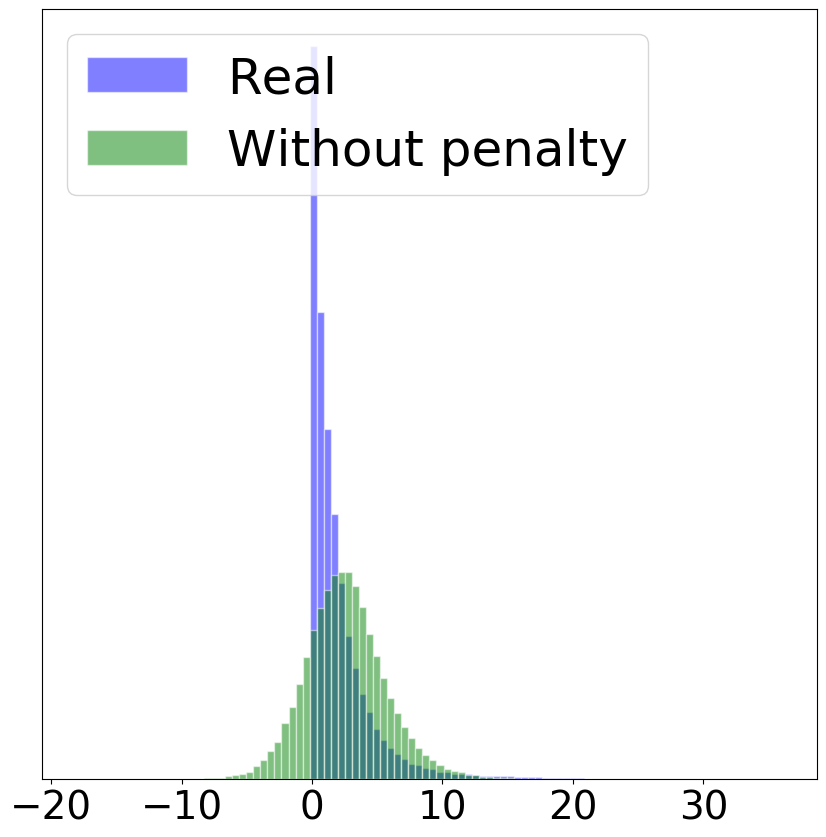}}
\subfigure{%
\label{far1_3}%
\includegraphics[width=1.7cm]{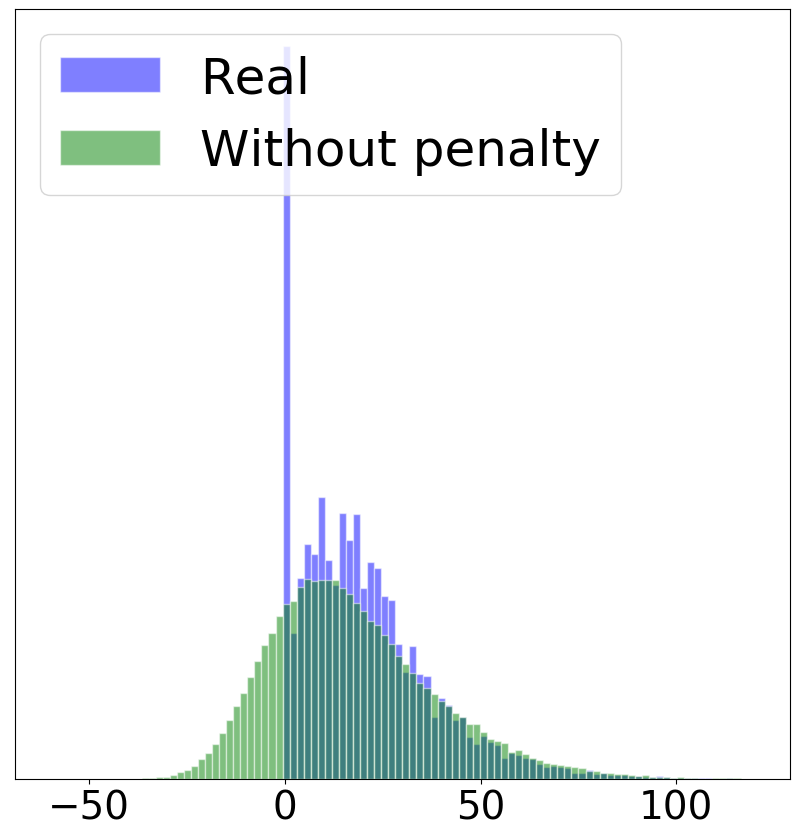}}
\subfigure{%
\label{far1_4}%
\includegraphics[width=1.7cm]{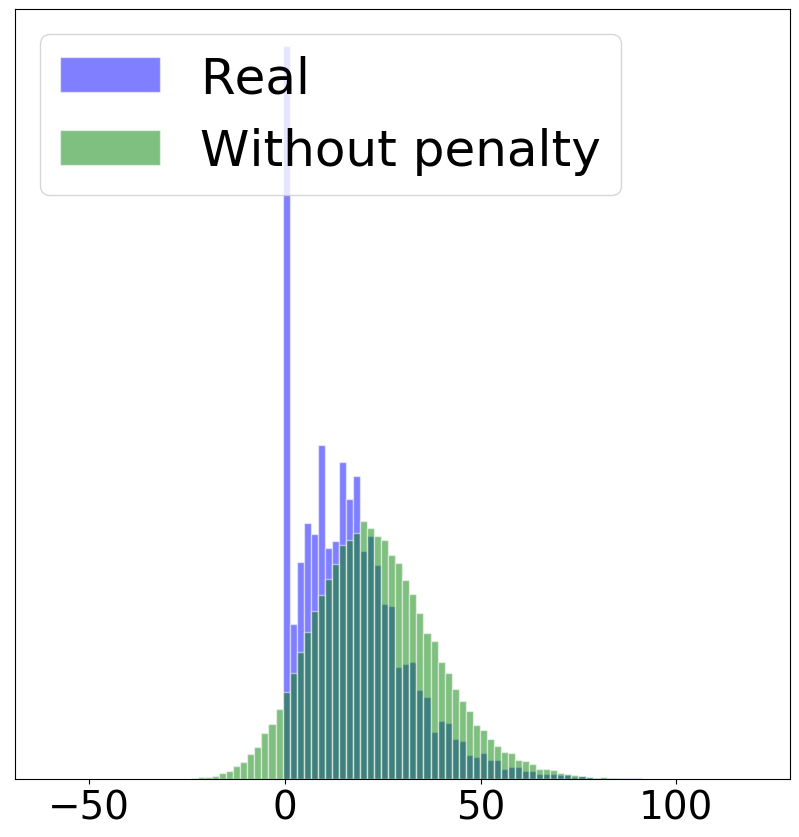}}
\subfigure{%
\label{far1_5}%
\includegraphics[width=1.7cm]{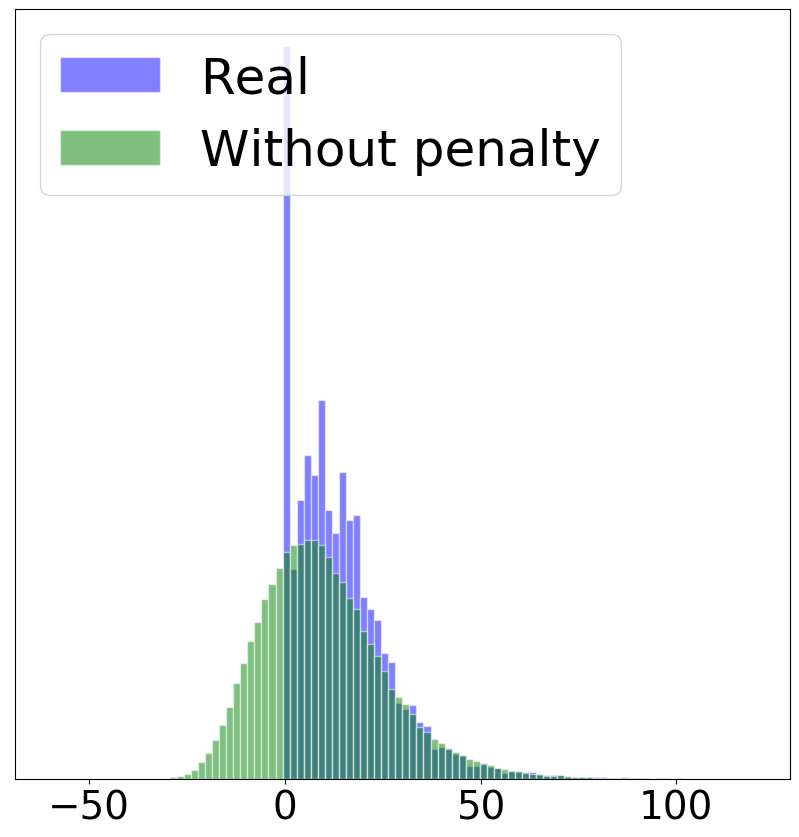}}
\subfigure{%
\label{far1_6}%
\includegraphics[width=1.7cm]{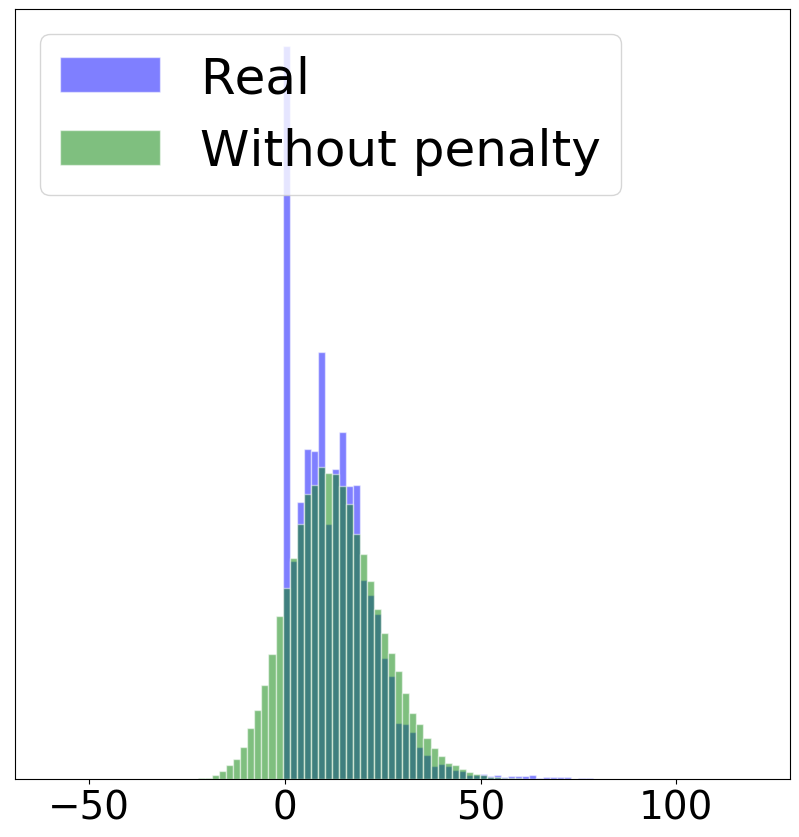}}
\subfigure{%
\label{far1_9}%
\includegraphics[width=1.7cm]{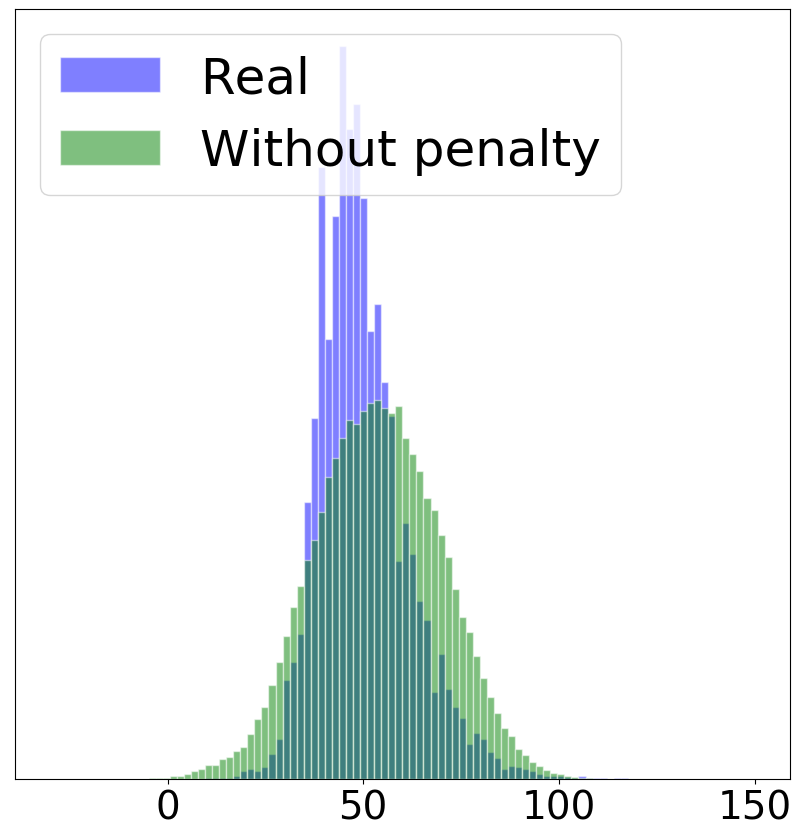}}
\subfigure{%
\label{far1_8}%
\includegraphics[width=1.7cm]{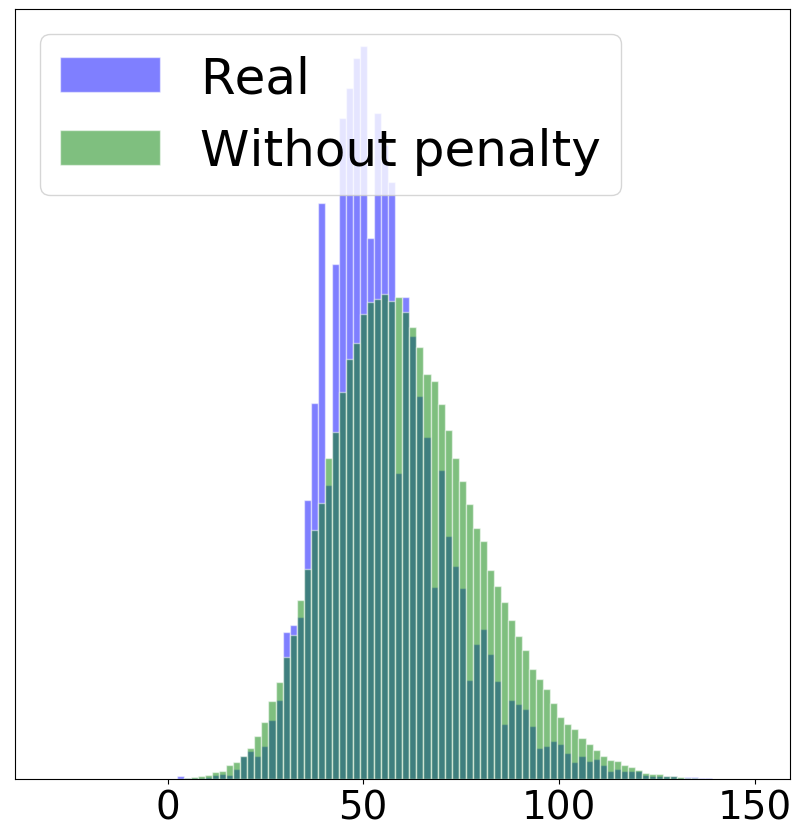}}
\subfigure{%
\label{far1_7}%
\includegraphics[width=1.7cm]{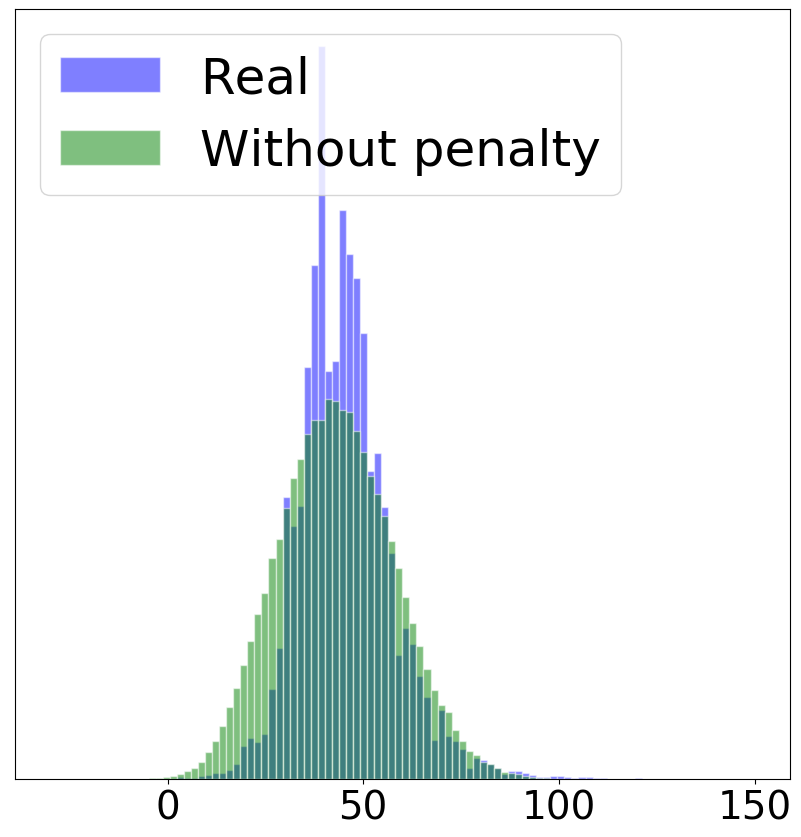}}

\subfigure{%
\label{far2_1}%
\includegraphics[width=1.7cm]{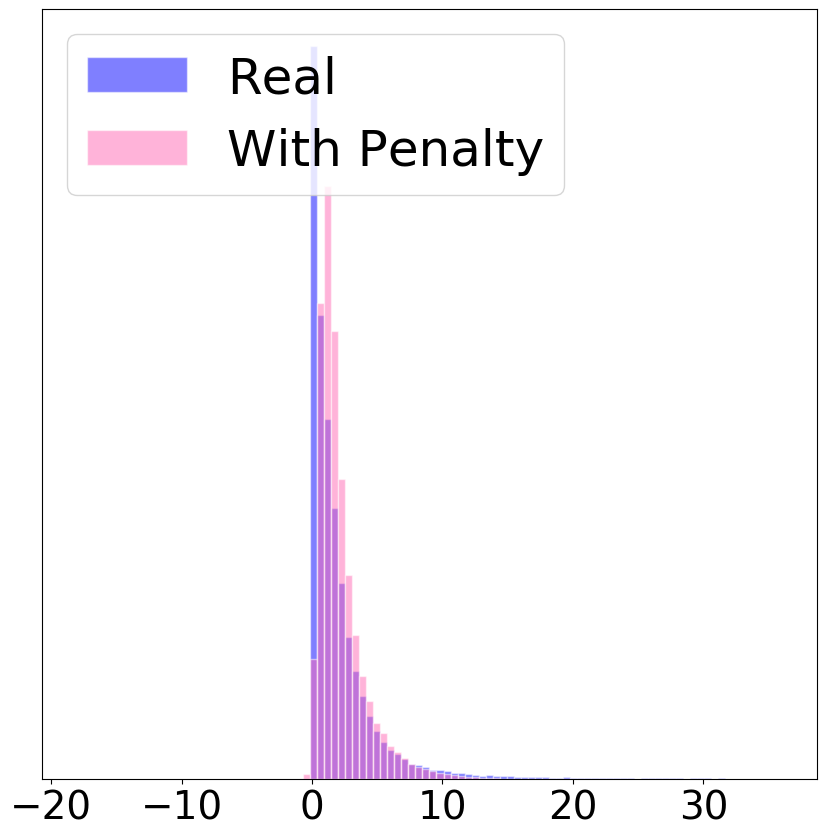}}
\subfigure{%
\label{far2_2}%
\includegraphics[width=1.7cm]{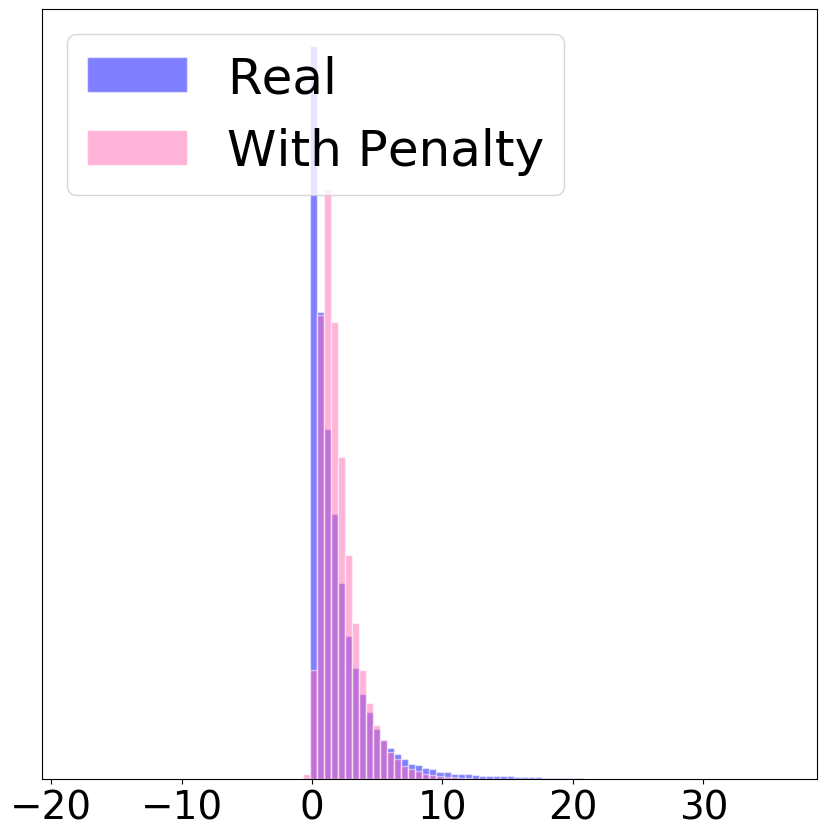}}
\subfigure{%
\label{far2_3}%
\includegraphics[width=1.7cm]{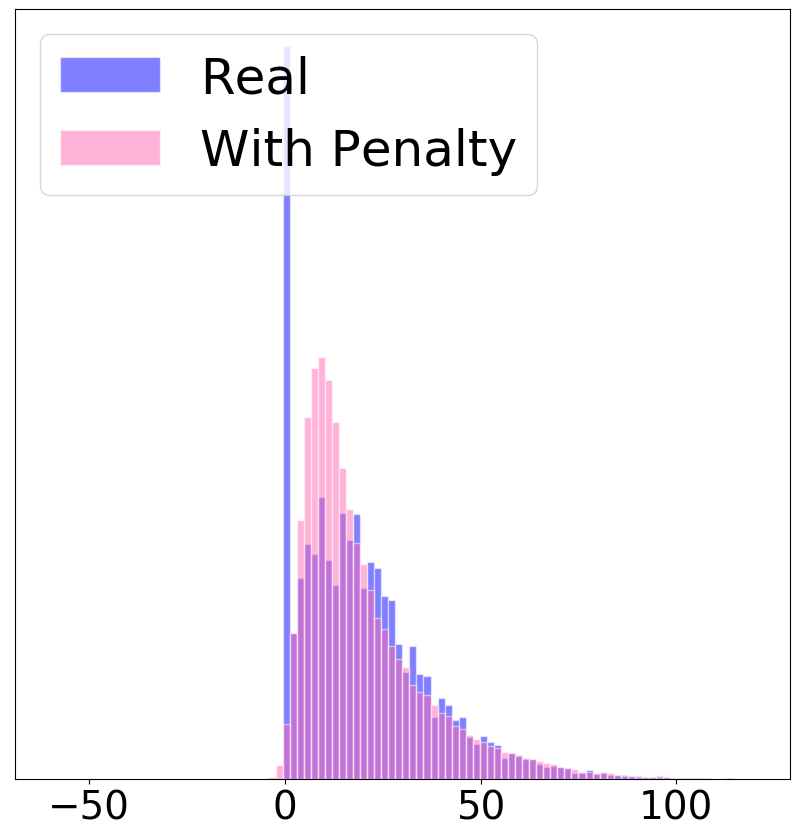}}
\subfigure{%
\label{far2_4}%
\includegraphics[width=1.7cm]{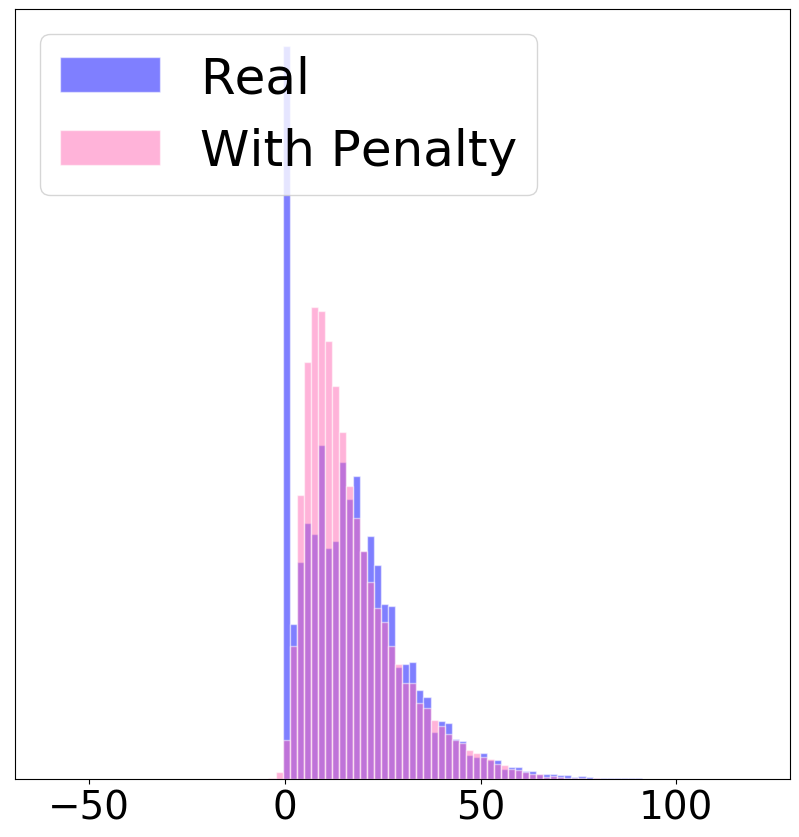}}
\subfigure{%
\label{far2_5}%
\includegraphics[width=1.7cm]{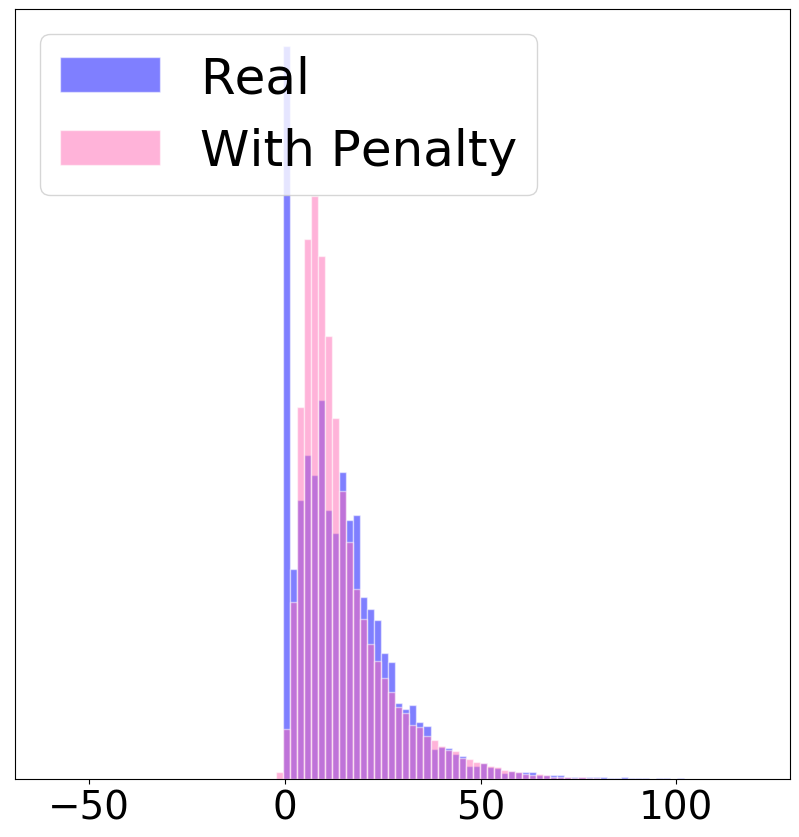}}
\subfigure{%
\label{far2_6}%
\includegraphics[width=1.7cm]{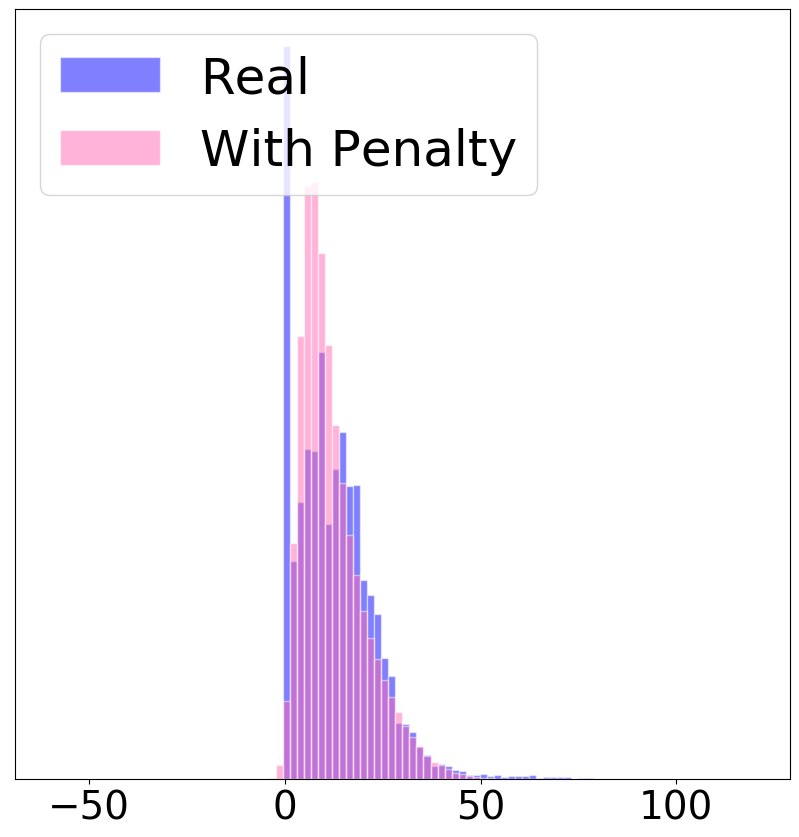}}
\subfigure{%
\label{far2_9}%
\includegraphics[width=1.7cm]{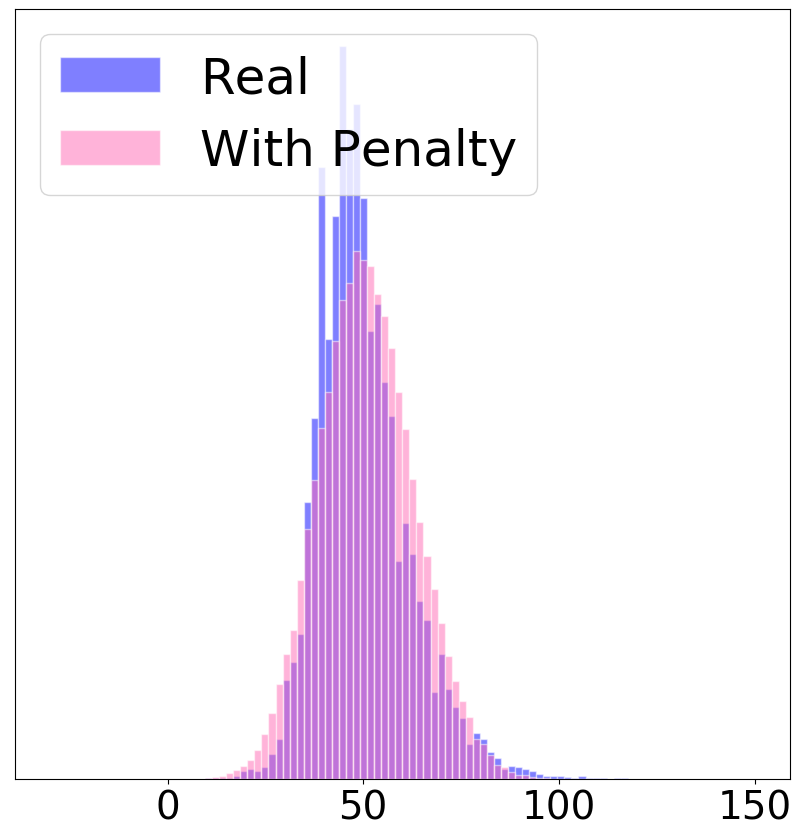}}
\subfigure{%
\label{far2_8}%
\includegraphics[width=1.7cm]{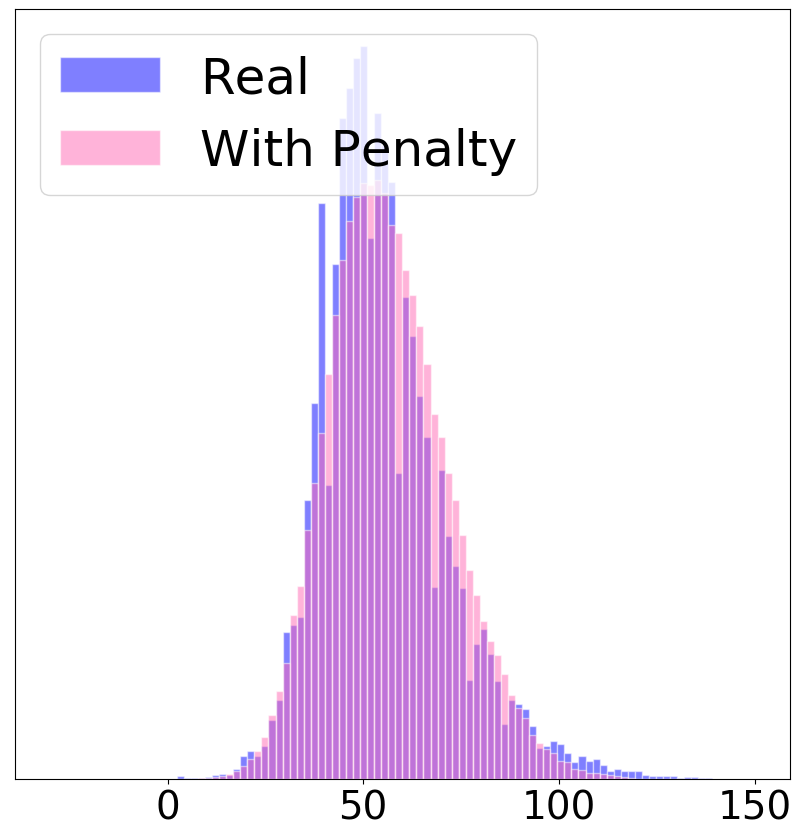}}
\subfigure{%
\label{far2_7}%
\includegraphics[width=1.7cm]{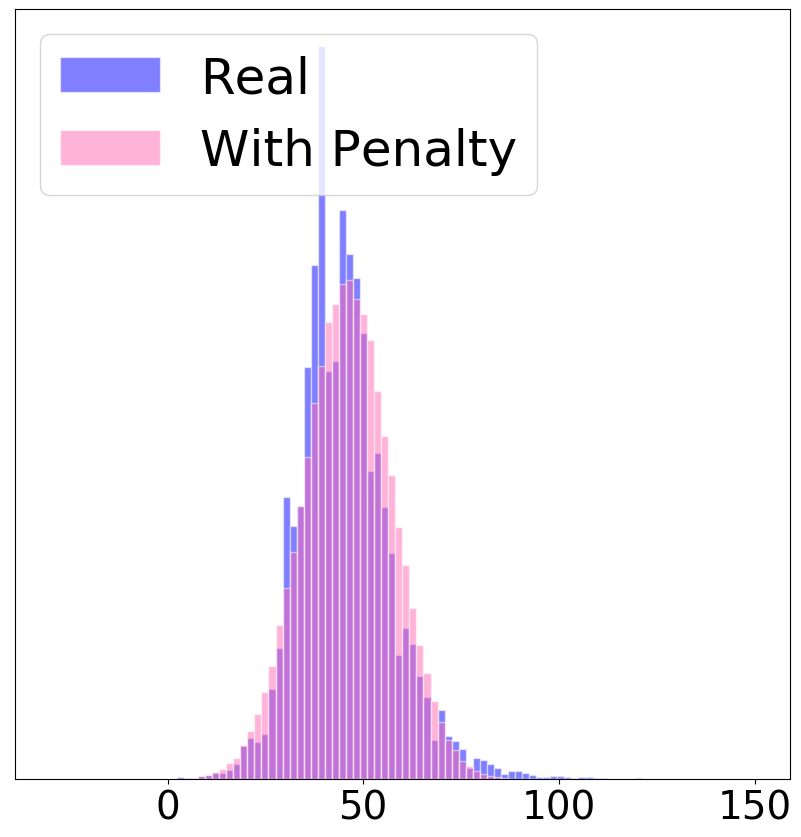}}
\vspace*{-3mm}

\subfigure{%
\label{far3_1}%
\includegraphics[width=1.7cm]{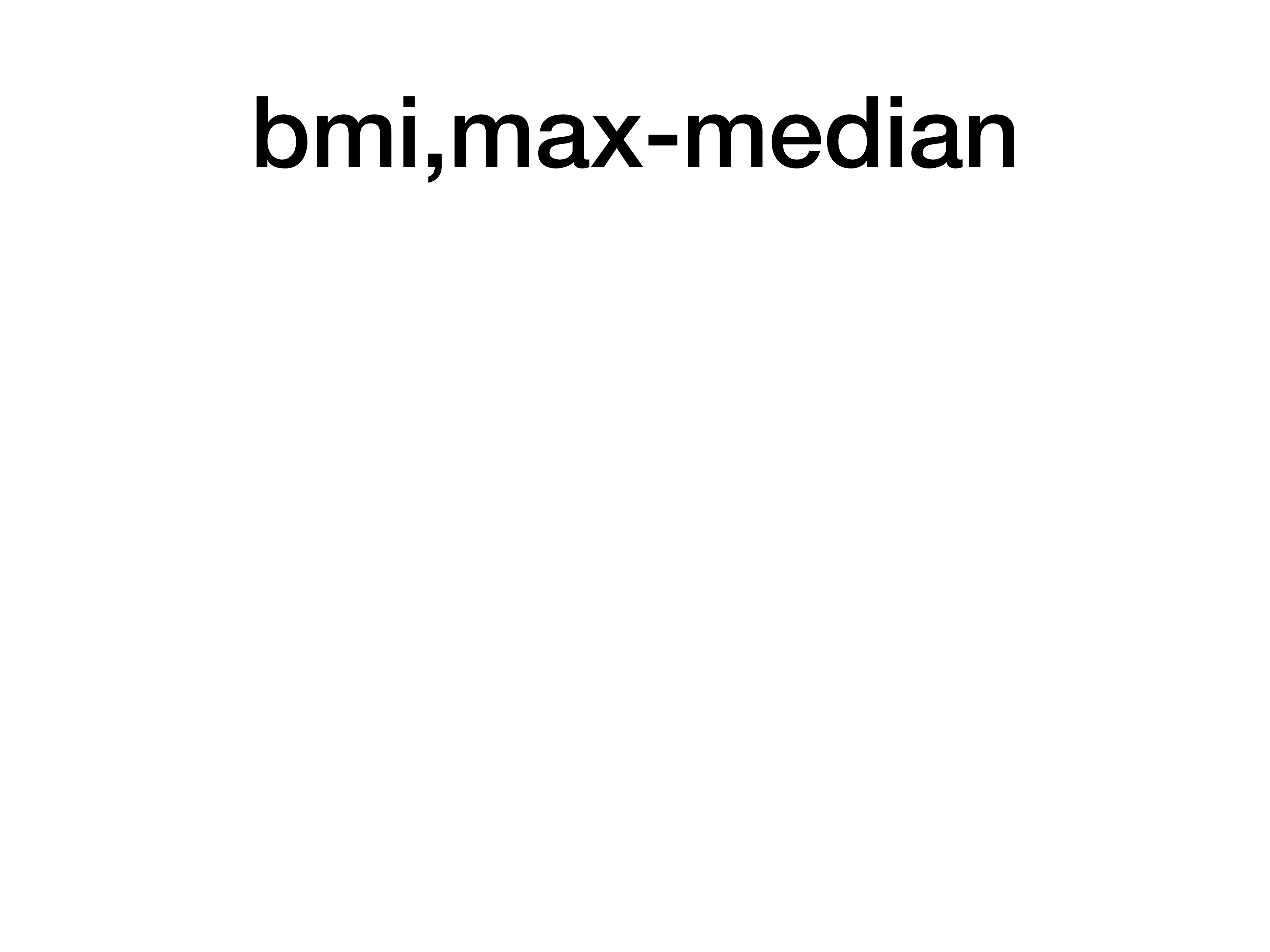}}
\subfigure{%
\label{far3_2}%
\includegraphics[width=1.7cm]{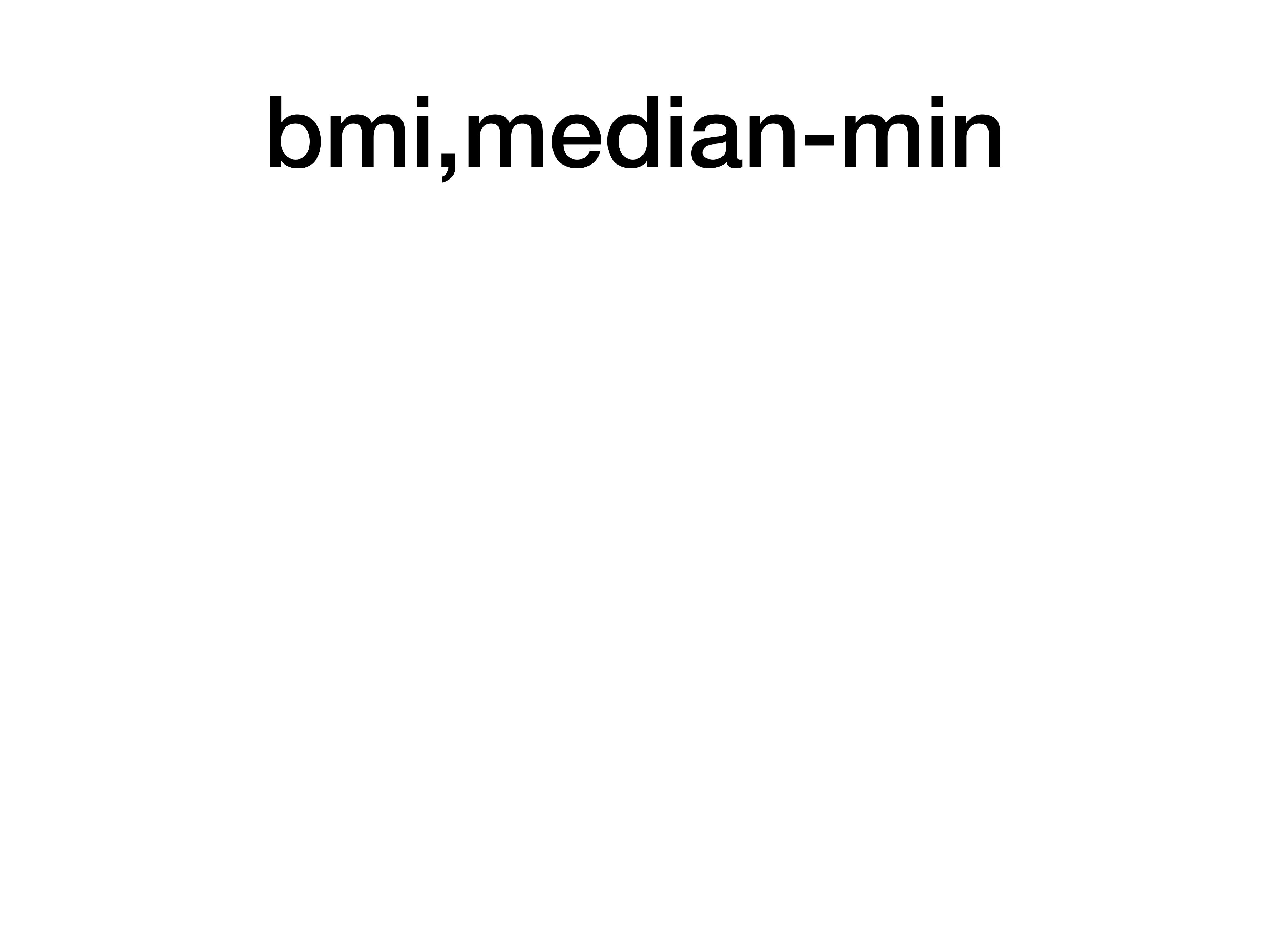}}
\subfigure{%
\label{far3_3}%
\includegraphics[width=1.7cm]{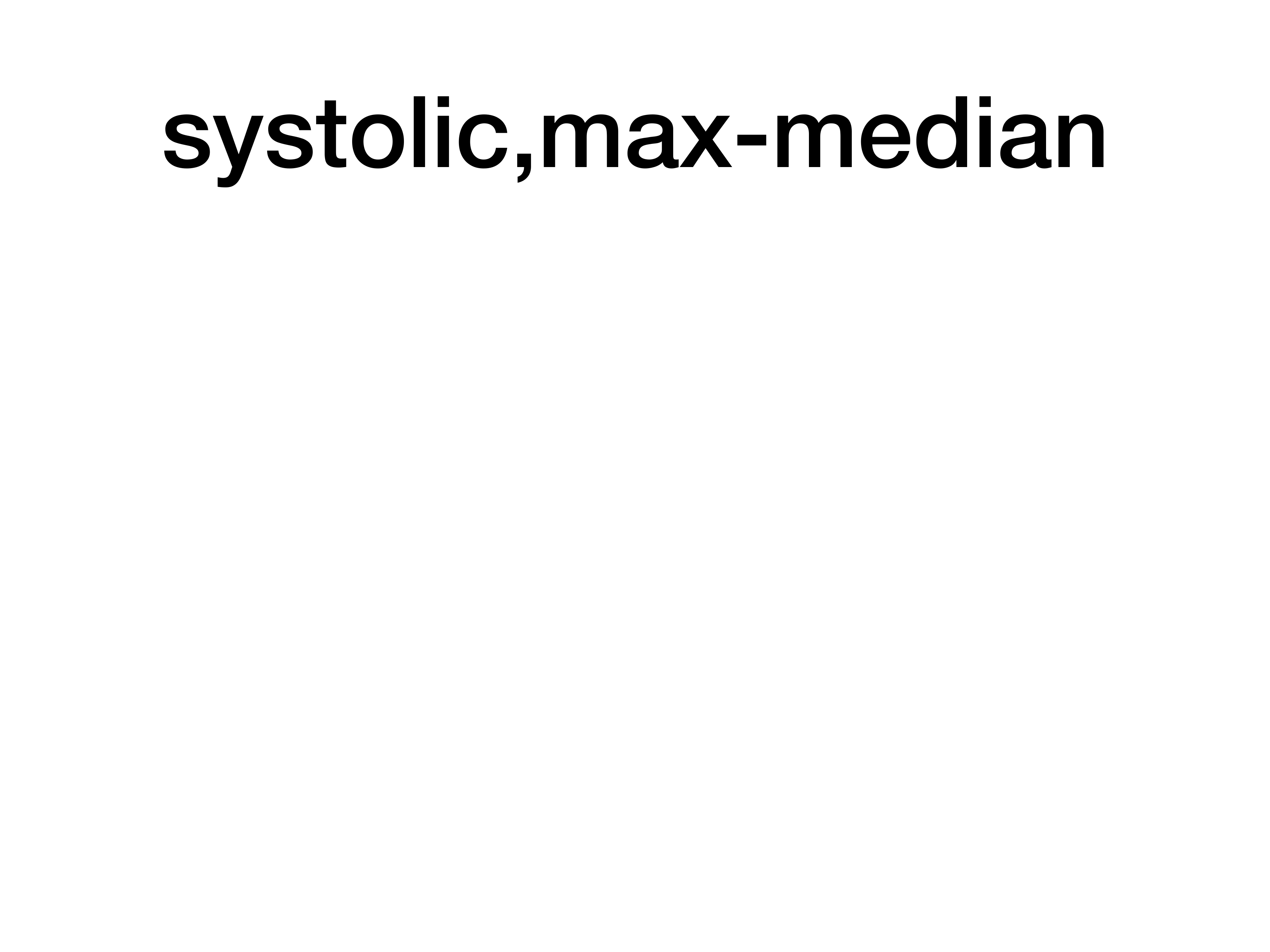}}
\subfigure{%
\label{far3_4}%
\includegraphics[width=1.7cm]{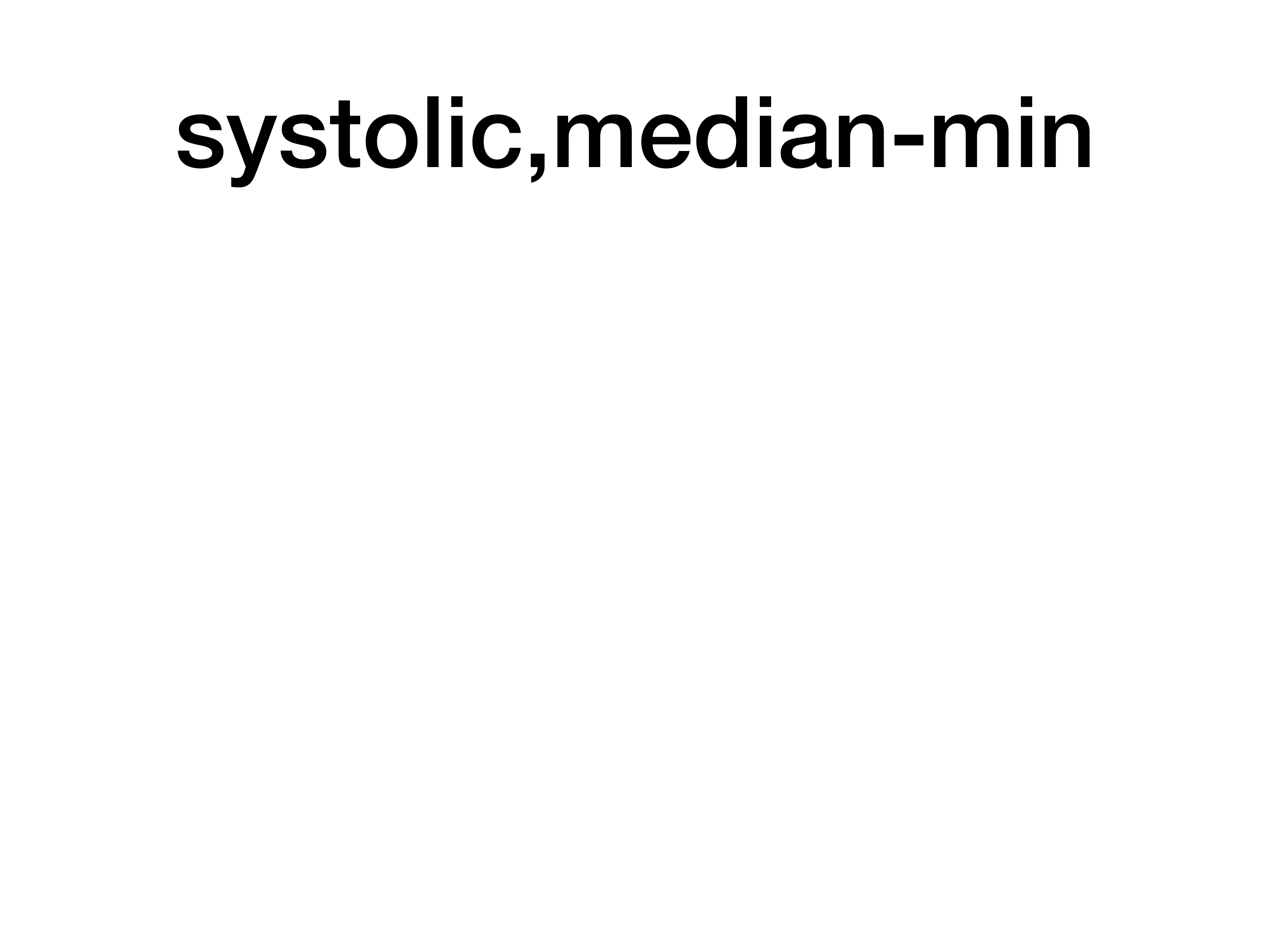}}
\subfigure{%
\label{far3_5}%
\includegraphics[width=1.7cm]{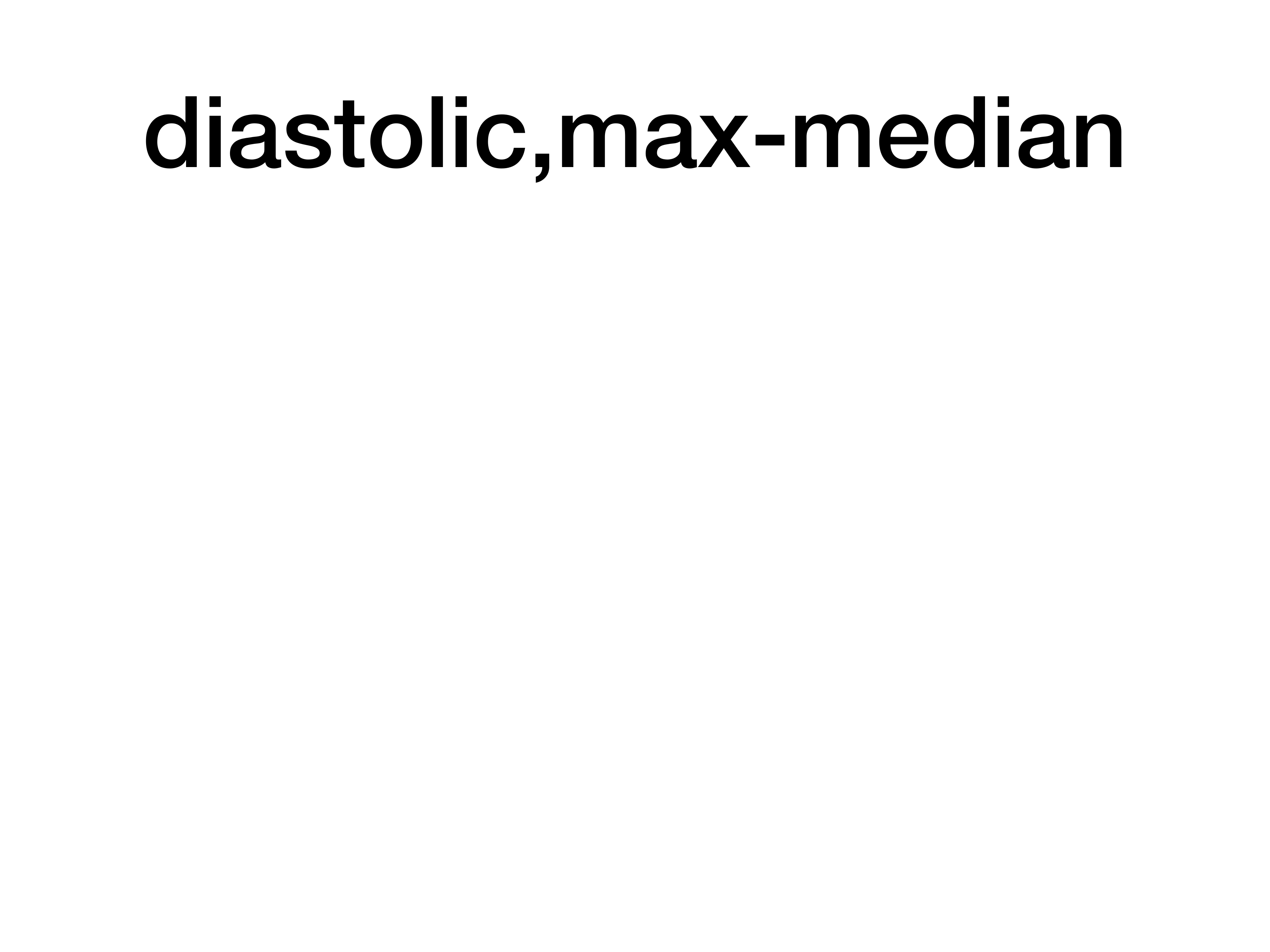}}
\subfigure{%
\label{far3_6}%
\includegraphics[width=1.7cm]{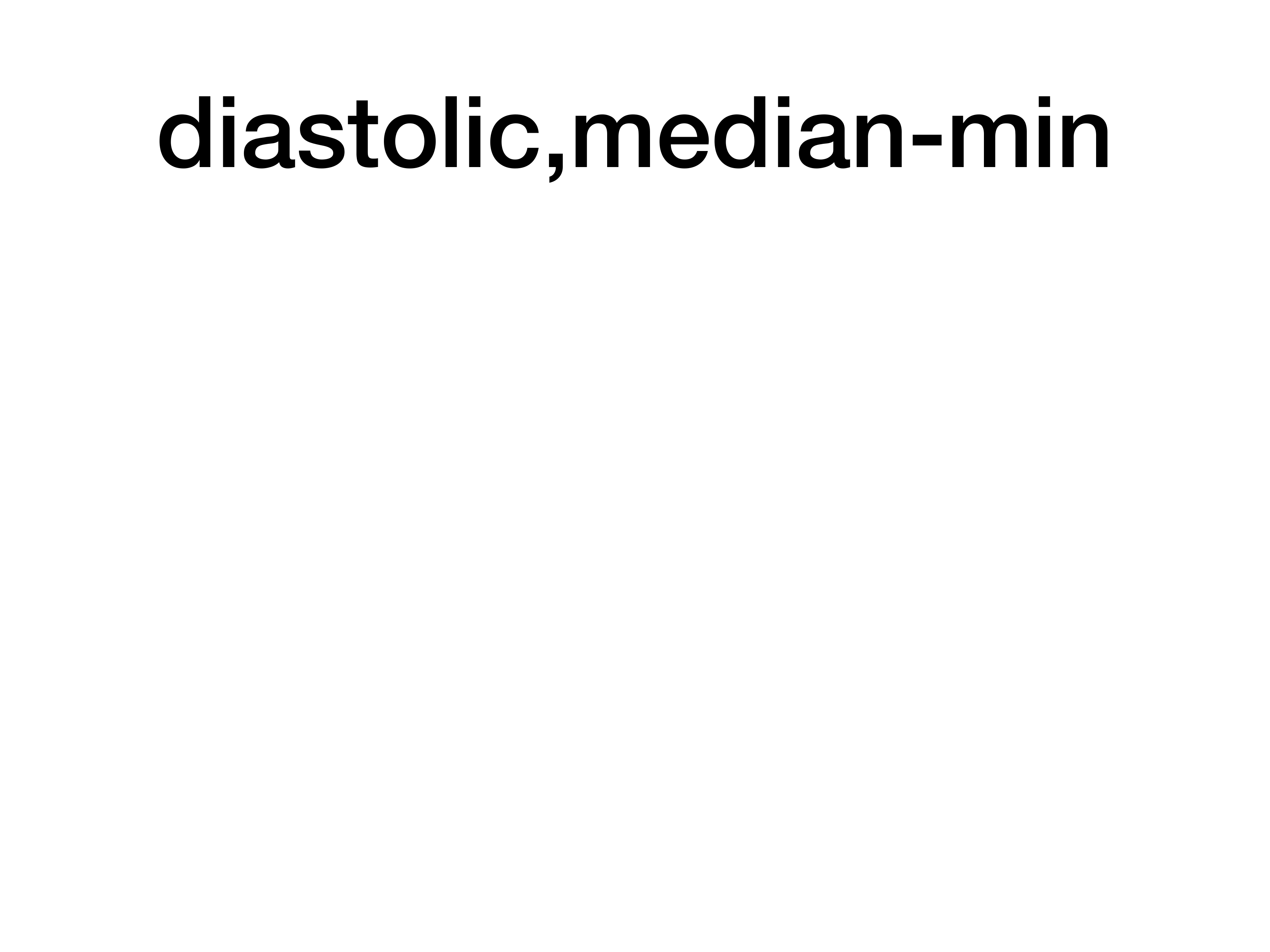}}
\subfigure{%
\label{far3_8}%
\includegraphics[width=1.7cm]{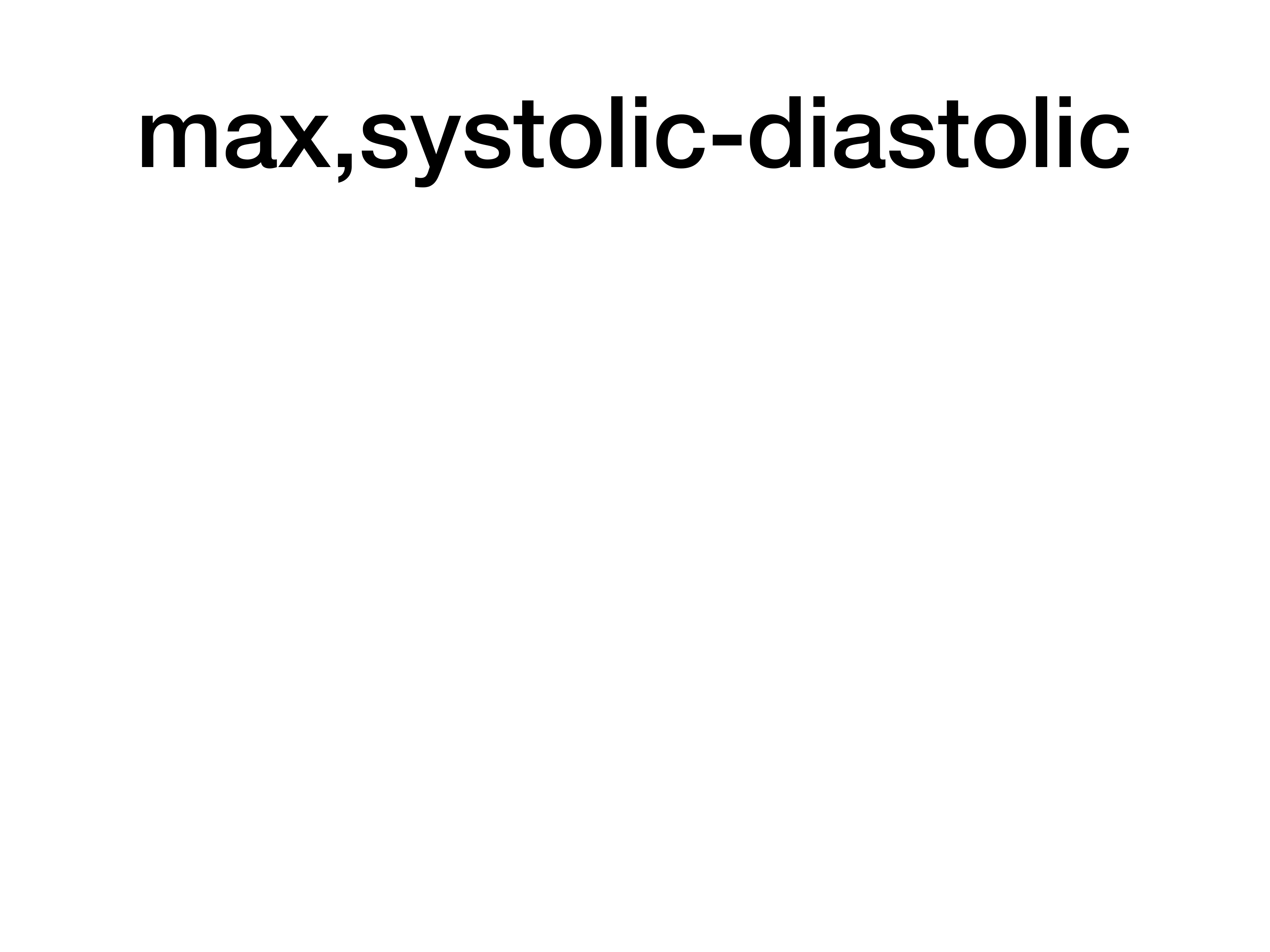}}
\subfigure{%
\label{far3_9}%
\includegraphics[width=1.7cm]{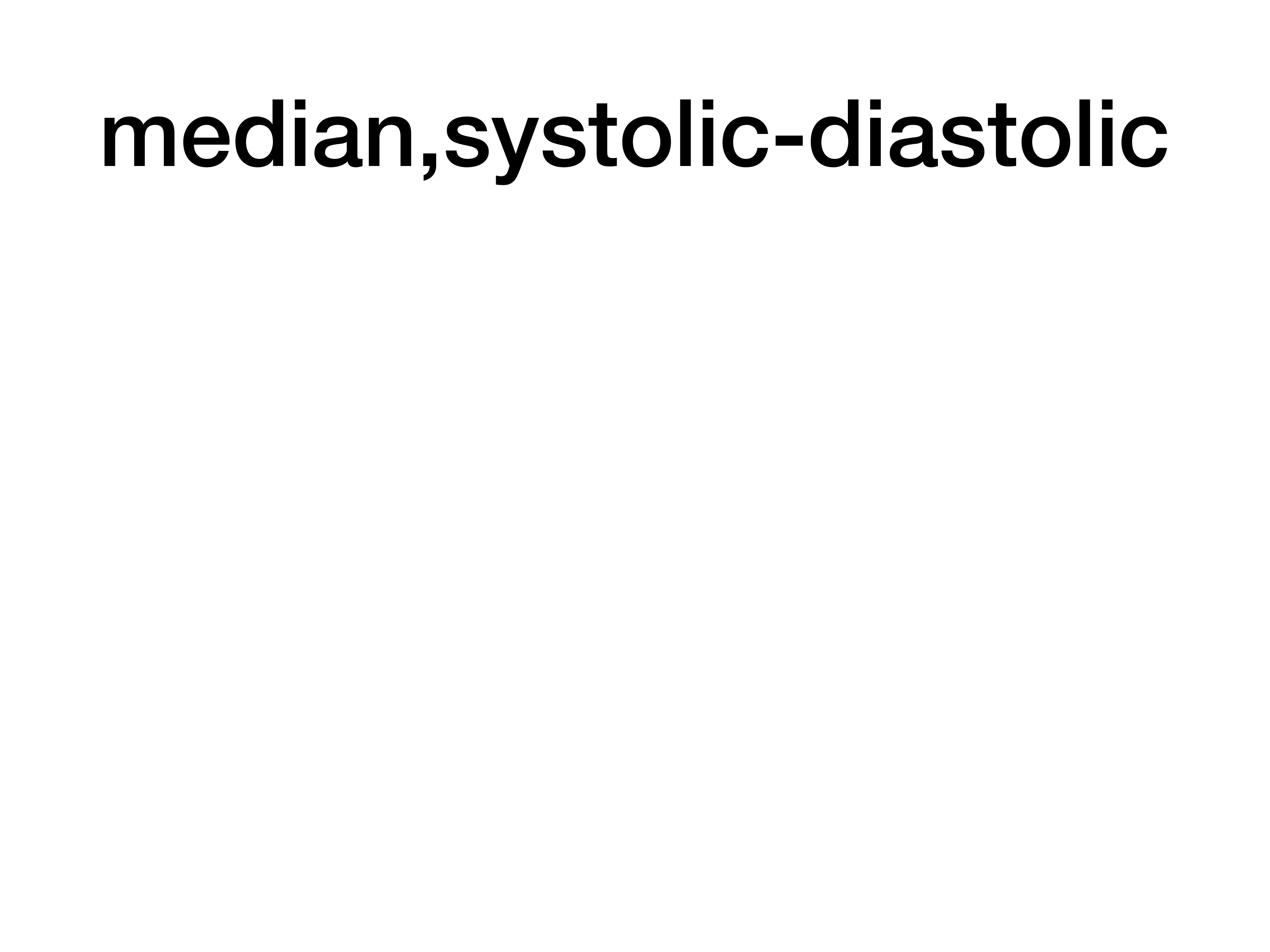}}
\subfigure{%
\label{far3_7}%
\includegraphics[width=1.7cm]{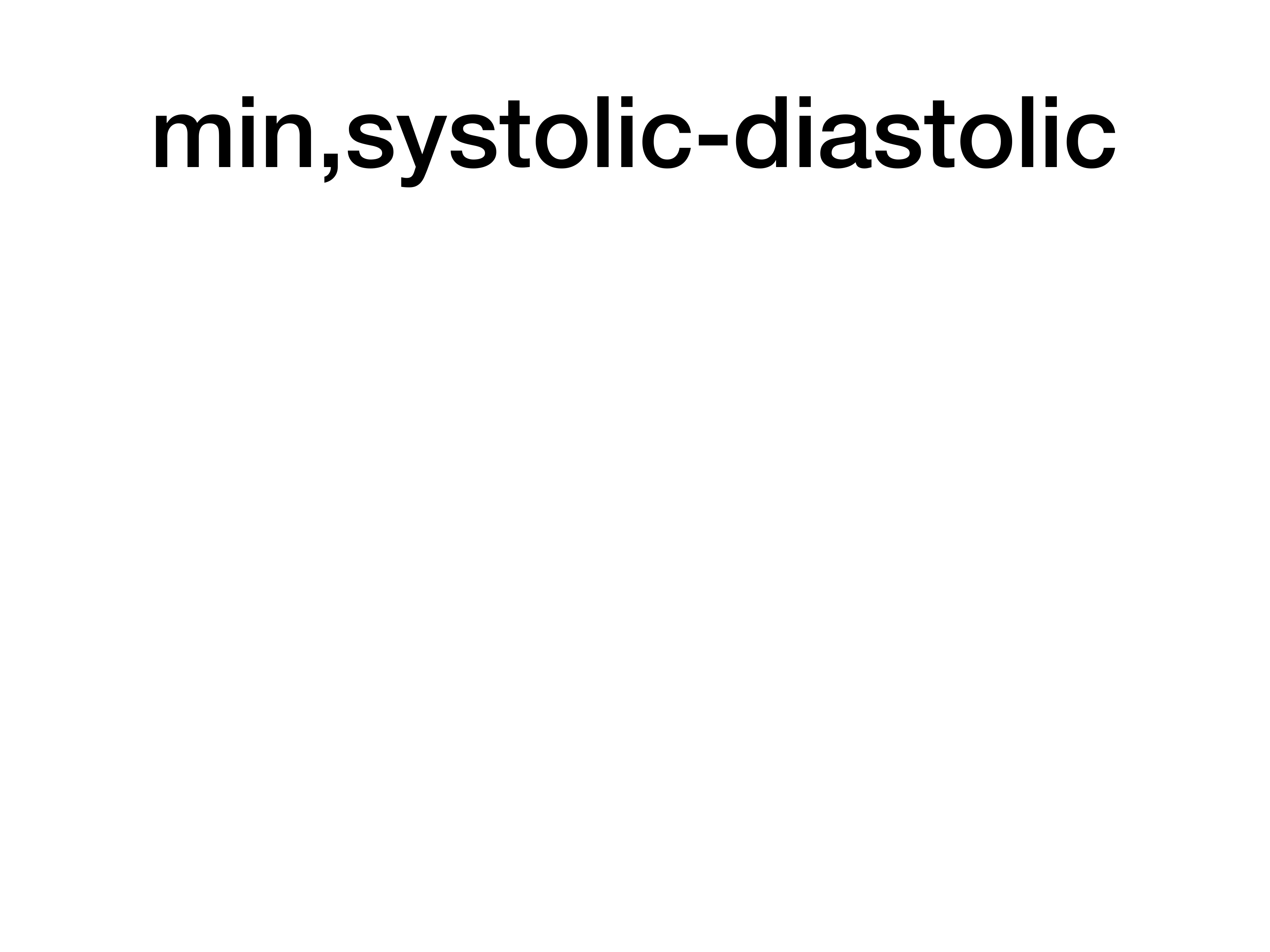}}
\vspace*{-20mm}
\caption{Constraint violation test. \emph{Without penalty} corresponds to the model HGAN without constraint violation penalty. \emph{With penalty} corresponds to HGAN. Distributions marked with \emph{Real} are built from real EHR data.}\label{cvt}
\end{figure}

\textbf{Constraint Violation Test (CVT).}
To evaluate the effectiveness of the constraint violation penalty, we consider the following scenarios. First, for each vital sign (BMI, systolic, and diastolic pressure), we compute the record-level difference (\emph{max}-\emph{median}) and (\emph{median}-\emph{min}). Second, for each of the basic statistics (\emph{max}, \emph{median}, and \emph{min}), we construct the distribution of the difference in record-level systolic and diastolic pressure. Third, we adopt two baselines to compare our model with. The first baseline is the real data, while the second is HGAN without the constraint violation penalty. Figure \ref{cvt} shows the CVT results, where the first row is the comparison of real \emph{vs} HGAN and the second row is real \emph{vs} HGAN without the constraint violation penalty. As can be seen in the first six columns, all of the difference distributions built from HGAN without a penalty have negative values, which suggests violations of the constraints. By contrast, HGAN always ensured the difference was positive. From the final three columns, it can be seen that HGAN (with the penalty) yield a more similar distribution with the real data than HGAN without the penalty. This indicates that HGAN is able to solve the feature constraint violation problem in our generation settings.

\begin{figure}[ht]%
\centering
\subfigure[Real,Precision]{%
\label{far_1}%
\includegraphics[width=4.0cm]{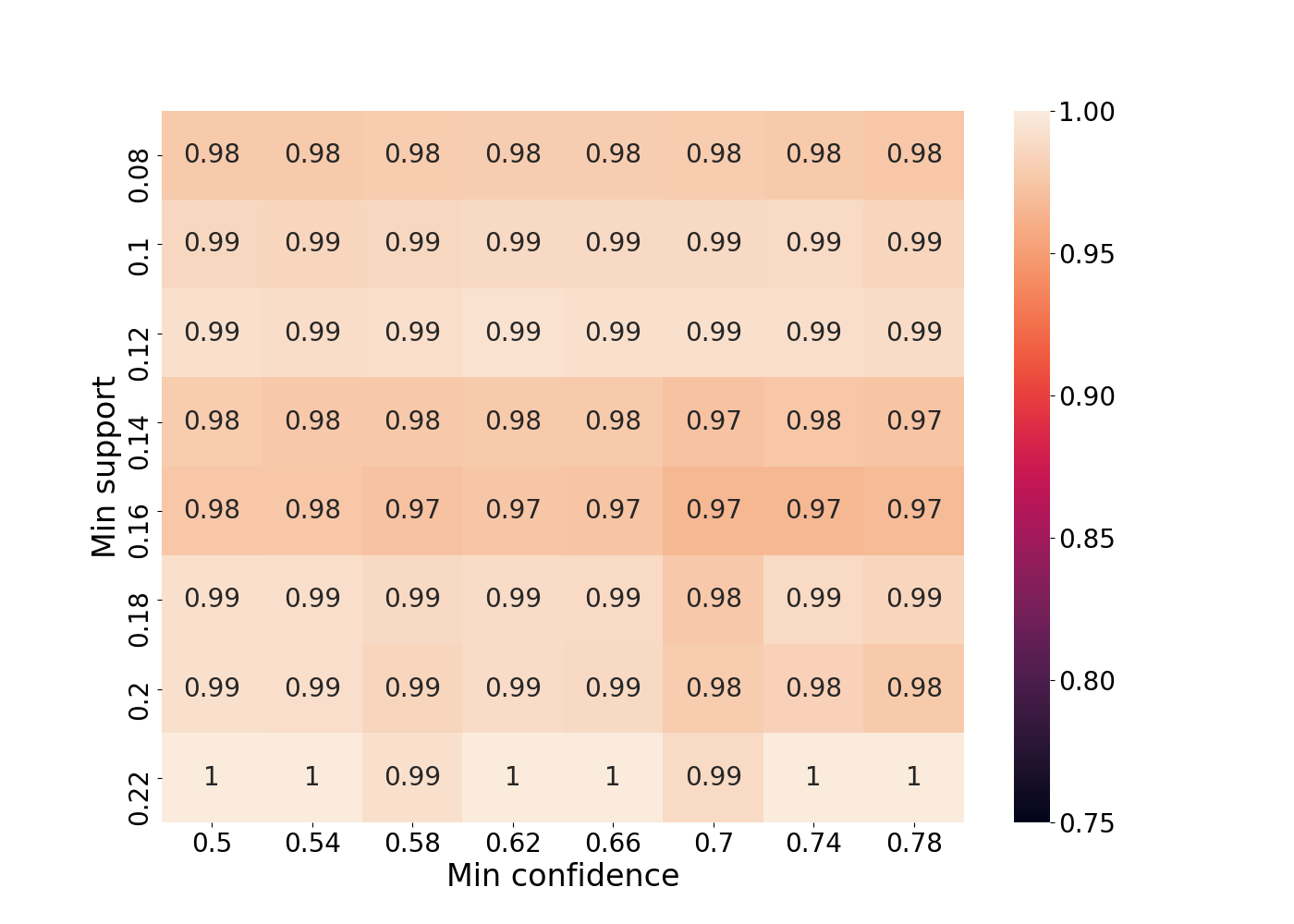}}
\subfigure[EMR-CWGAN,Precision]{%
\label{far_2}%
\includegraphics[width=4.0cm]{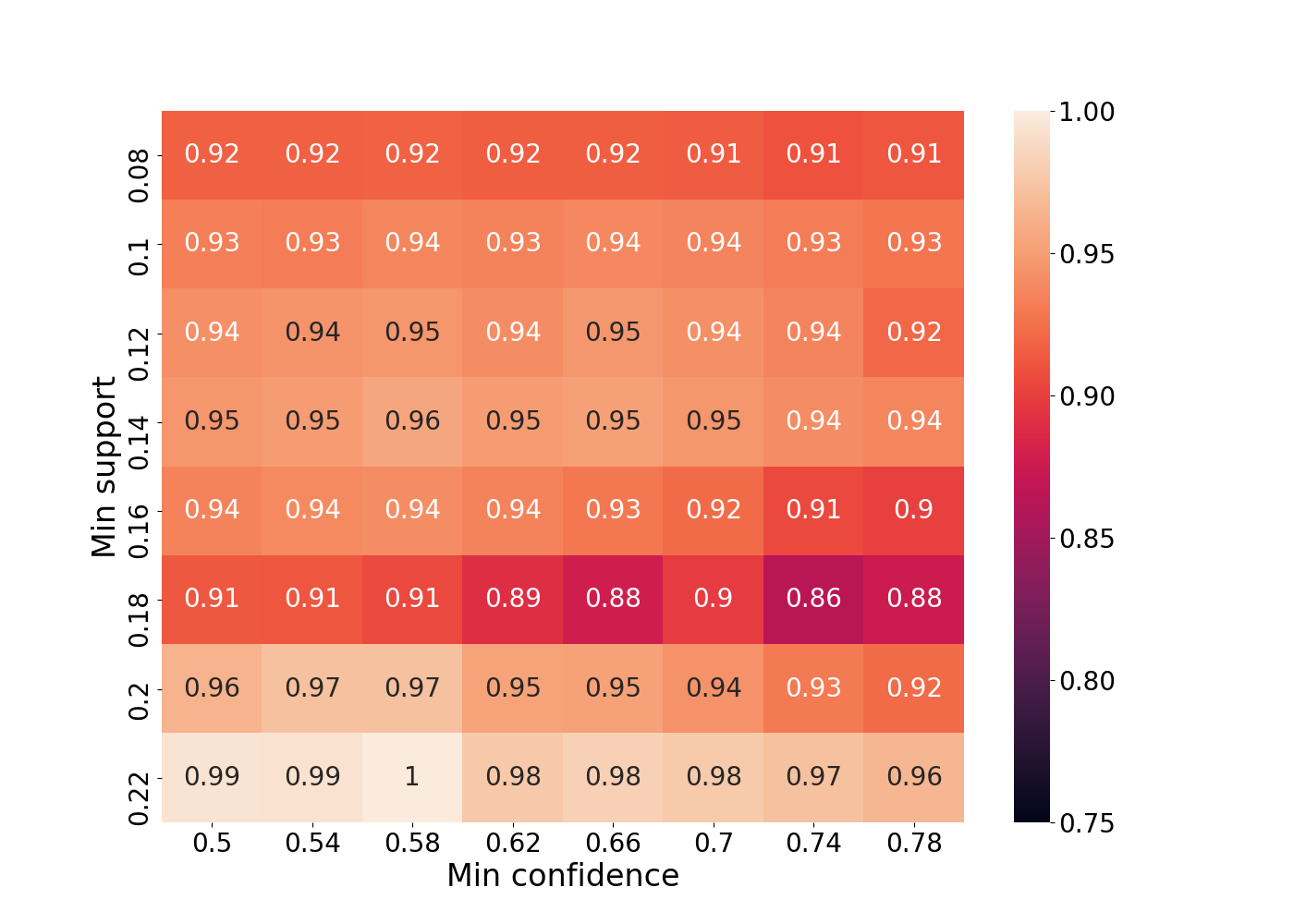}}
\subfigure[HGAN-U,Precision]{%
\label{far_3}%
\includegraphics[width=4.0cm]{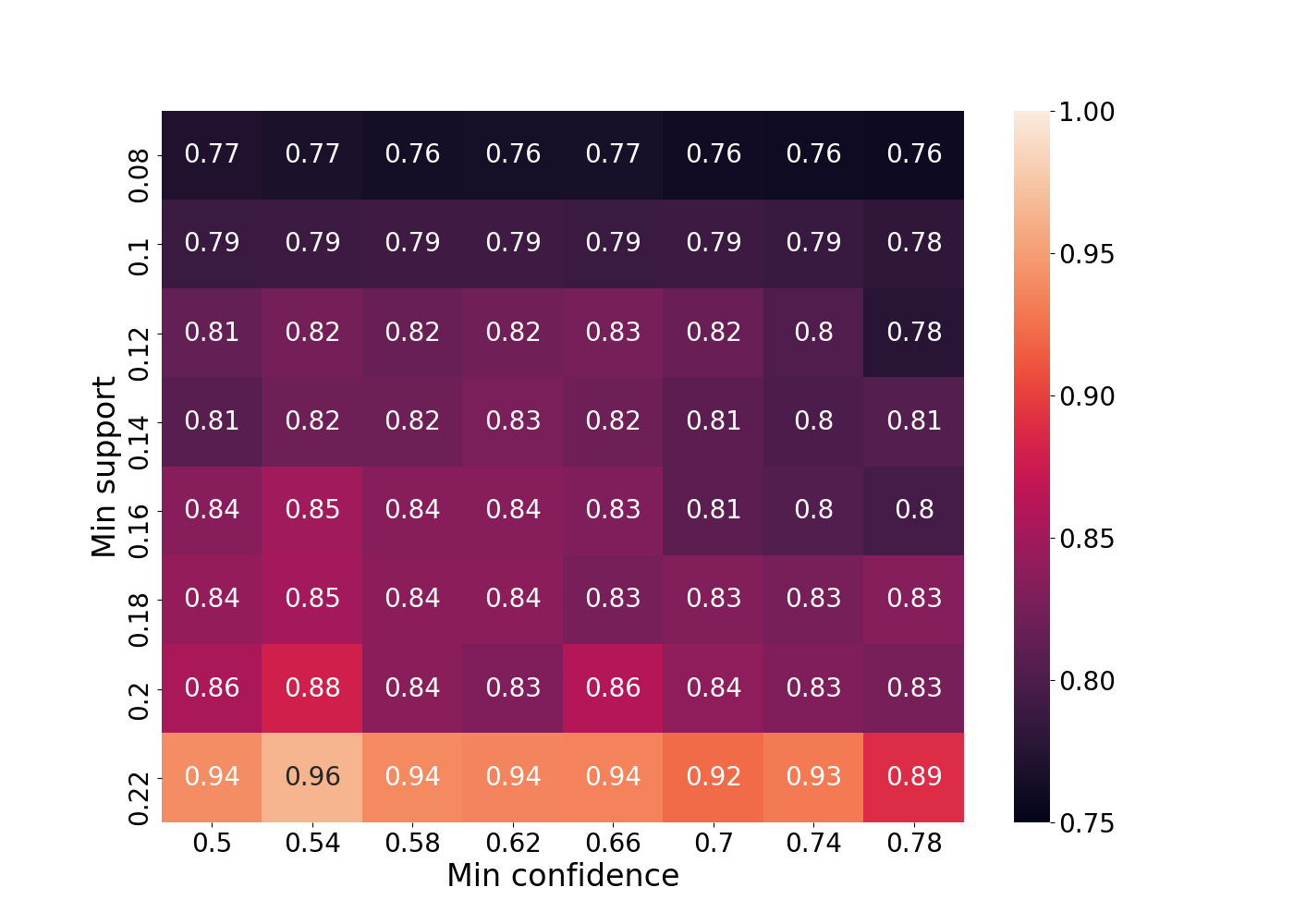}}
\subfigure[HGAN,Precision]{%
\label{far_4}%
\includegraphics[width=4.0cm]{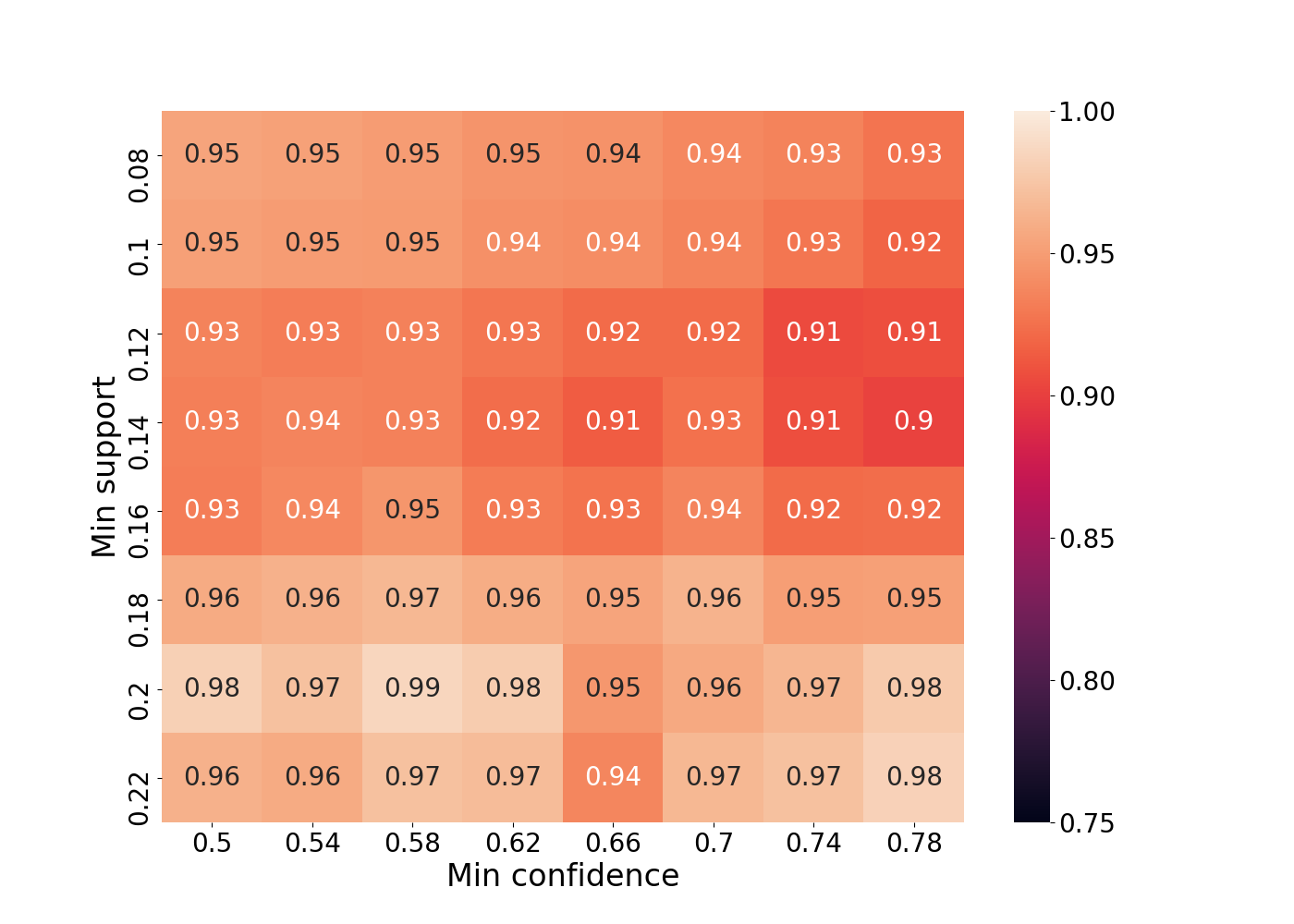}}%
\vspace*{-3mm}

\subfigure[Real,Recall]{%
\label{far_5}%
\includegraphics[width=4.0cm]{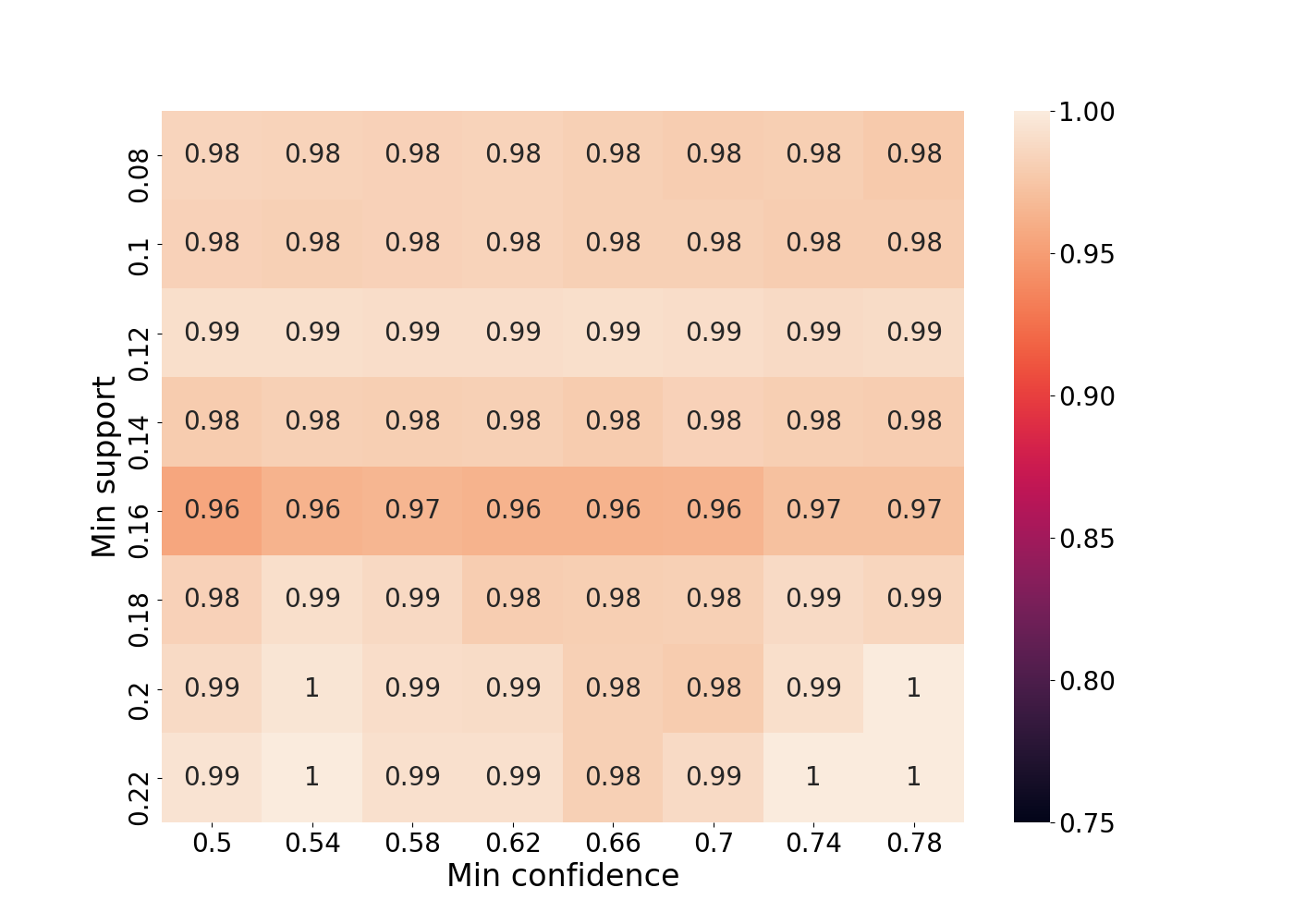}}
\subfigure[EMR-CWGAN,Recall]{%
\label{far_6}%
\includegraphics[width=4.0cm]{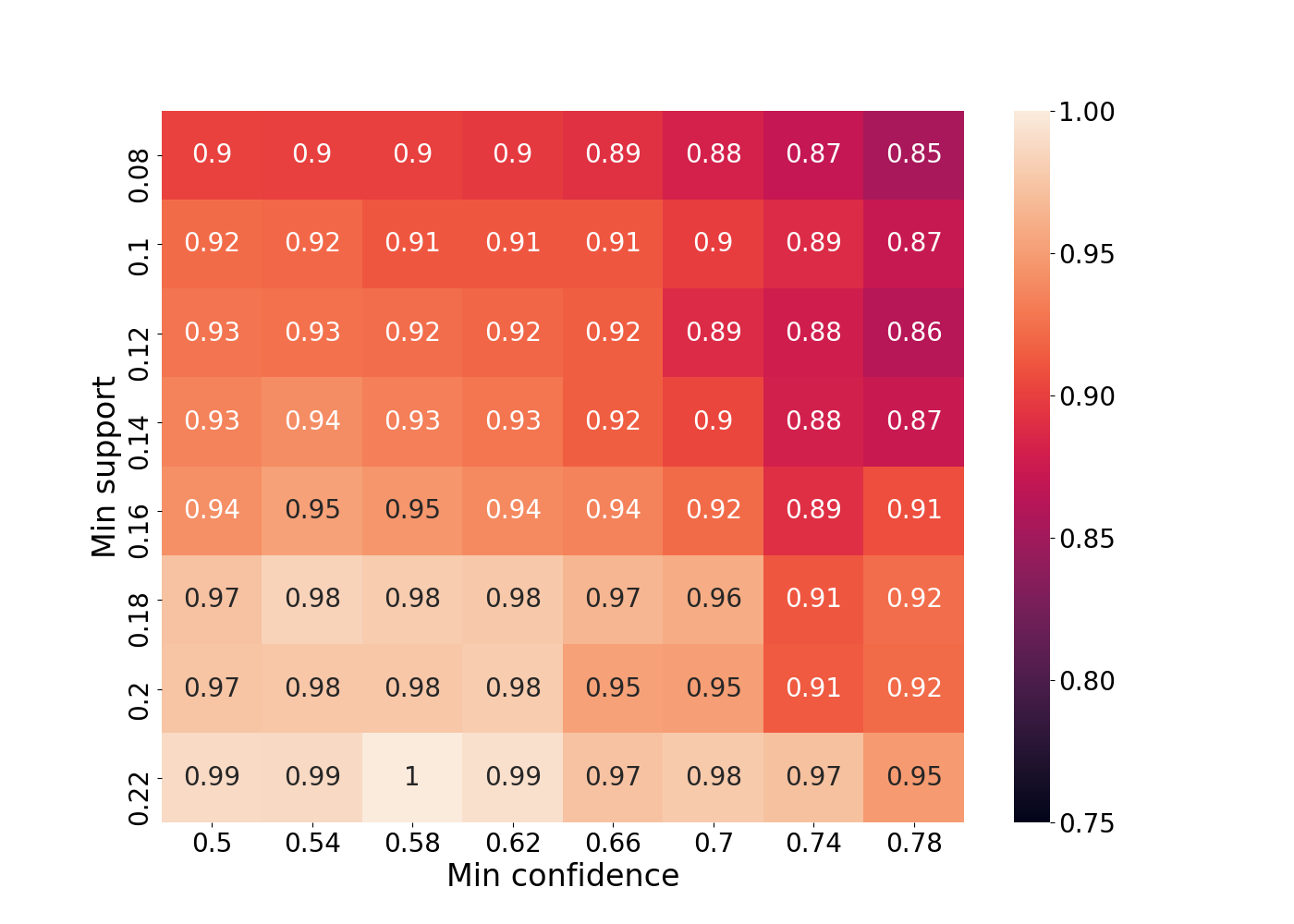}}
\subfigure[HGAN-U,Recall]{%
\label{far_7}%
\includegraphics[width=4.0cm]{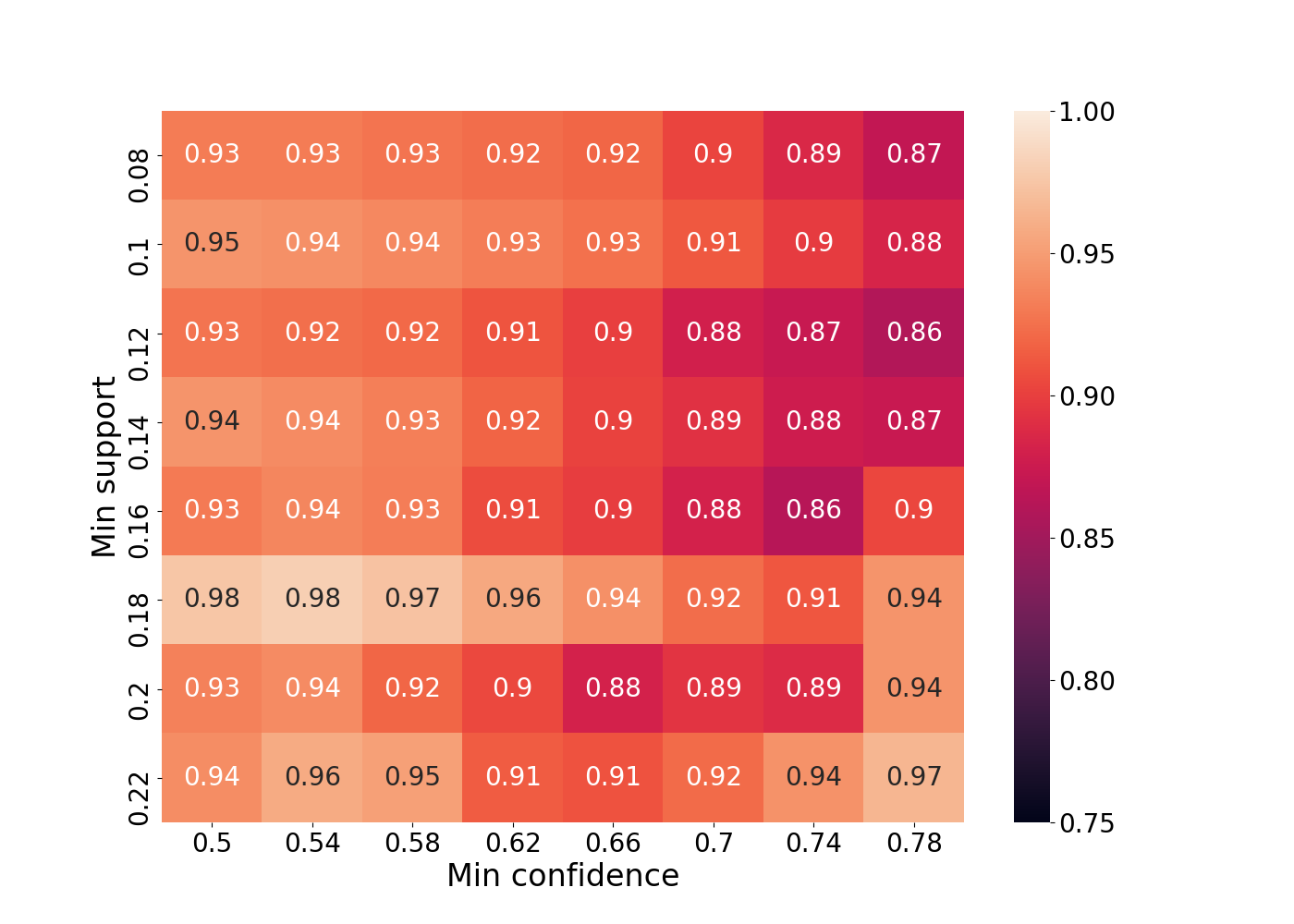}}
\subfigure[HGAN,Recall]{%
\label{far_8}%
\includegraphics[width=4.0cm]{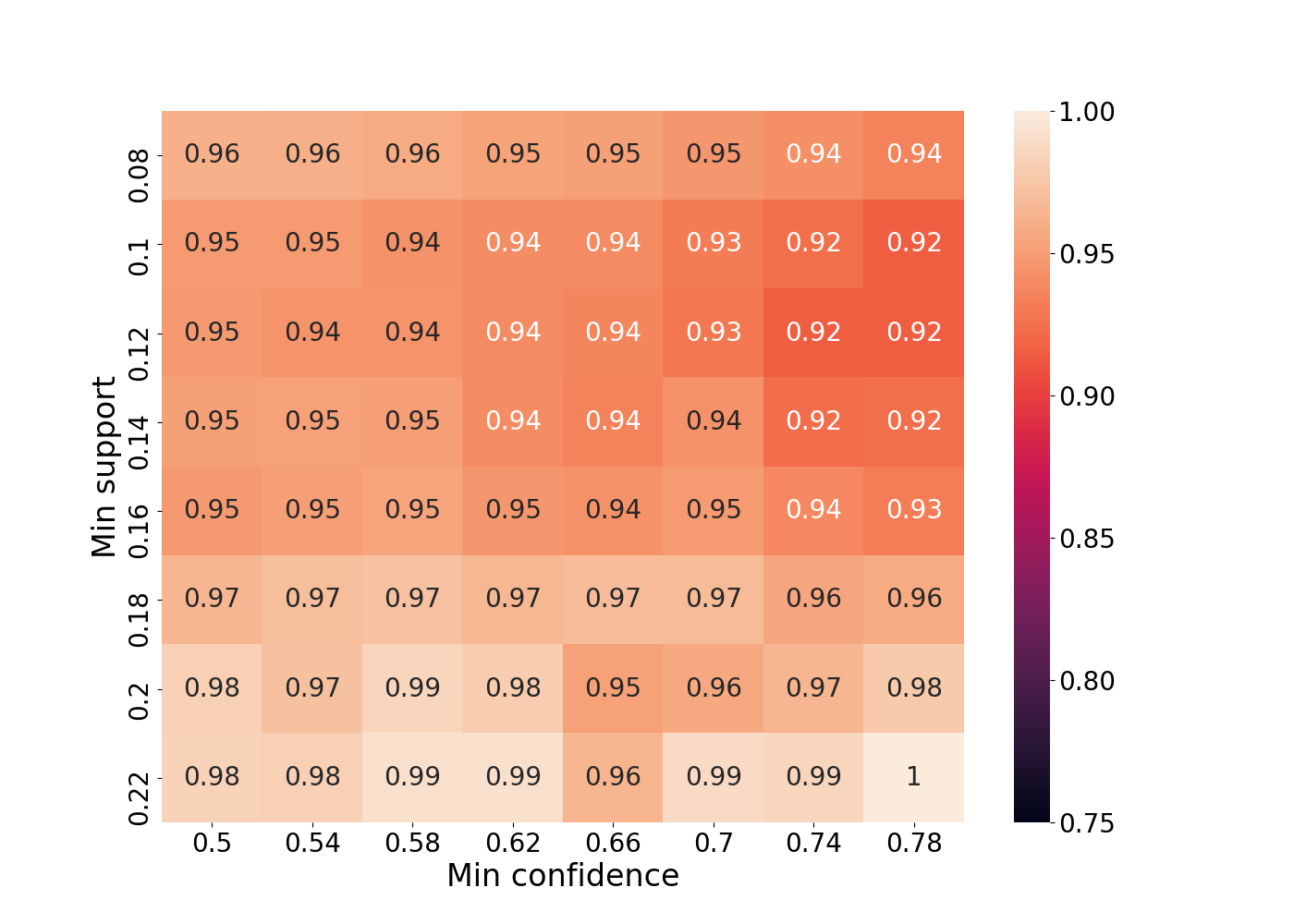}}%
\vspace*{-10mm}
\caption{Frequent association rules. The cell numbers in subfigures \ref{far_1}-\ref{far_4} and \ref{far_5}-\ref{far_8} show the precision and recall of the association rules (for ICD and CPT codes), respectively. Figures \ref{far_1} and \ref{far_5} correspond to the real \emph{vs} real setting.}\label{far}
\end{figure}

\textbf{Frequent Association Rules (FAR).}
Figure \ref{far} shows the results for FAR. Given a range of thresholds for support ($min_s \in [0.08,0.22]$) and confidence ($min_c \in [0.50,0.78]$),  we measure both the precision (Figures \ref{far_1}--\ref{far_4}) and recall (Figures \ref{far_5}--\ref{far_8}) of association rules (learned from ICD and CPT codes).  
There are multiple observations to highlight. 
First, as can be seen from Figures \ref{far_1} and \ref{far_5}, both the precision and recall from two equal-sized subsets of real EHR data are close to 1. This indicates that the association rules in the real data are robust. 
Second, by comparing Figure \ref{far_4} with \ref{far_3} and Figure \ref{far_8} with \ref{far_7}, it can be observed that HGAN almost always outperforms HGAN-U in terms of precision and recall.
Third, by comparing Figures \ref{far_8} with \ref{far_6}, it can be seen that the recall of HGAN is greater than EMR-CWGAN in the majority of conditions.
However, there is no obvious domination in precision when comparing Figures \ref{far_4} with \ref{far_2}.
These results suggest that HGAN is more capable at maintaining the record-level feature association (or record-wise consistency) than other baselines.

\begin{figure}[ht]%
\captionsetup[subfigure]{justification=centering}
\centering
\subfigure[Age,Real]{%
\label{ccd_1}%
\includegraphics[width=2.5cm]{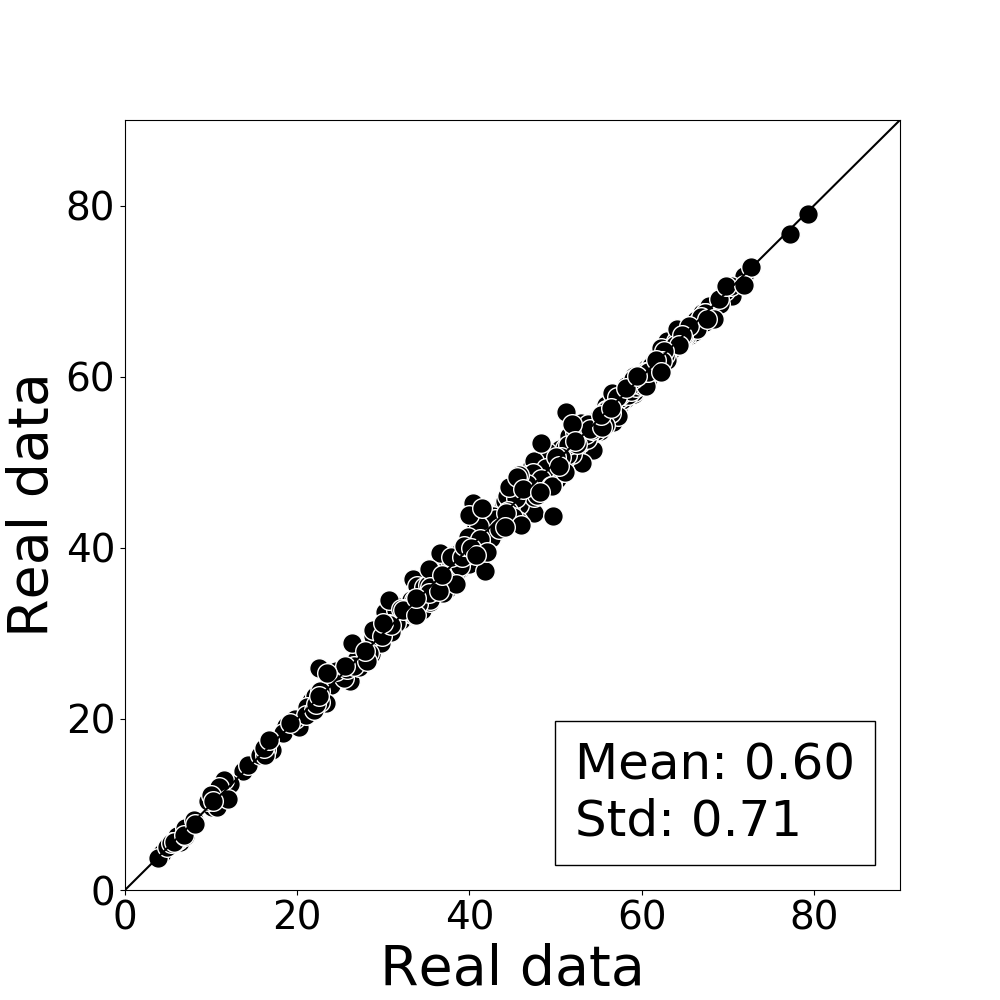}} 
\subfigure[Age,HGAN-U]{%
\label{ccd_2}%
\includegraphics[width=2.5cm]{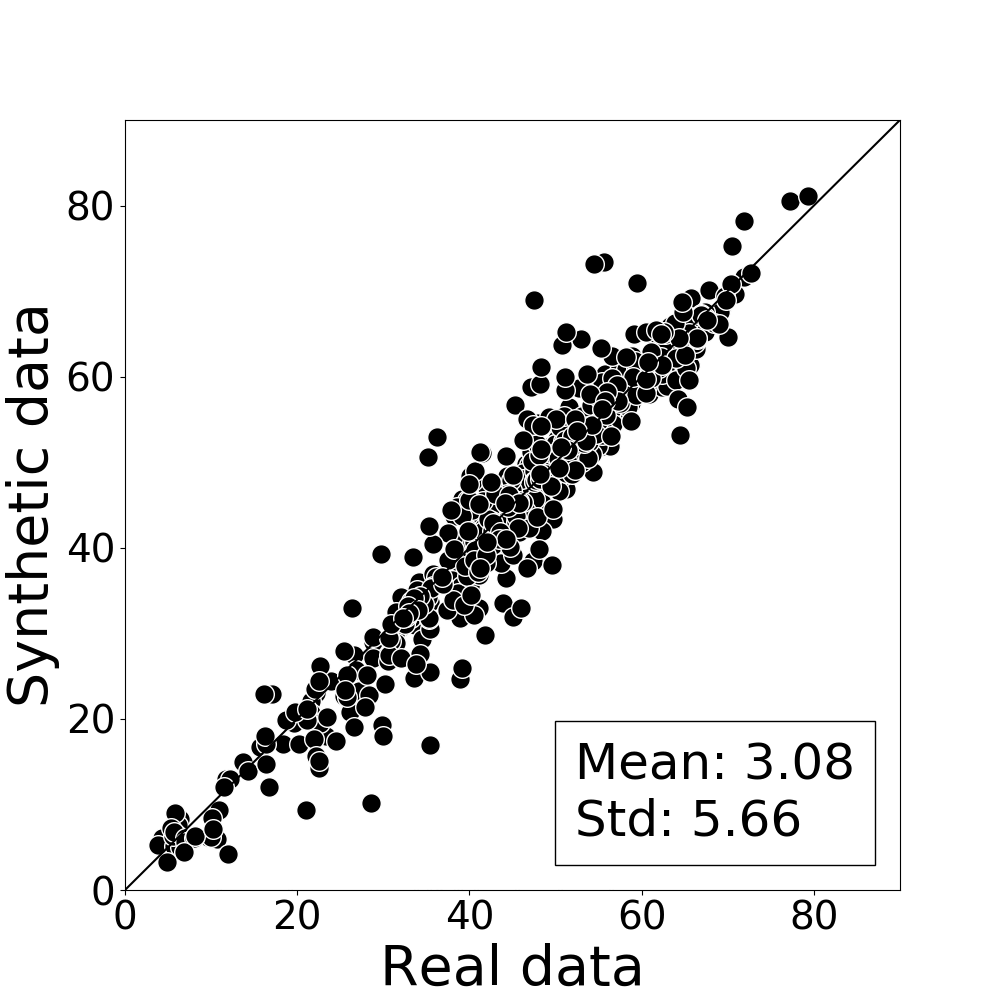}} 
\subfigure[Age,HGAN]{%
\label{ccd_3}%
\includegraphics[width=2.5cm]{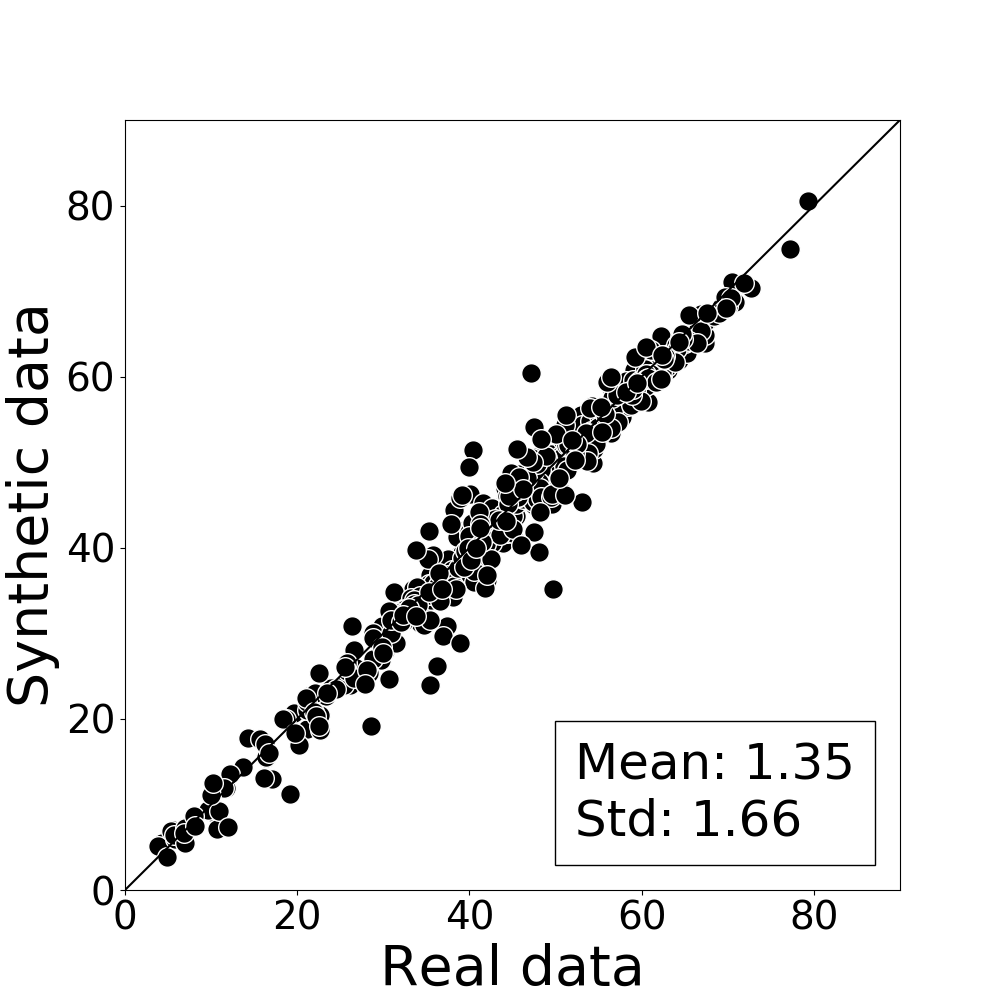}} 
\subfigure[Gender,Real]{%
\label{ccd_4}%
\includegraphics[width=2.5cm]{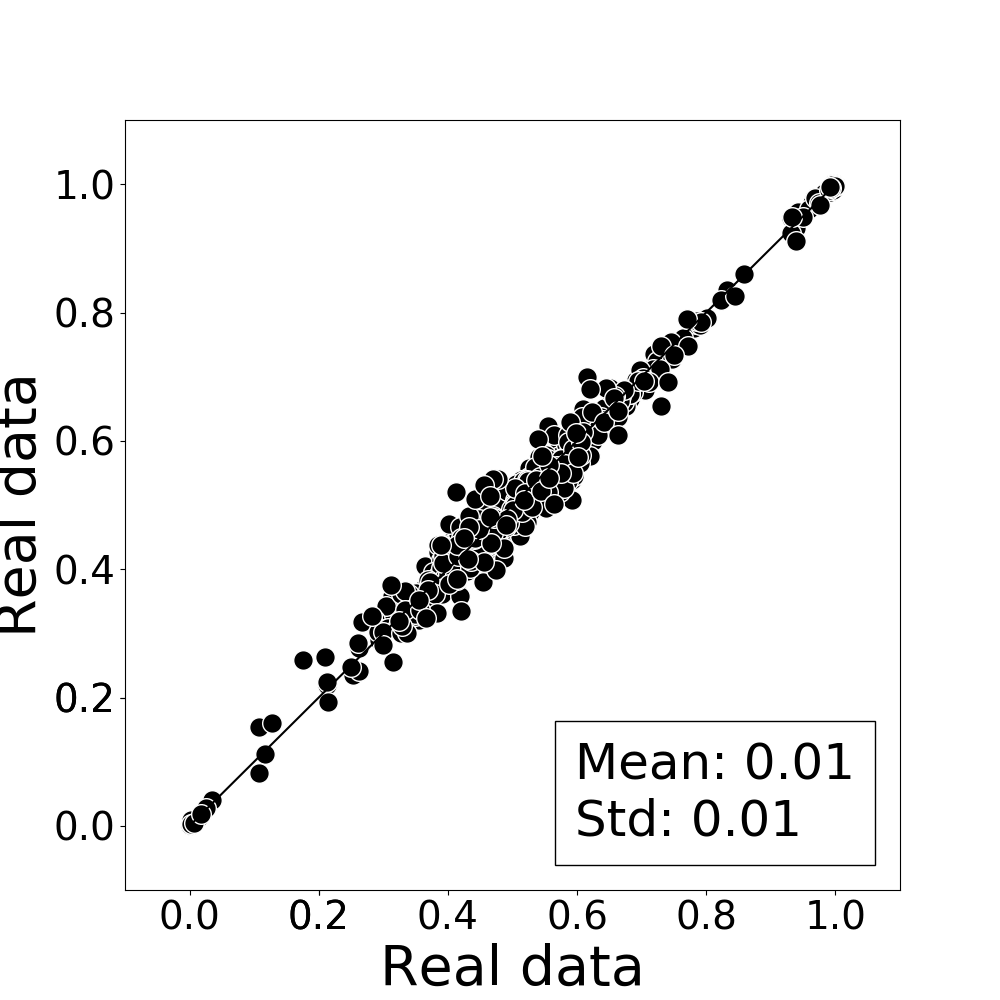}} 
\subfigure[Gender,HGAN-U]{%
\label{ccd_5}%
\includegraphics[width=2.5cm]{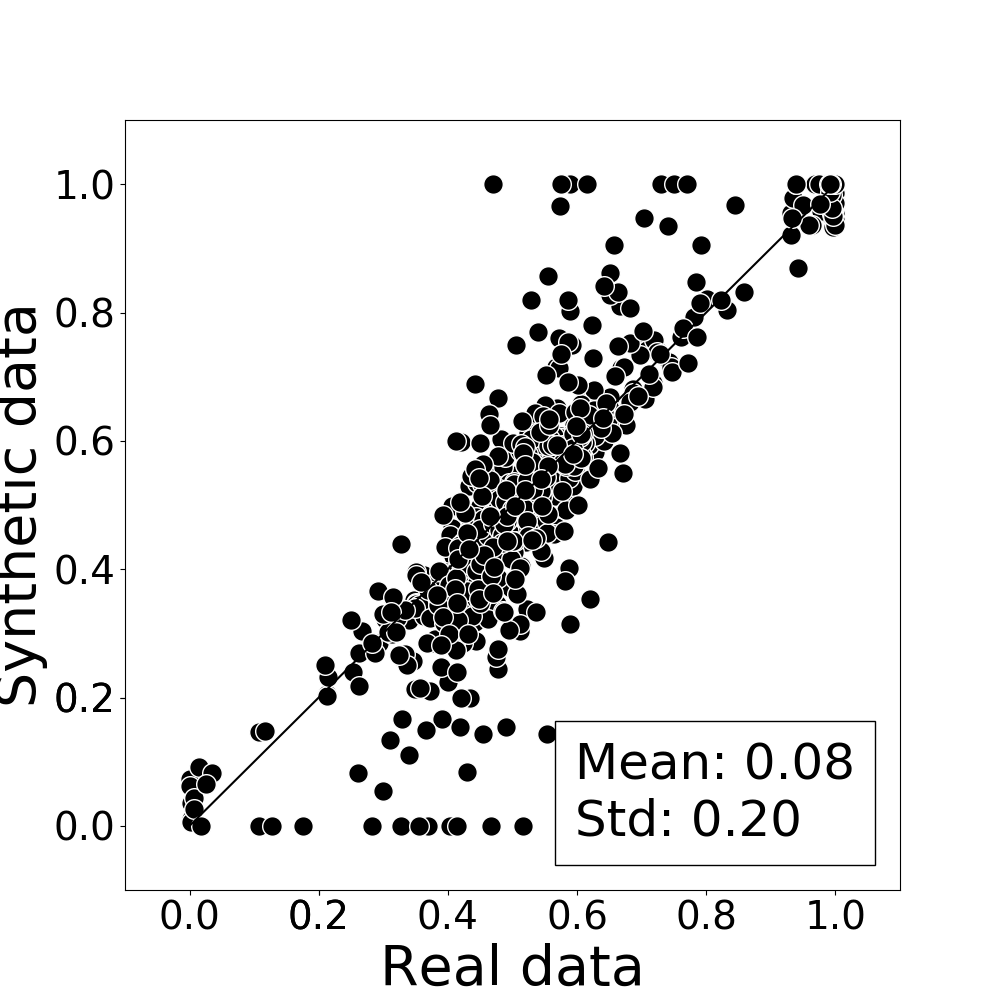}} 
\subfigure[Gender,HGAN]{%
\label{ccd_6}%
\includegraphics[width=2.5cm]{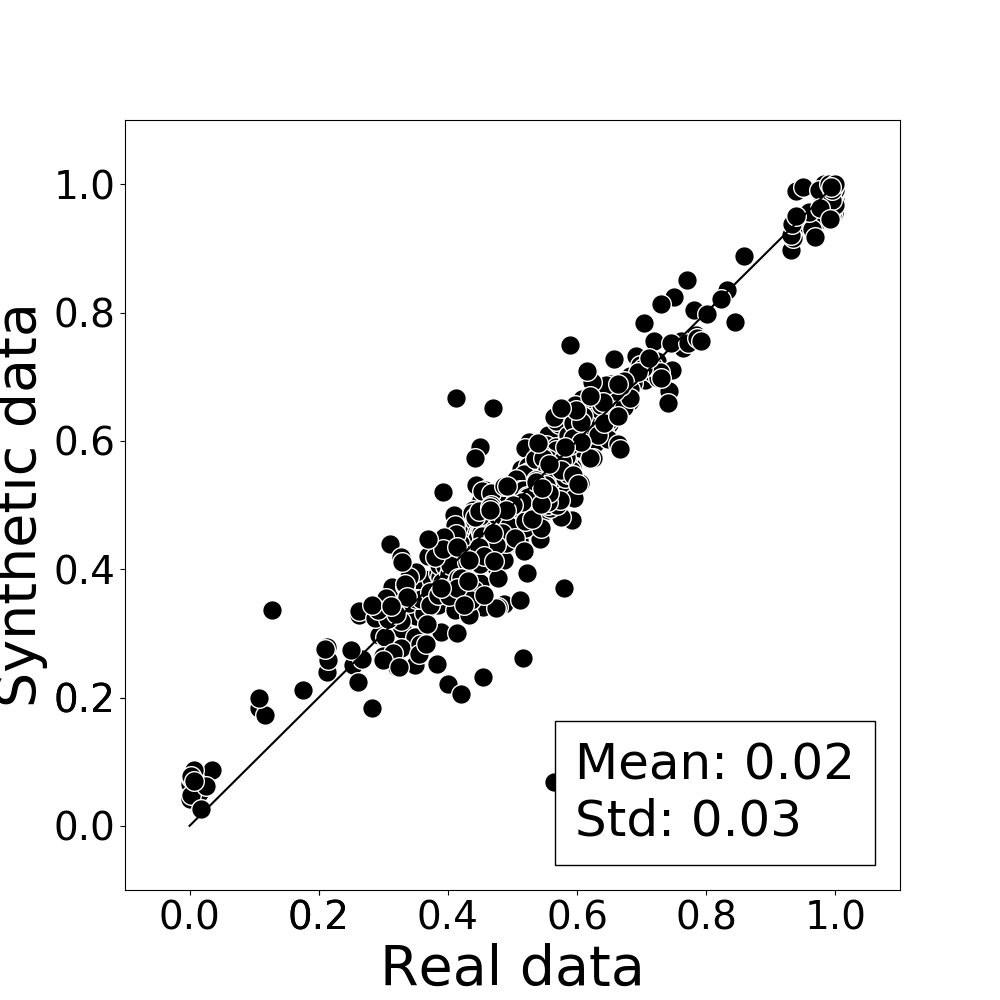}}
\vspace*{-10mm}
\caption{Cross-type conditional distribution on age and gender. Subfigures \ref{ccd_1}--\ref{ccd_3} show the mean age for each code in the real \emph{vs} real and real \emph{vs} synthetic settings. Subfigures \ref{ccd_4}--\ref{ccd_6} show the proportion of genders that are female for each code.}\label{cond_age_sex}
\end{figure}

\begin{figure}[ht]%
\captionsetup[subfigure]{justification=centering}
\centering
\subfigure[Age,Real]{%
\label{std_ccd_1}%
\includegraphics[width=2.5cm]{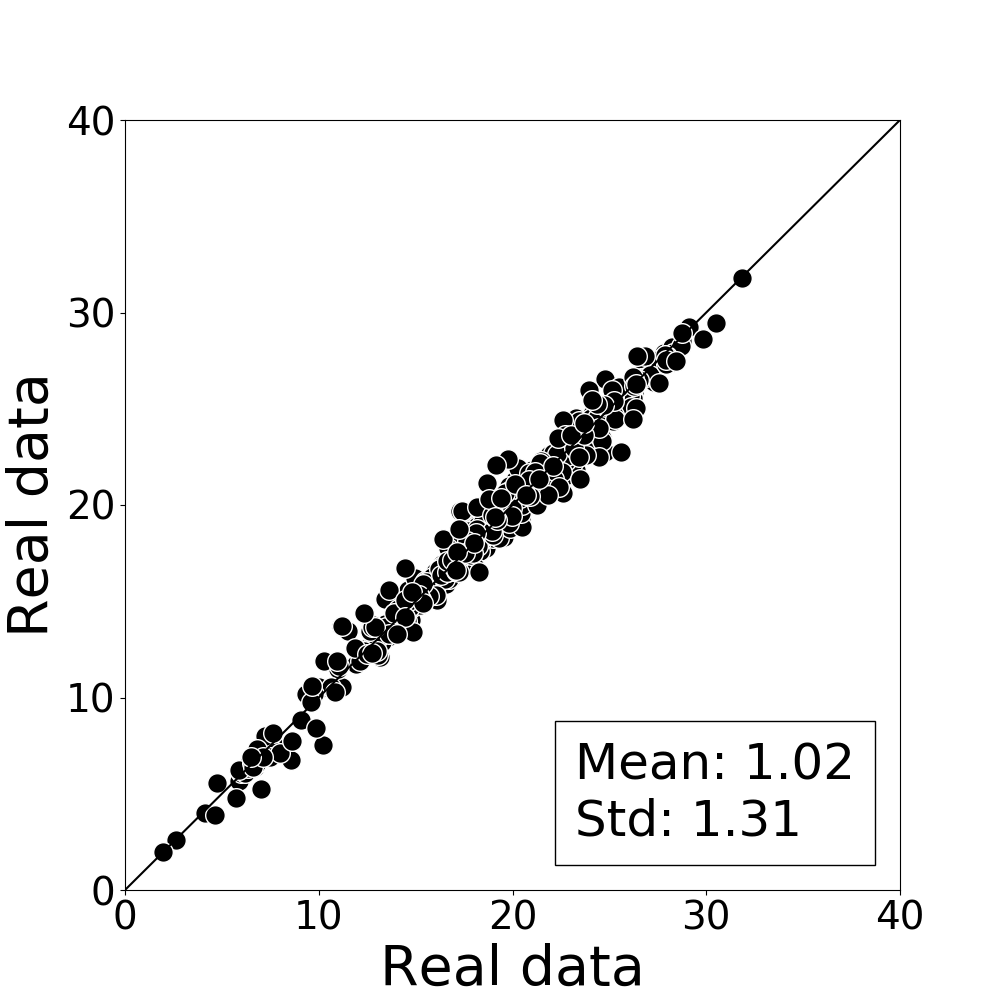}} 
\subfigure[Age,HGAN-U]{%
\label{std_ccd_2}%
\includegraphics[width=2.5cm]{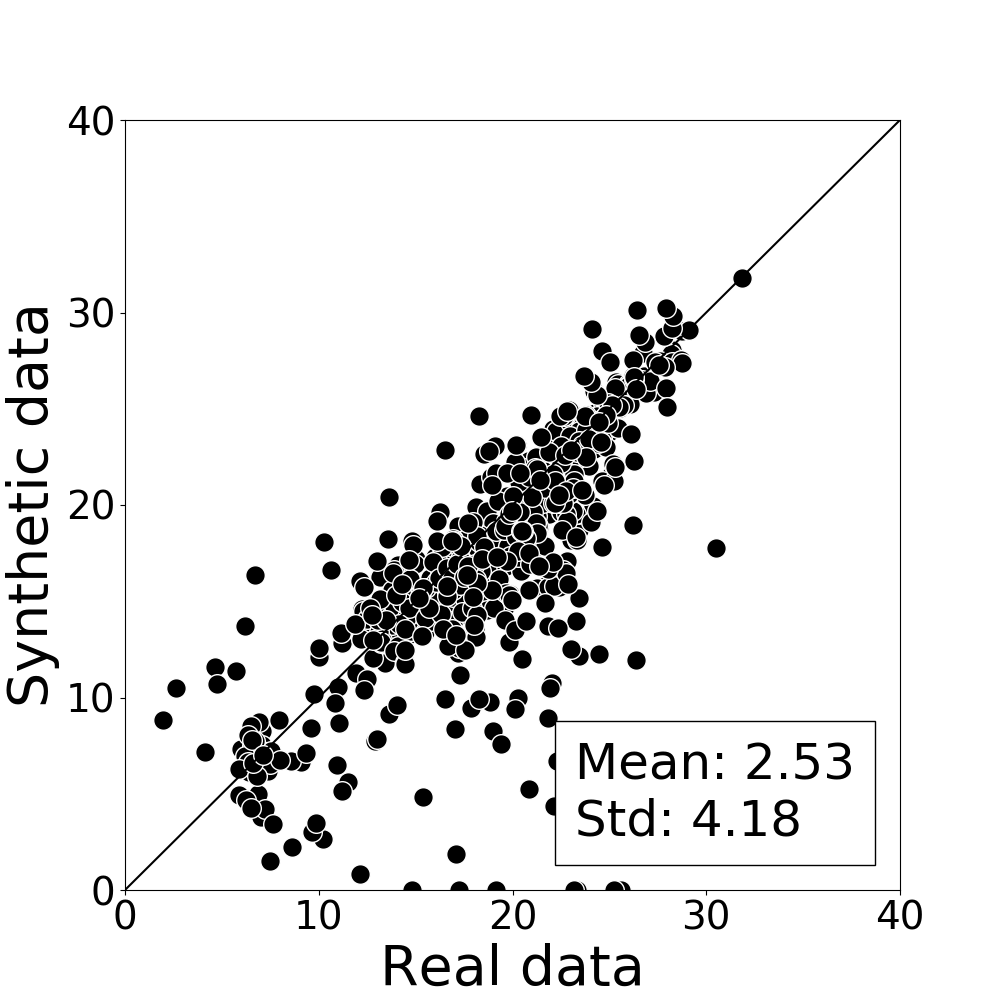}} 
\subfigure[Age,HGAN]{%
\label{std_ccd_3}%
\includegraphics[width=2.5cm]{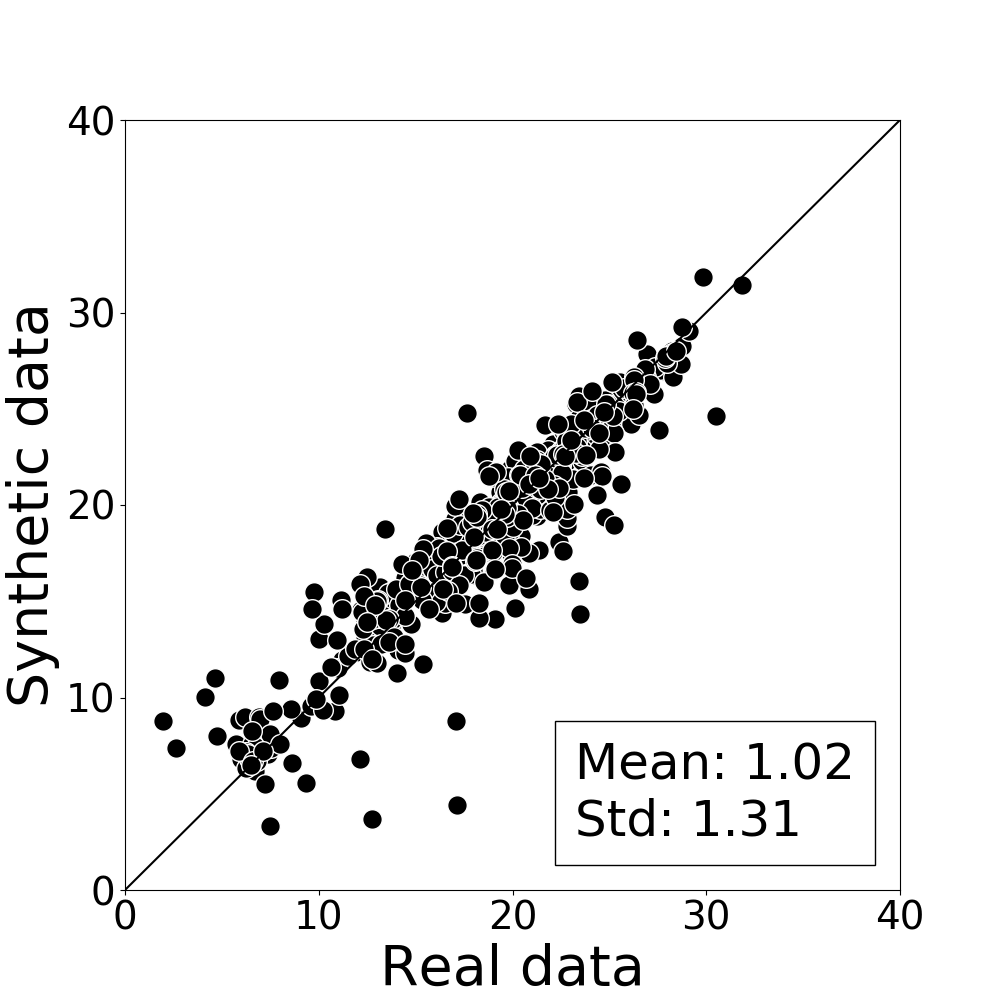}}
\vspace*{-10mm}
\caption{Cross-type conditional distribution on age and gender. Subfigures \ref{std_ccd_1}--\ref{std_ccd_3} show the standard deviation of age for each code in real \emph{vs} real and real \emph{vs} synthetic settings.}\label{cond_age_std}
\end{figure}

\begin{figure}[ht]%
\captionsetup[subfigure]{justification=centering}
\centering
\subfigure{%
\includegraphics[width=2.5cm]{./real.pdf}}
\subfigure{%
\label{ccd1_1}%
\includegraphics[width=2.cm]{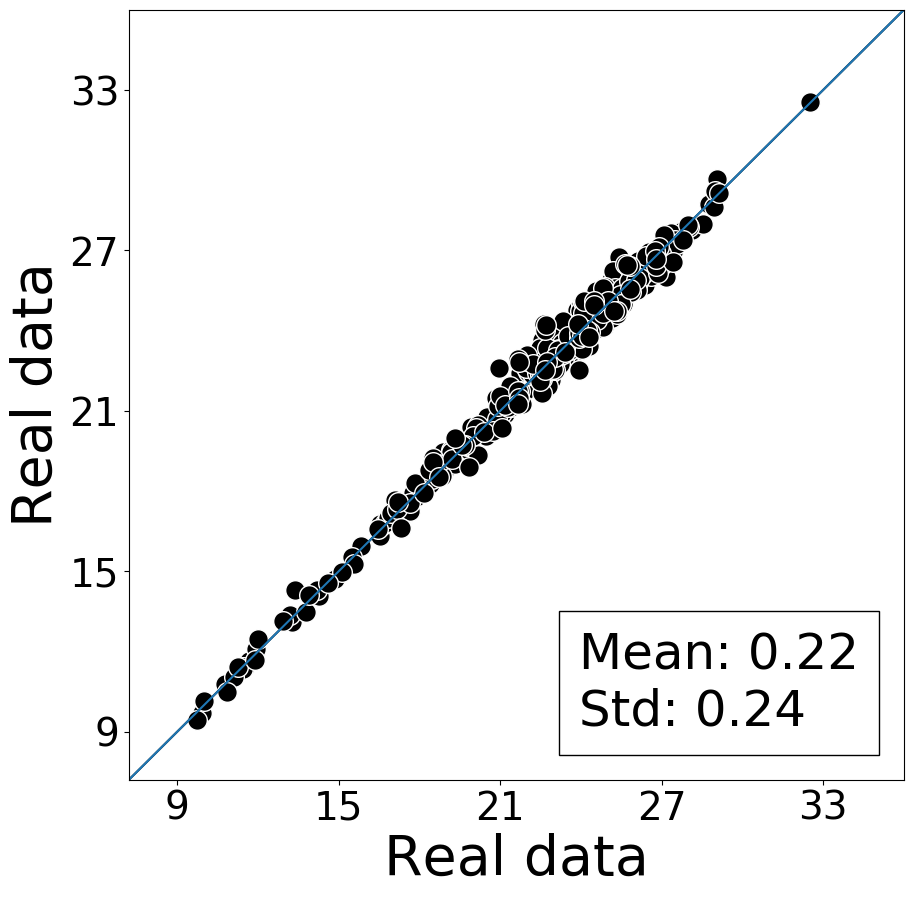}} 
\subfigure{%
\label{ccd1_3}%
\includegraphics[width=2.cm]{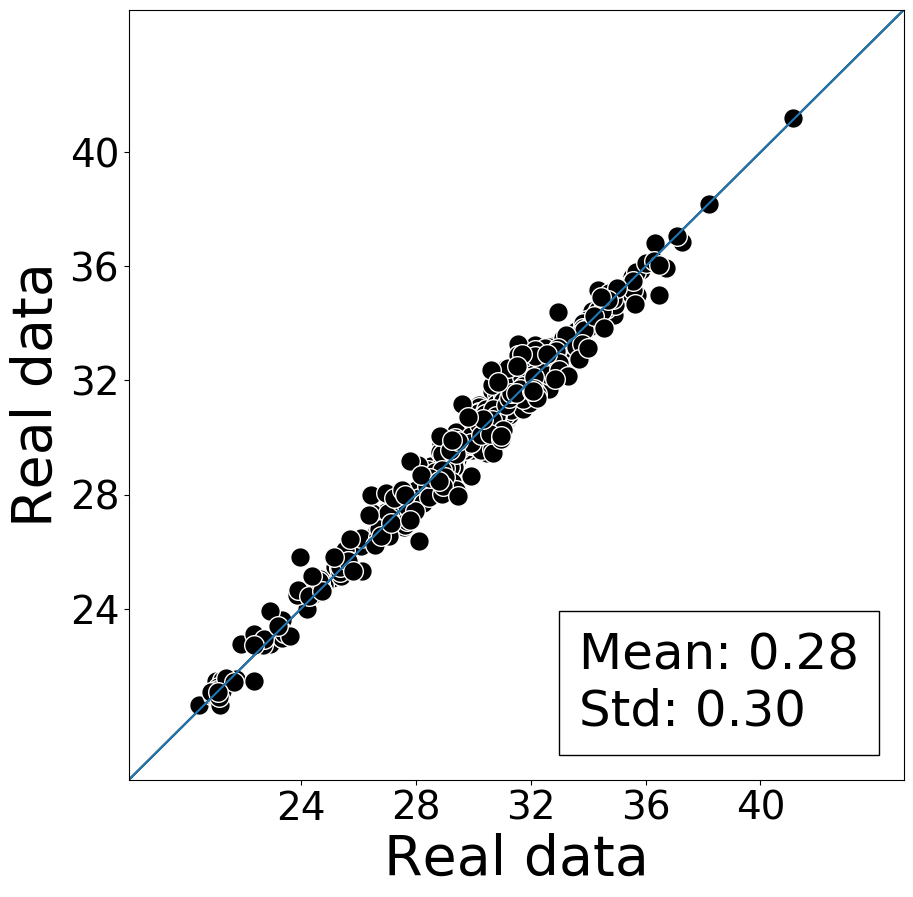}}
\subfigure{%
\label{ccd1_4}%
\includegraphics[width=2.cm]{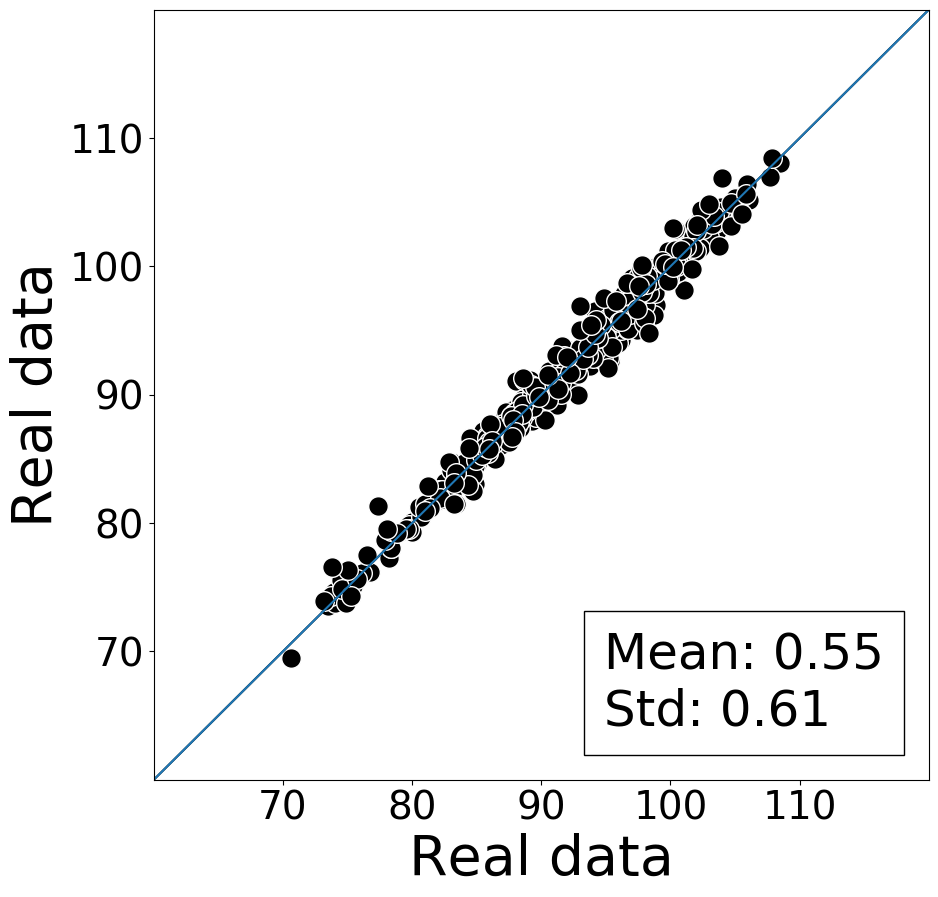}} 
\subfigure{%
\label{ccd1_6}%
\includegraphics[width=2.cm]{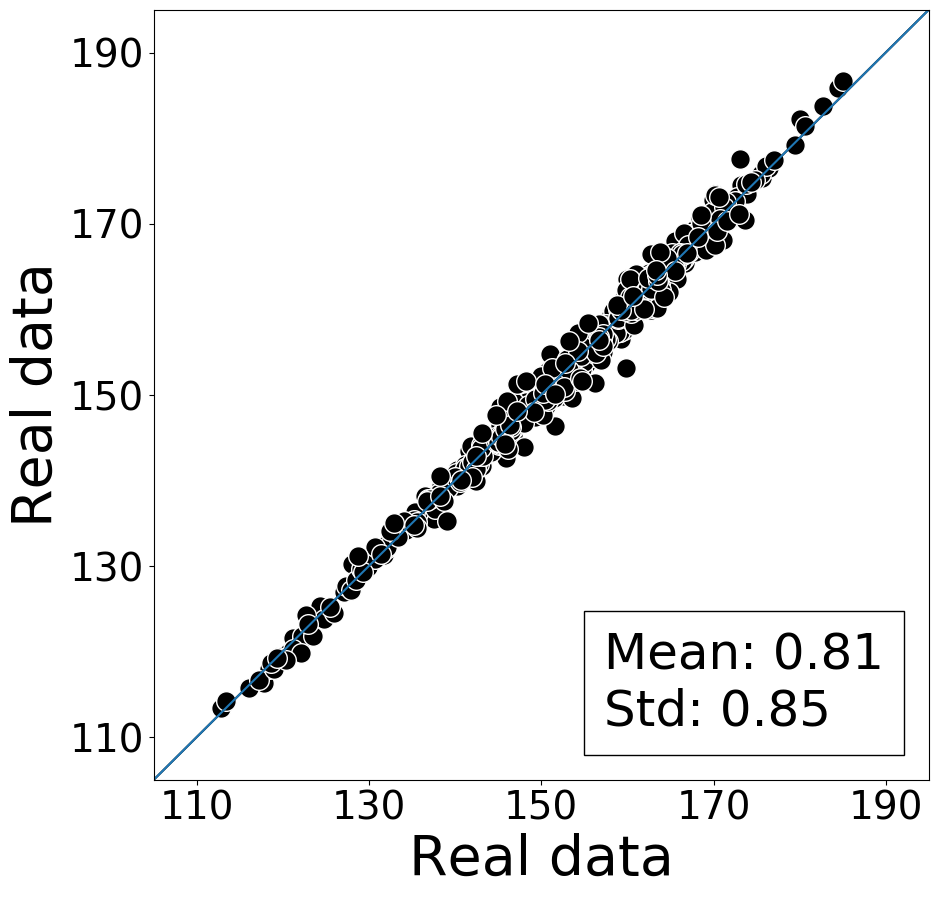}}
\subfigure{%
\label{ccd1_7}%
\includegraphics[width=2.cm]{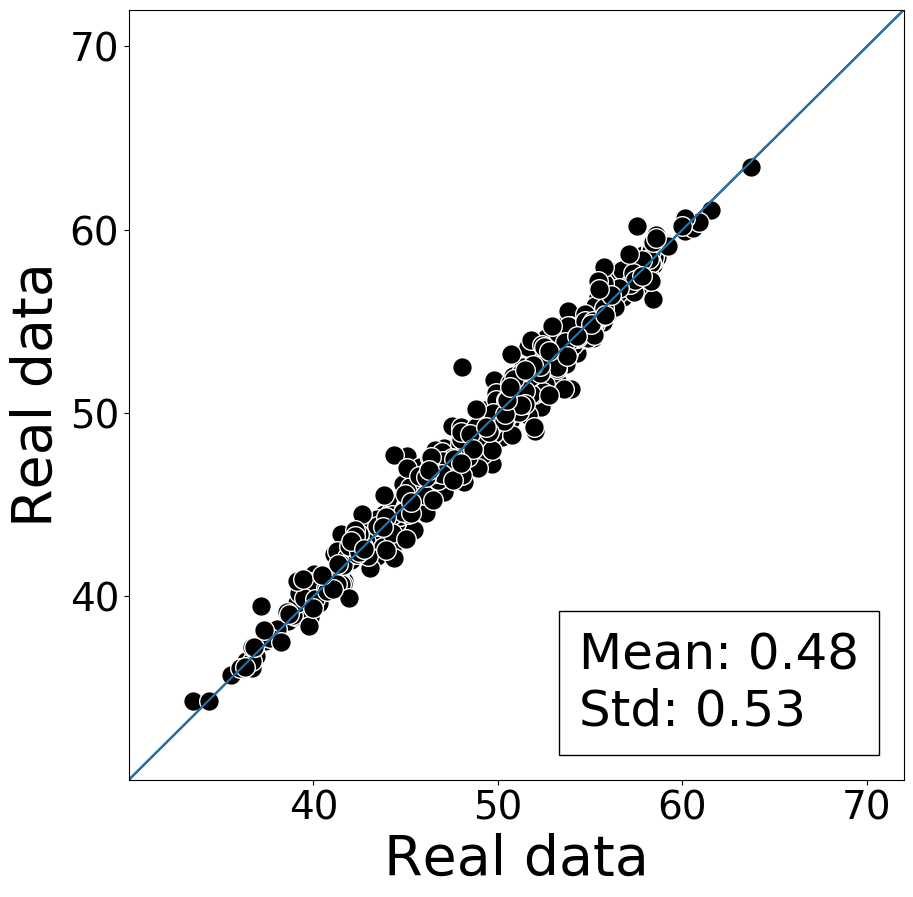}}
\subfigure{%
\label{ccd1_9}%
\includegraphics[width=2.cm]{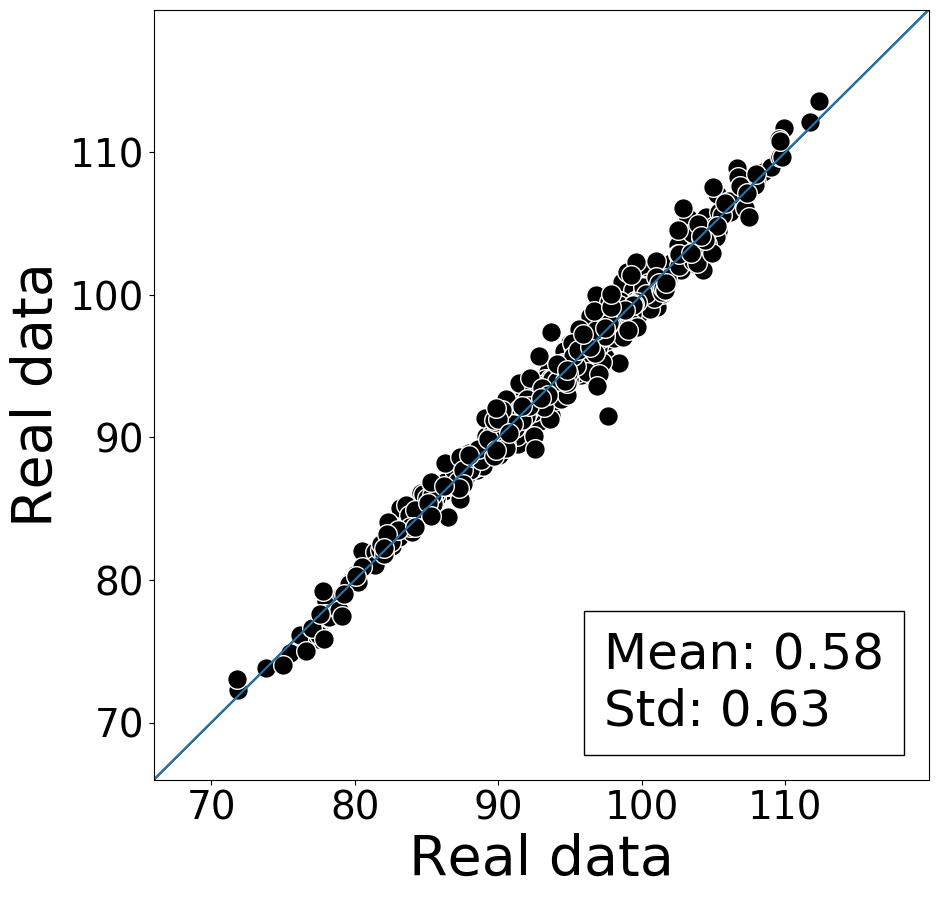}}
\vspace*{-2mm}

\subfigure{%
\includegraphics[width=2.5cm]{./bn_relu.pdf}}
\subfigure{%
\label{ccd3_1}%
\includegraphics[width=2.cm]{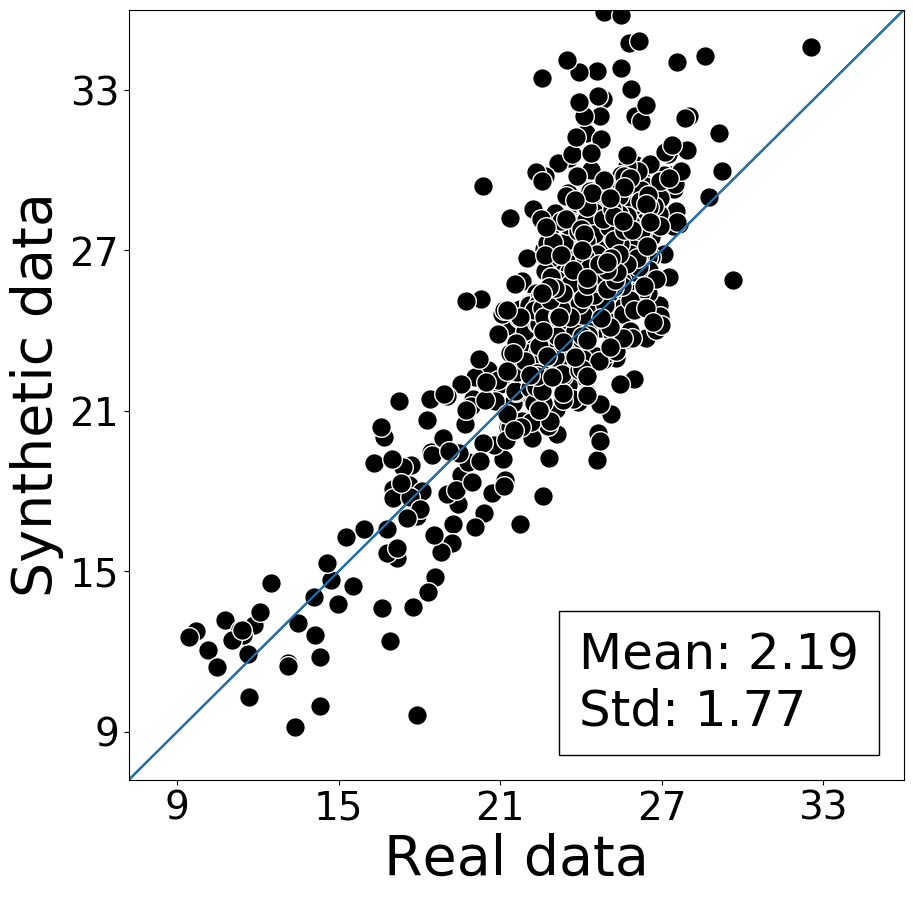}} 
\subfigure{%
\label{ccd3_3}%
\includegraphics[width=2.cm]{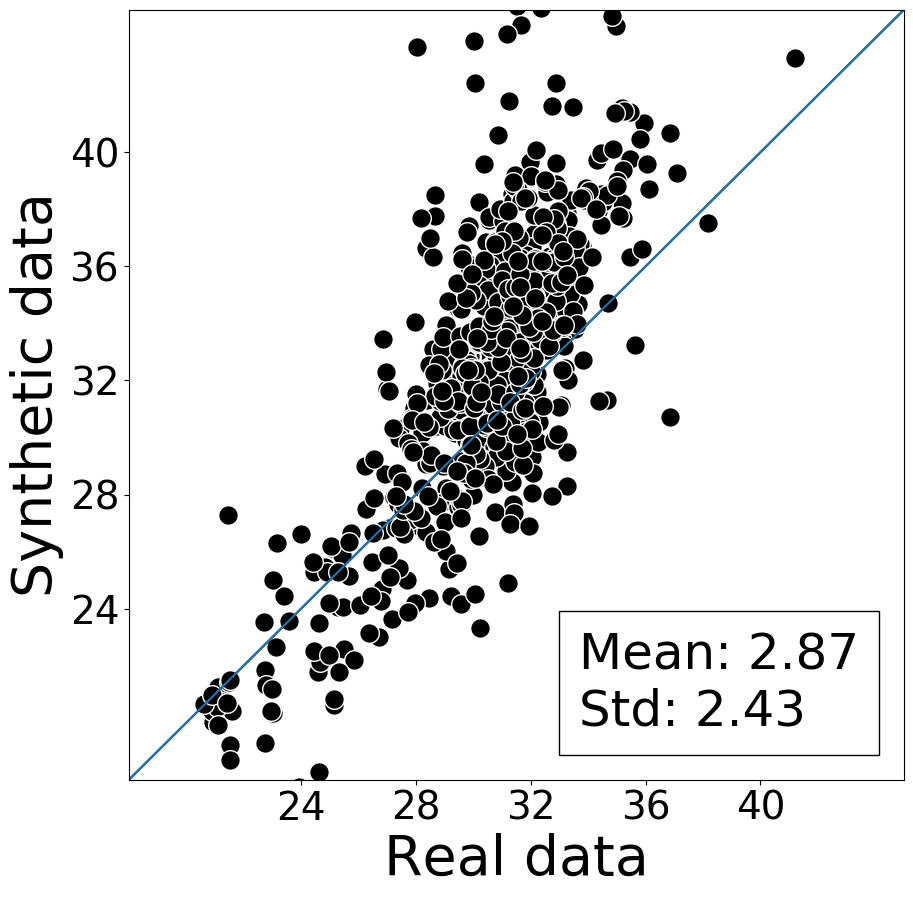}}
\subfigure{%
\label{ccd3_4}%
\includegraphics[width=2.cm]{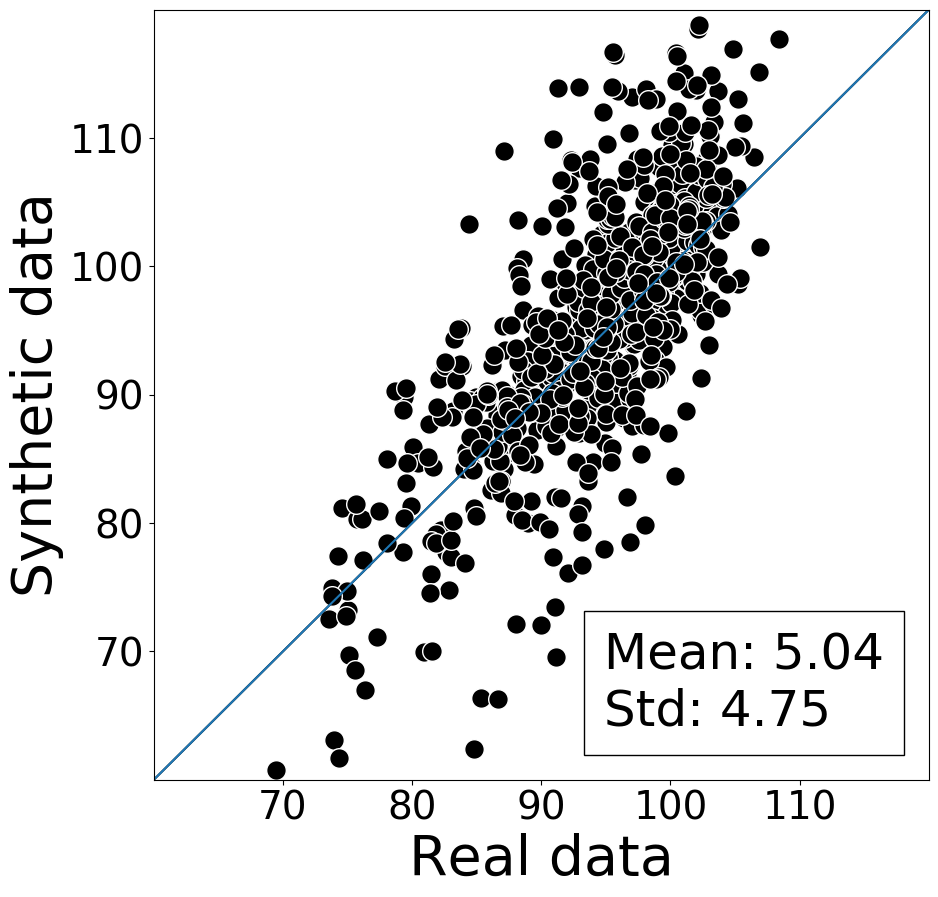}} 
\subfigure{%
\label{ccd3_6}%
\includegraphics[width=2.cm]{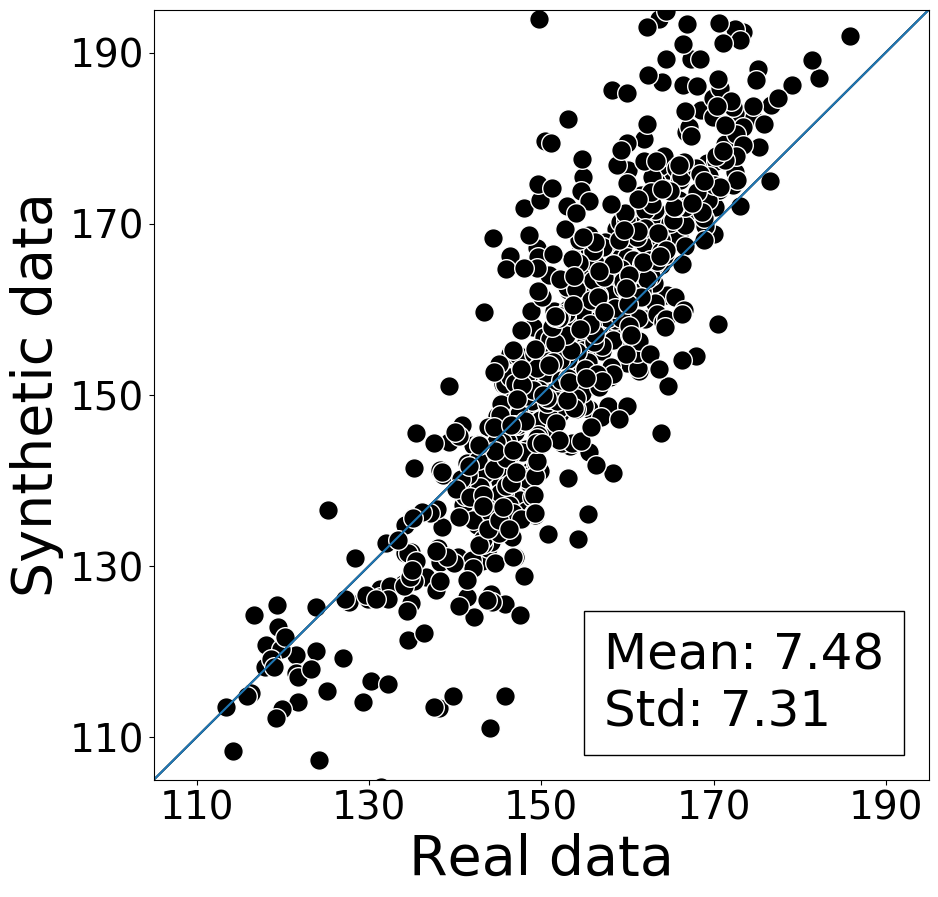}}
\subfigure{%
\label{ccd3_7}%
\includegraphics[width=2.cm]{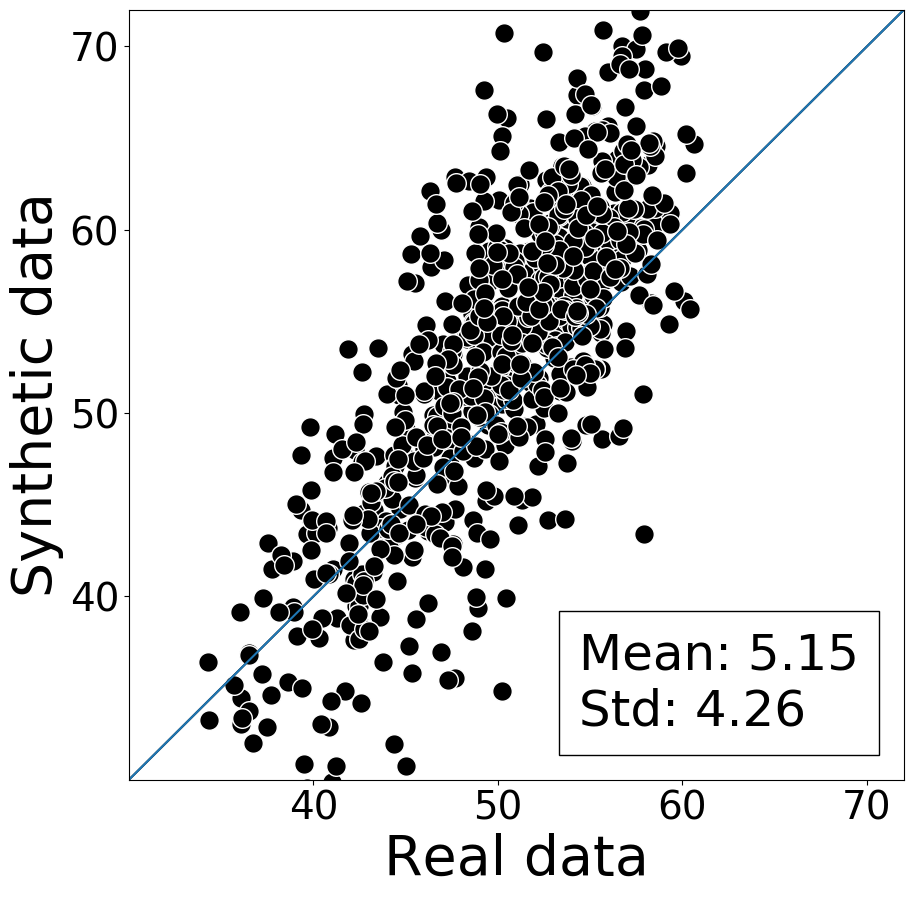}}
\subfigure{%
\label{ccd3_9}%
\includegraphics[width=2.cm]{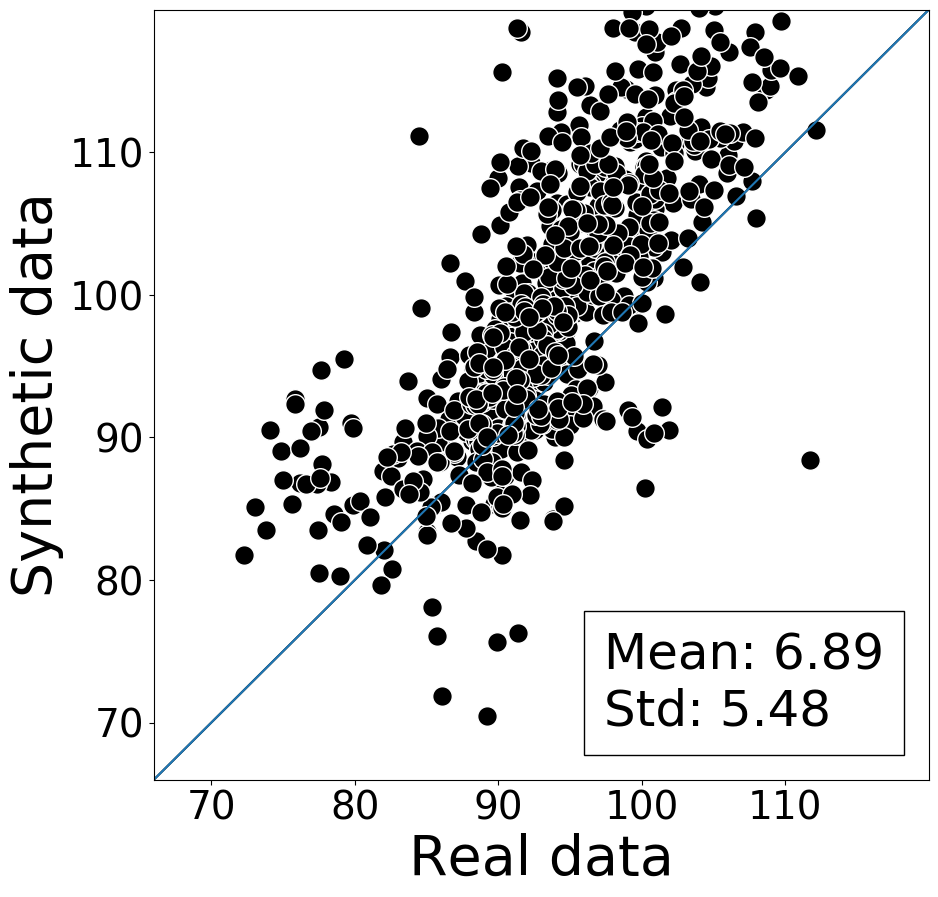}}
\vspace*{-2mm}

\subfigure{%
\includegraphics[width=2.5cm]{./relu_bn.pdf}}
\subfigure{%
\label{ccd2_1}%
\includegraphics[width=2.cm]{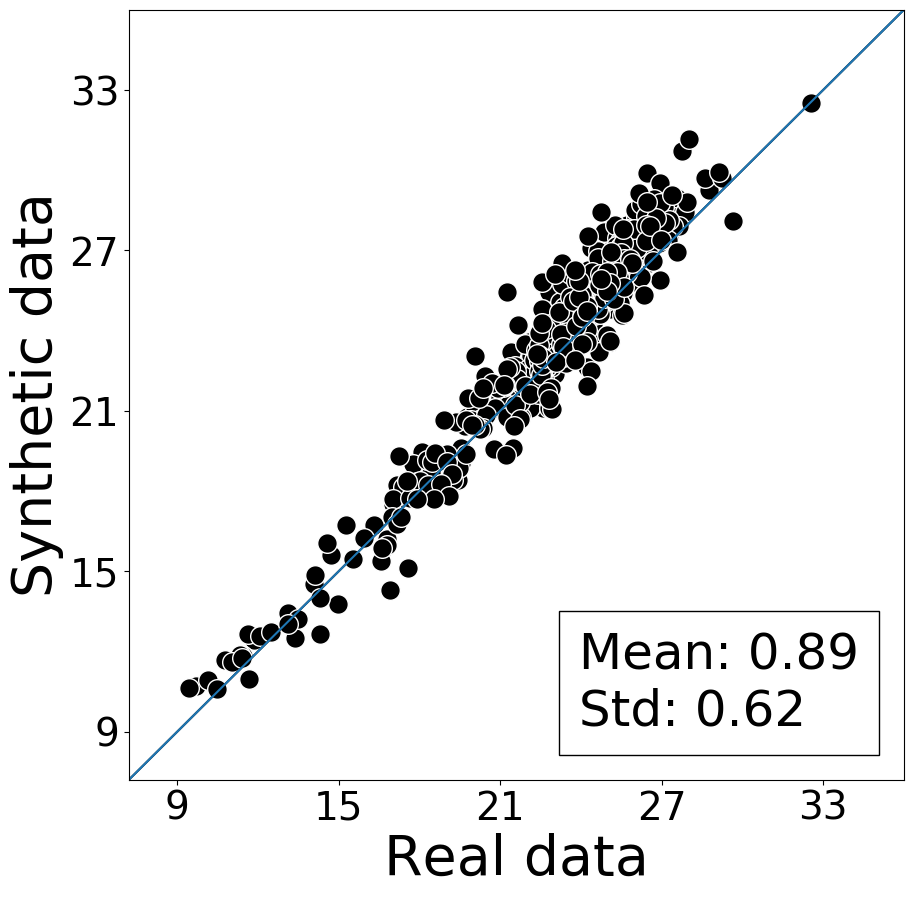}} 
\subfigure{%
\label{ccd2_3}%
\includegraphics[width=2.cm]{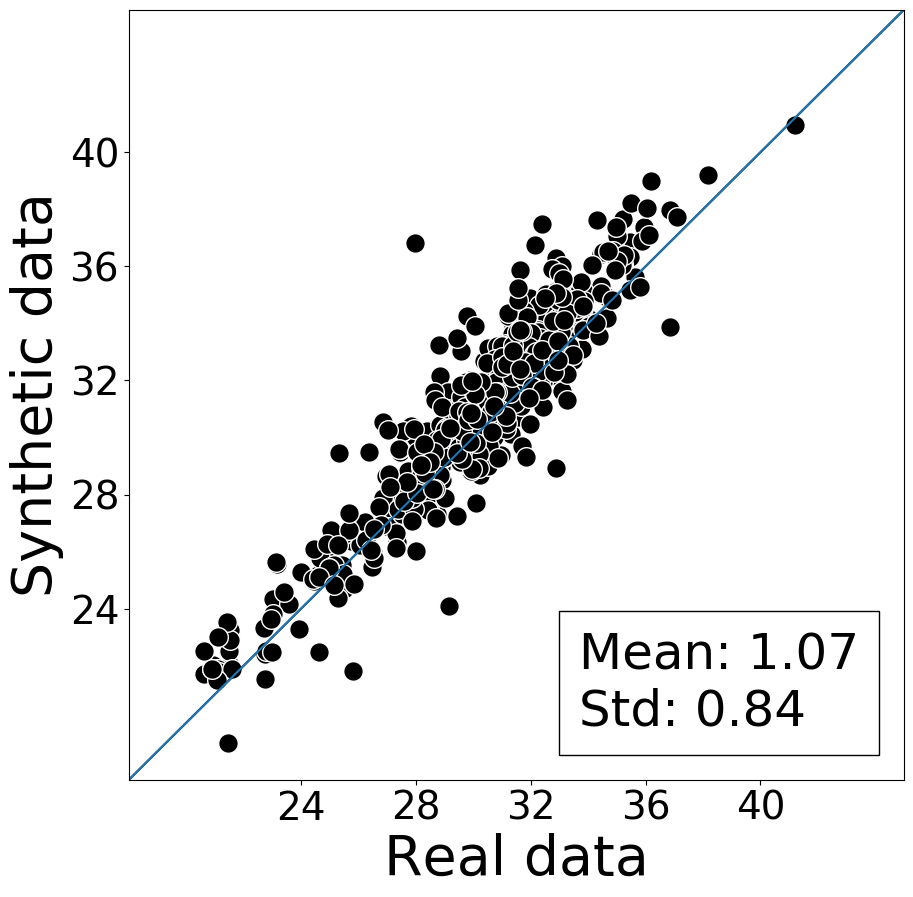}}
\subfigure{%
\label{ccd2_4}%
\includegraphics[width=2.cm]{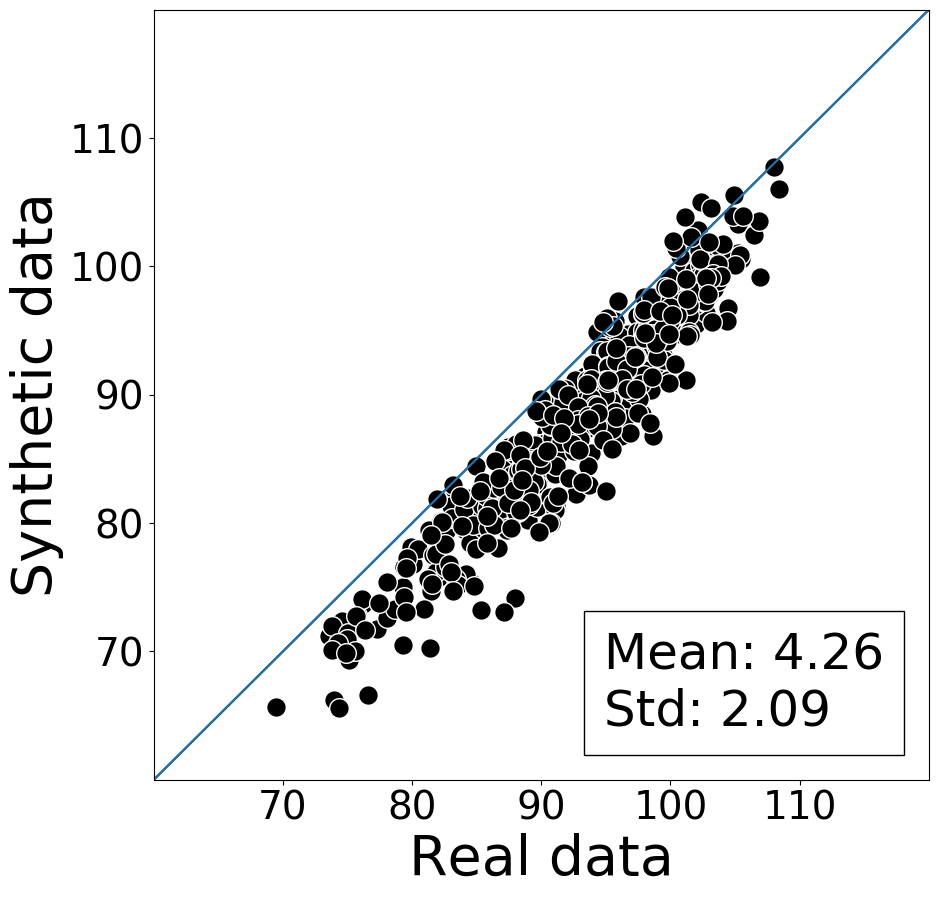}} 
\subfigure{%
\label{ccd2_6}%
\includegraphics[width=2.cm]{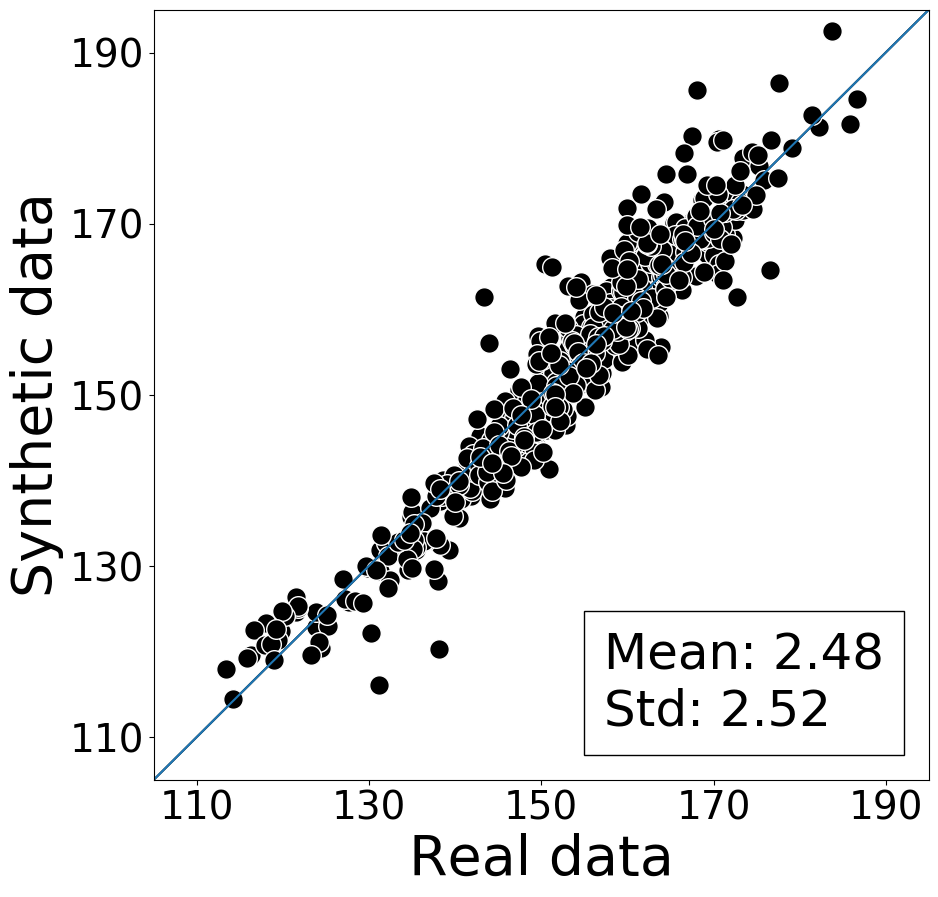}}
\subfigure{%
\label{ccd2_7}%
\includegraphics[width=2.cm]{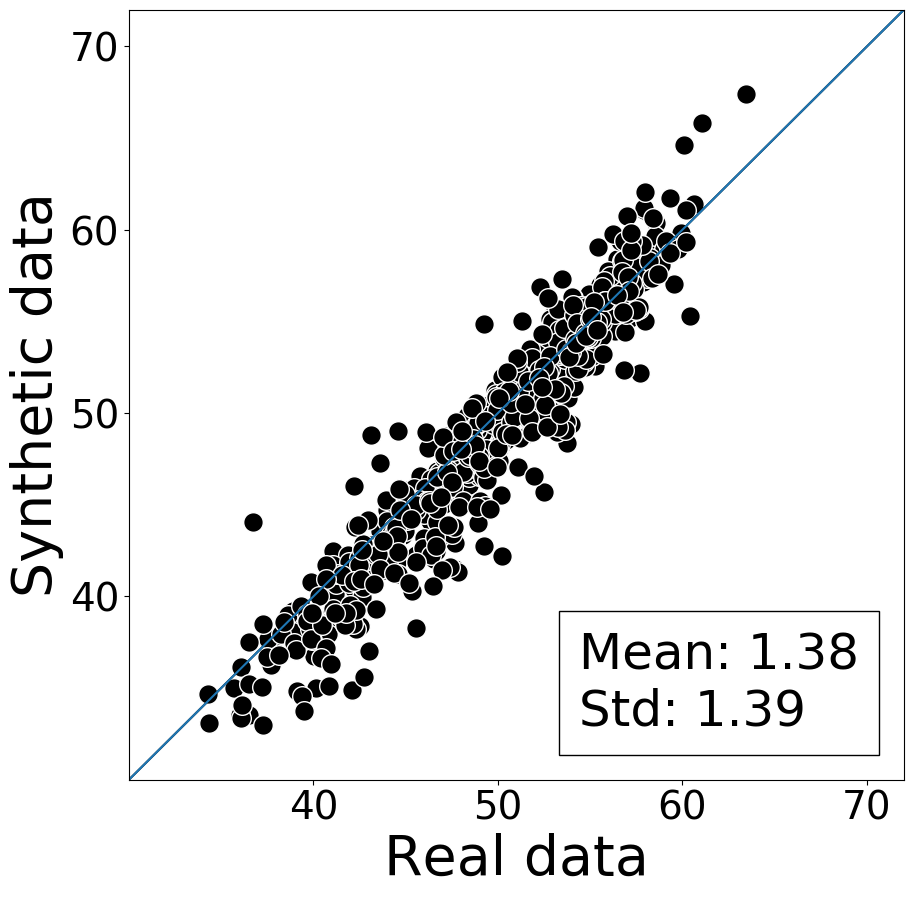}}
\subfigure{%
\label{ccd2_9}%
\includegraphics[width=2.cm]{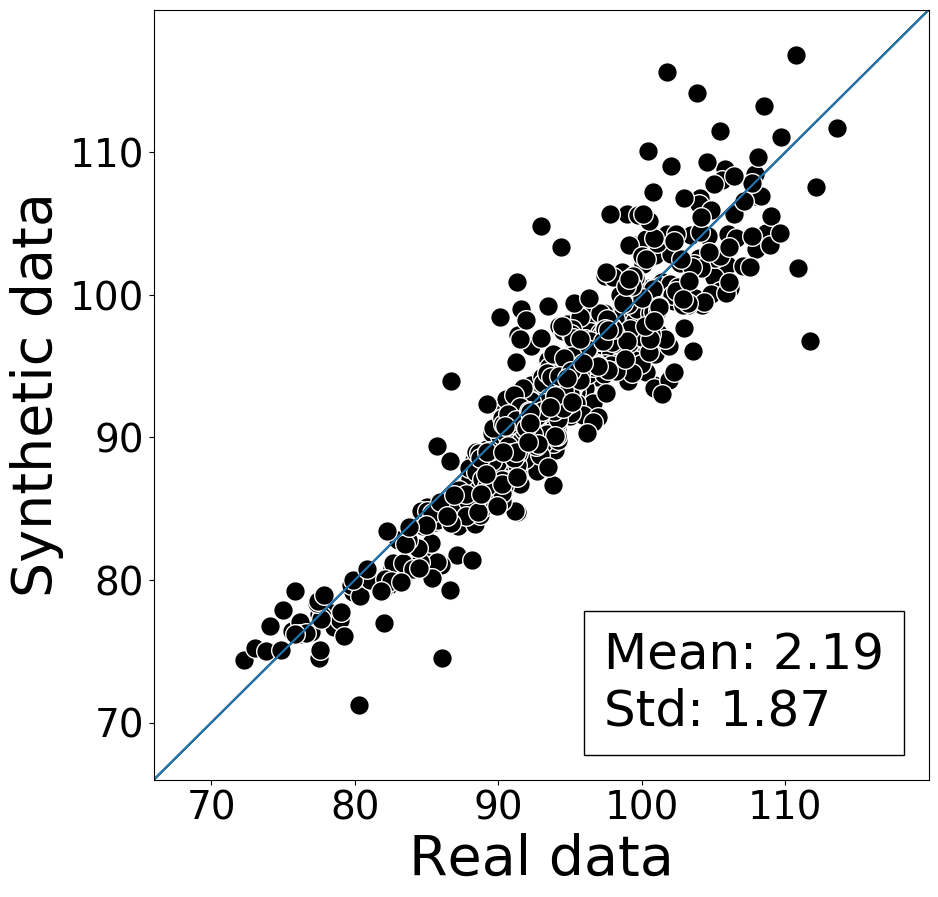}}
\vspace*{-1mm}

\subfigure{%
\includegraphics[width=2.5cm]{./d0.pdf}}
\subfigure{%
\label{ccd2_1}%
\includegraphics[width=2.cm]{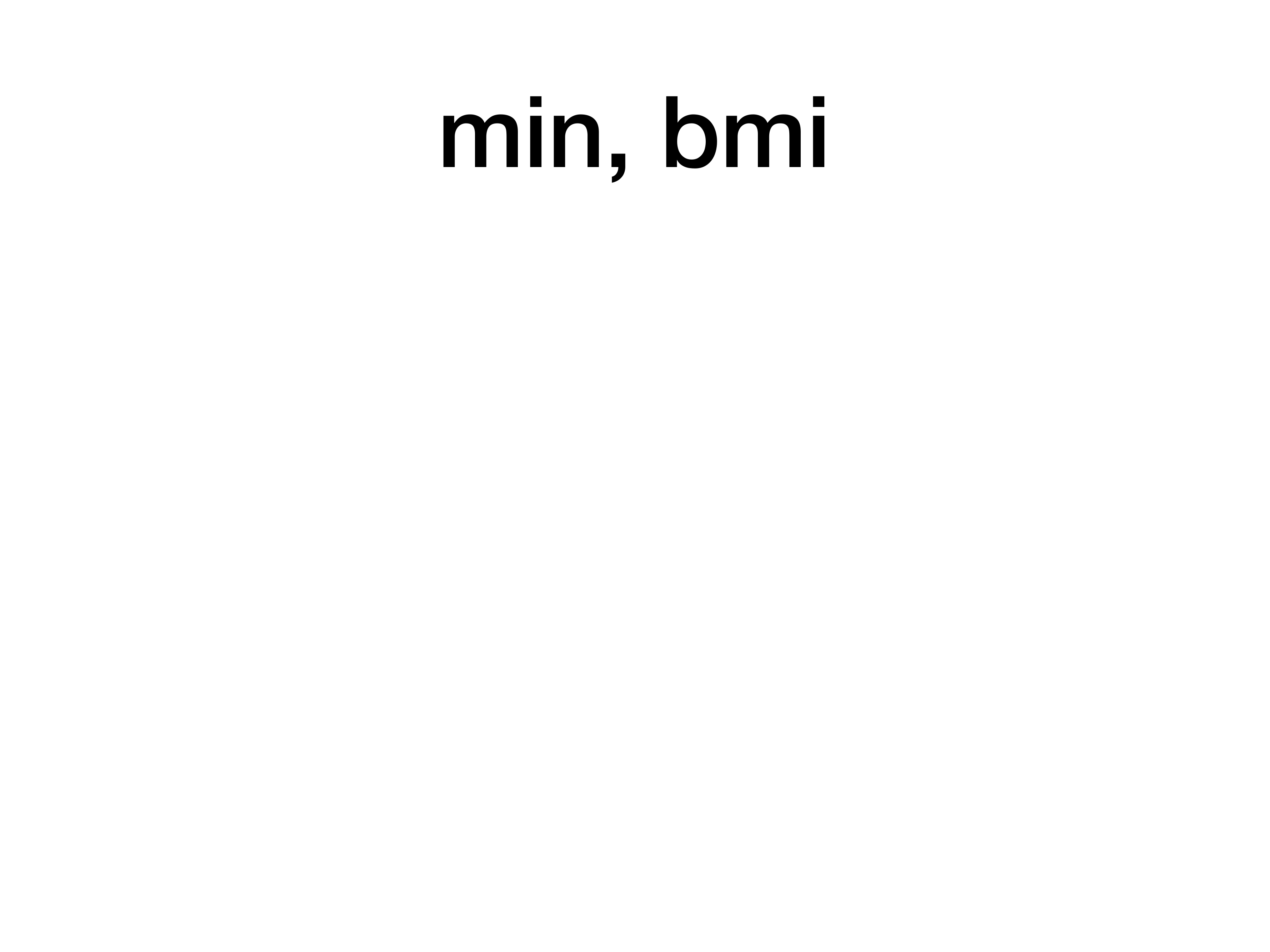}} 
\subfigure{%
\label{ccd2_3}%
\includegraphics[width=2.cm]{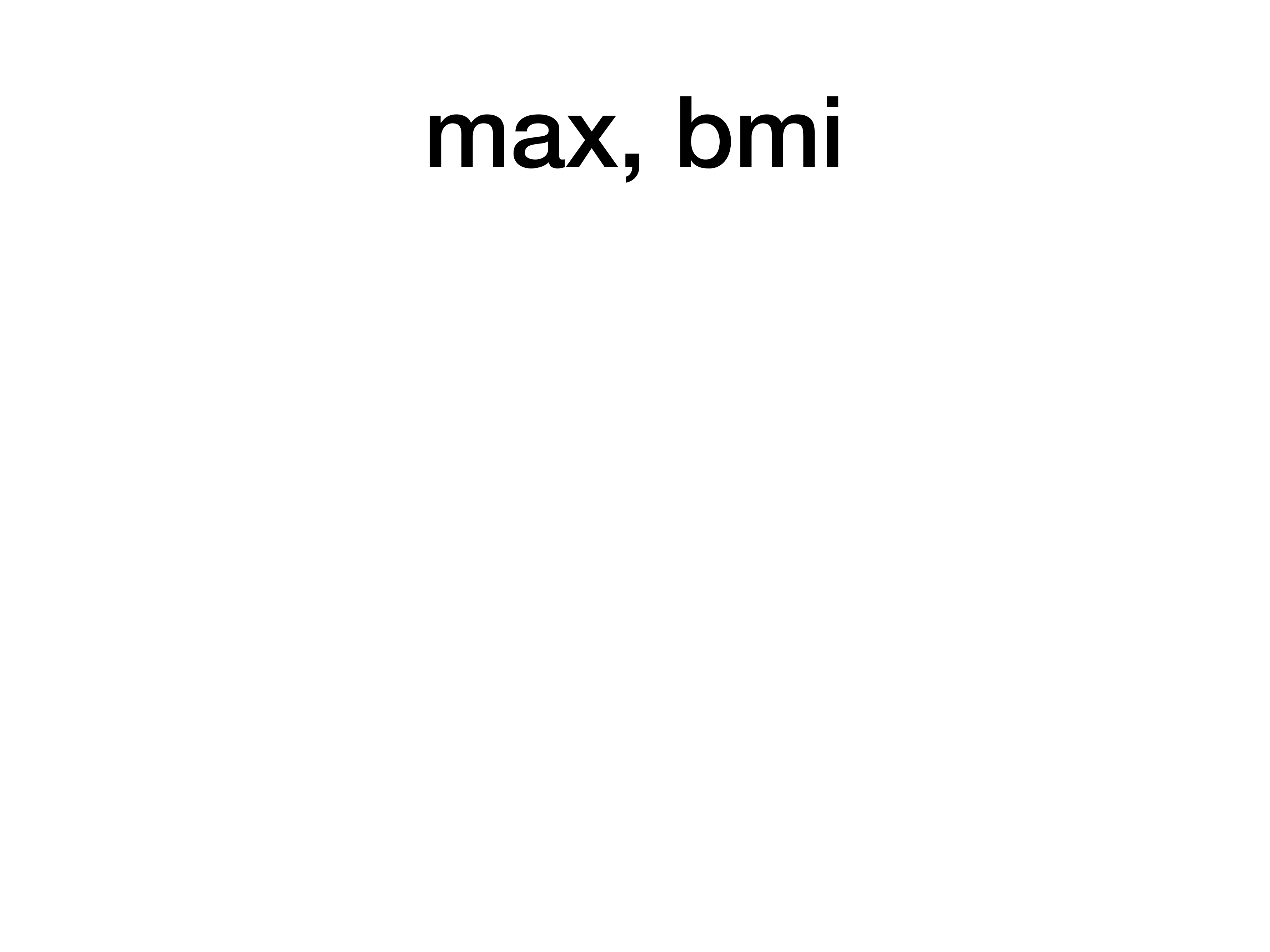}}
\subfigure{%
\label{ccd2_4}%
\includegraphics[width=2.cm]{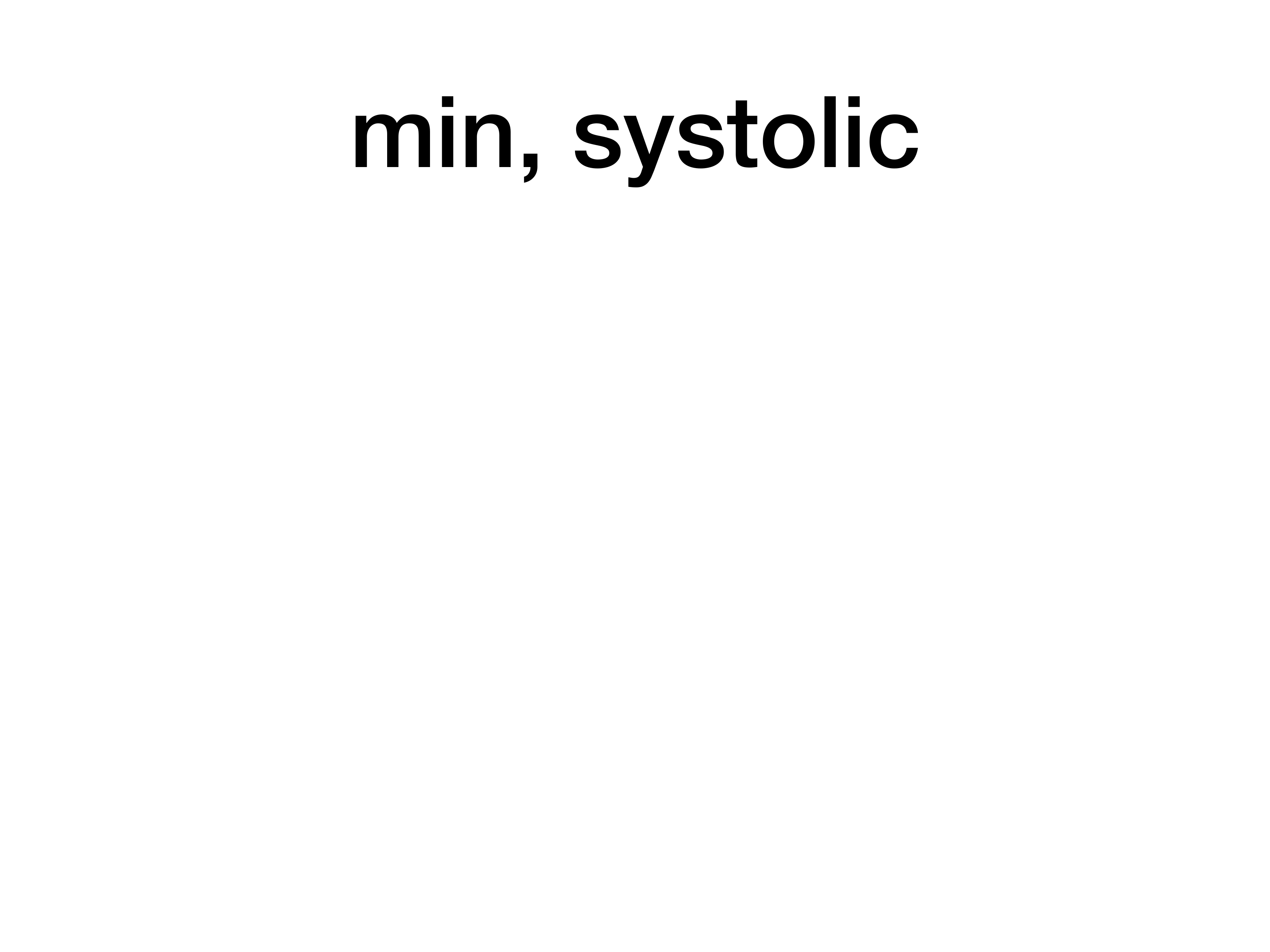}} 
\subfigure{%
\label{ccd2_6}%
\includegraphics[width=2.cm]{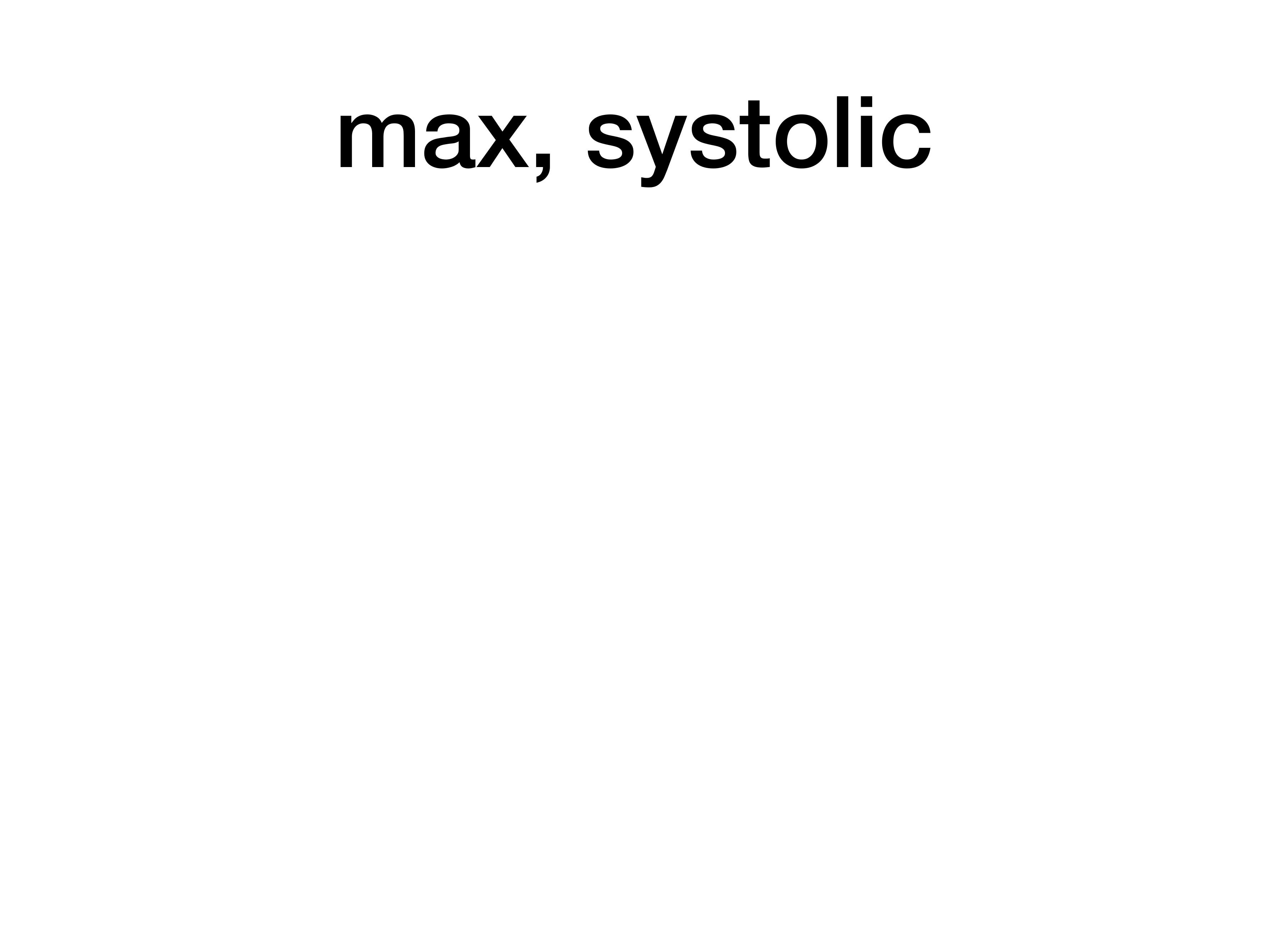}}
\subfigure{%
\label{ccd2_7}%
\includegraphics[width=2.cm]{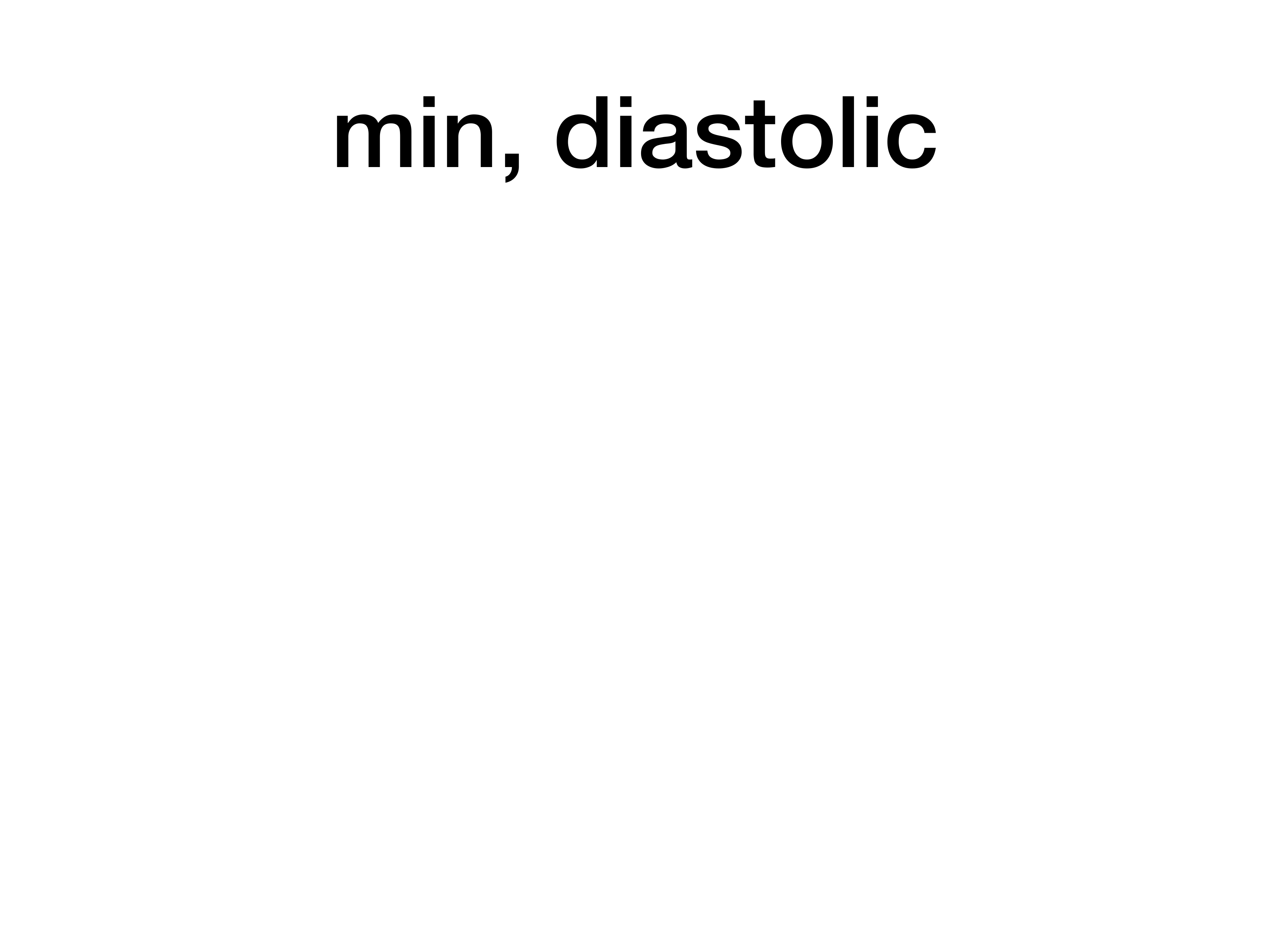}}
\subfigure{%
\label{ccd2_9}%
\includegraphics[width=2.cm]{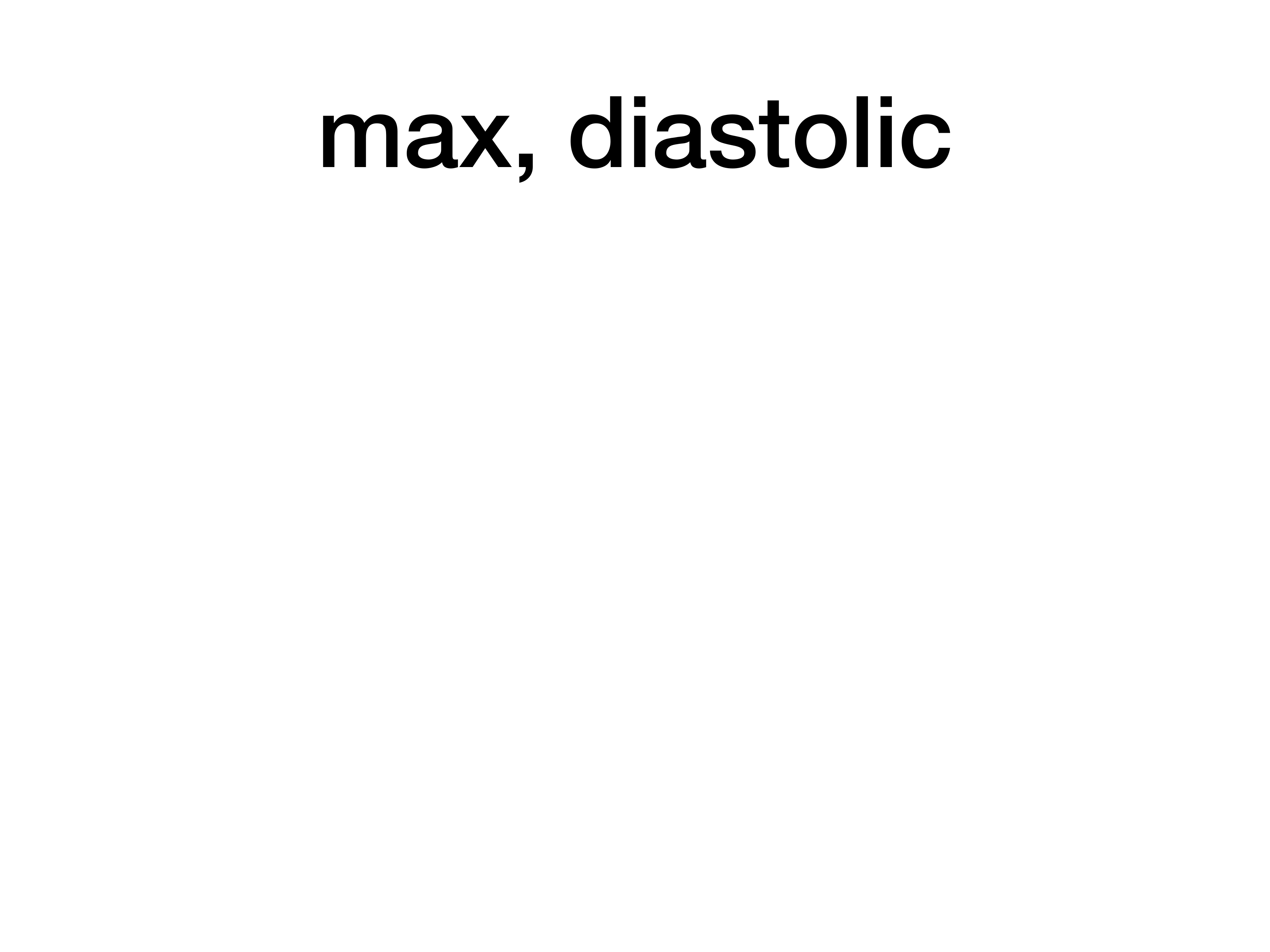}}
\vspace*{-20mm}
\caption{Cross-type conditional distribution on vital signs. The coordinates of each dot correspond to the means of the vital sign distributions on the real and synthetic dataset.}\label{cond_lab}
\end{figure}

\begin{figure}[ht]%
\captionsetup[subfigure]{justification=centering}
\centering
\subfigure{%
\includegraphics[width=2.5cm]{./real.pdf}}
\subfigure{%
\label{std_ccd1_1}%
\includegraphics[width=2.cm]{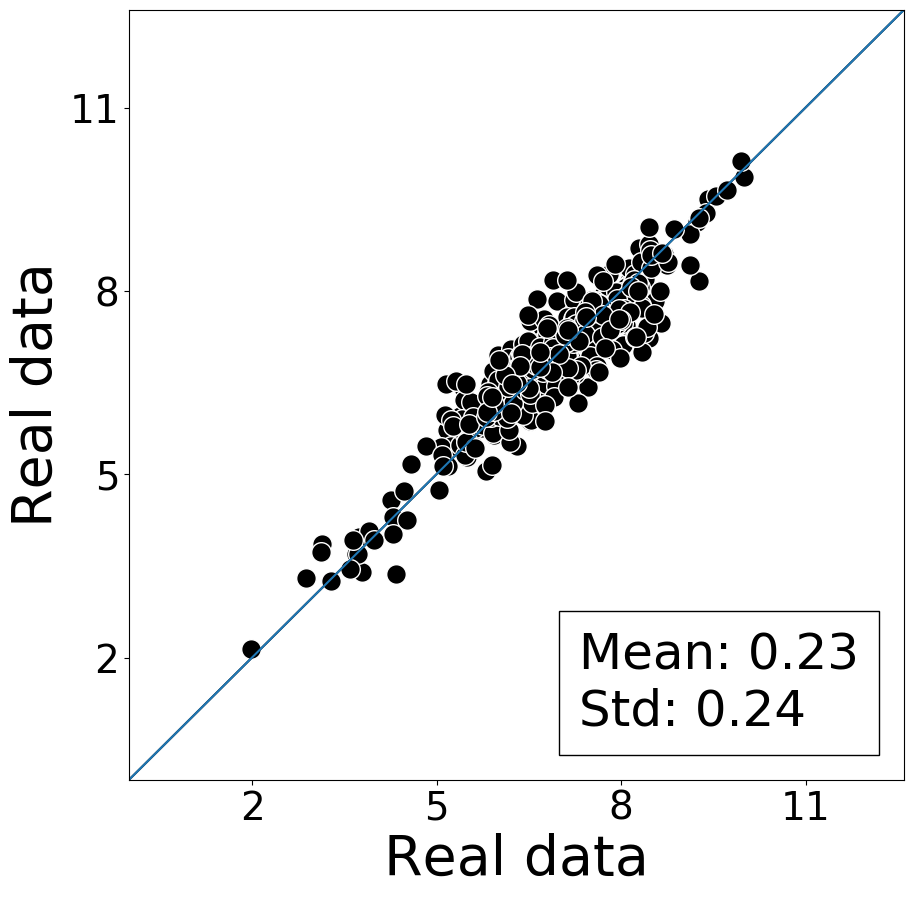}} 
\subfigure{%
\label{std_ccd1_3}%
\includegraphics[width=2.cm]{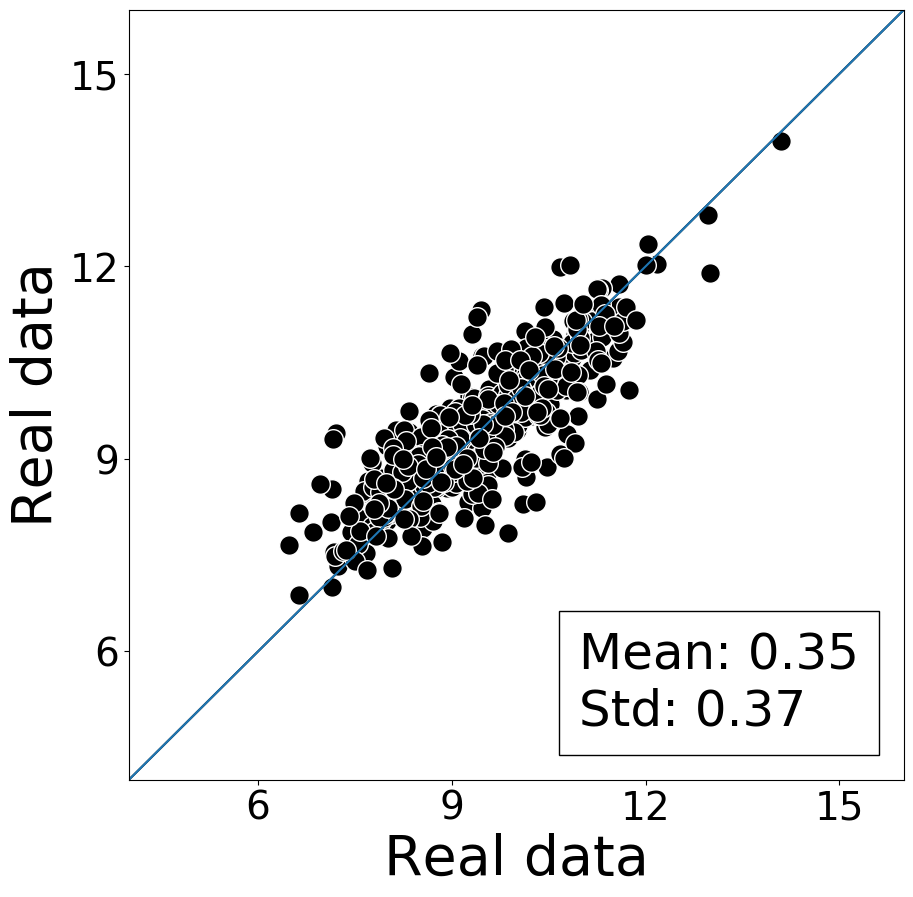}}
\subfigure{%
\label{std_ccd1_4}%
\includegraphics[width=2.cm]{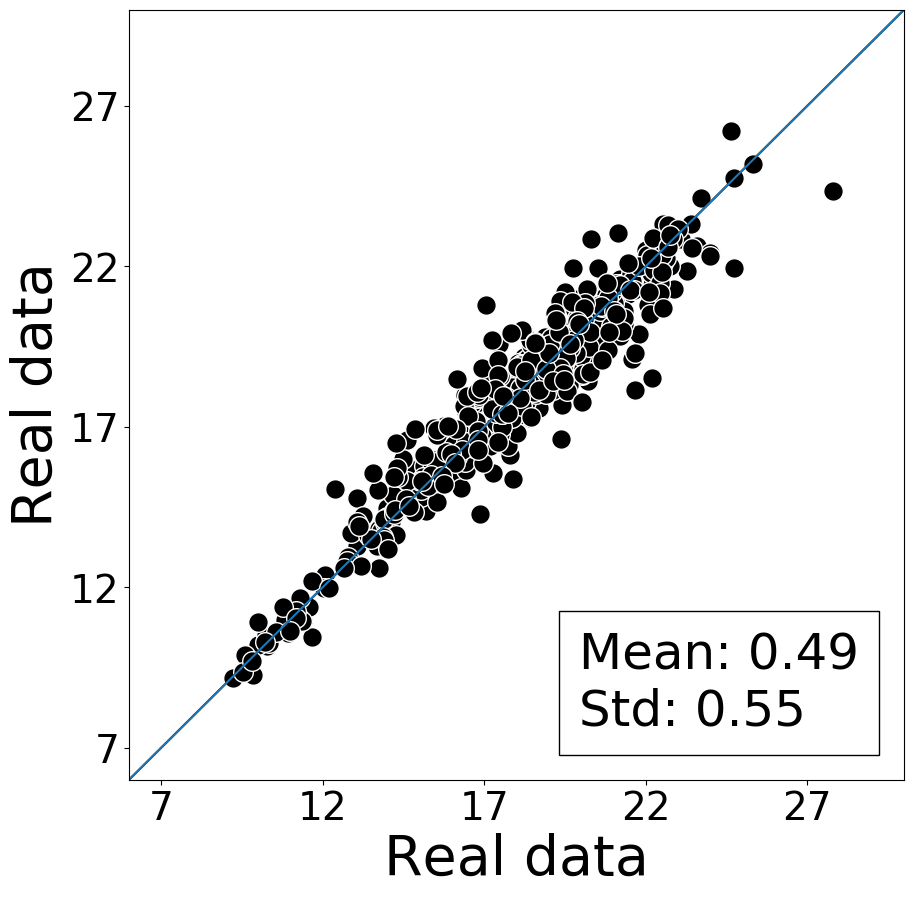}} 
\subfigure{%
\label{std_ccd1_6}%
\includegraphics[width=2.cm]{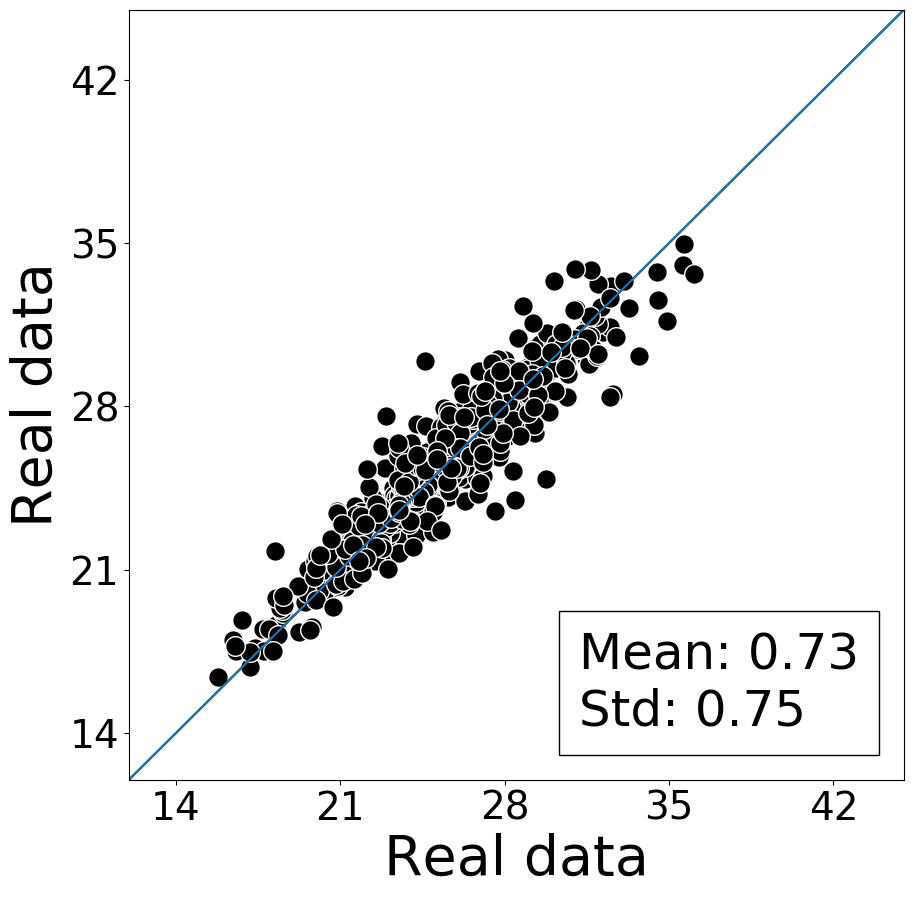}}
\subfigure{%
\label{std_ccd1_7}%
\includegraphics[width=2.cm]{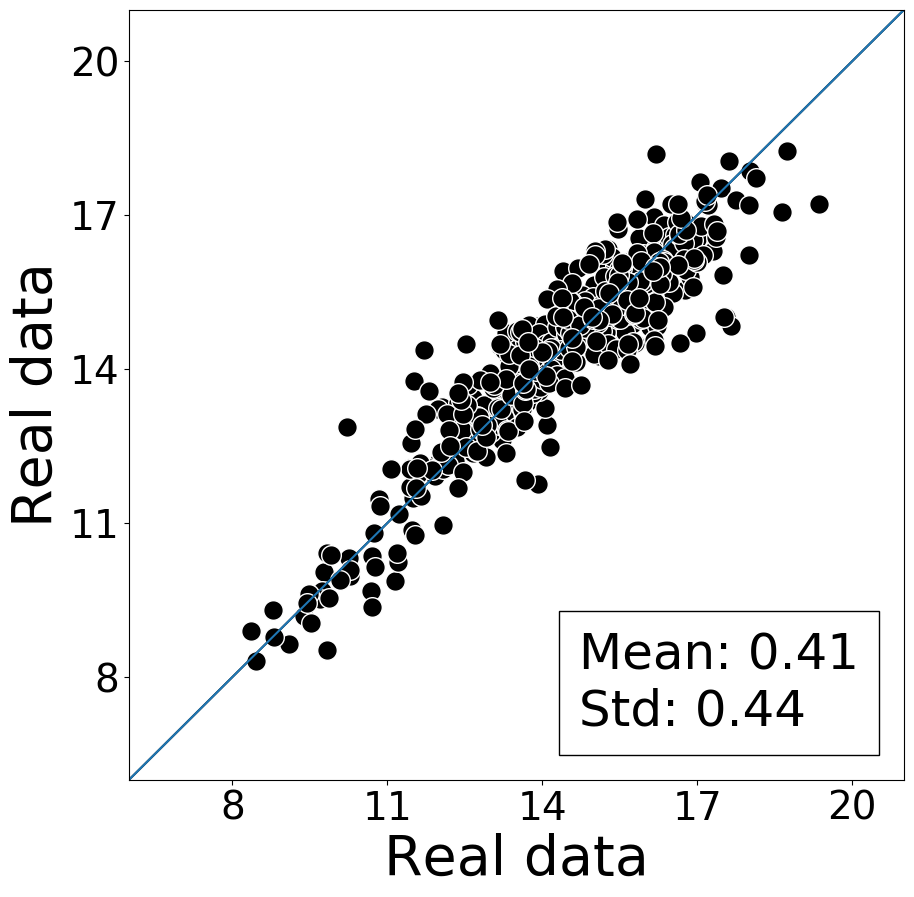}}
\subfigure{%
\label{std_ccd1_9}%
\includegraphics[width=2.cm]{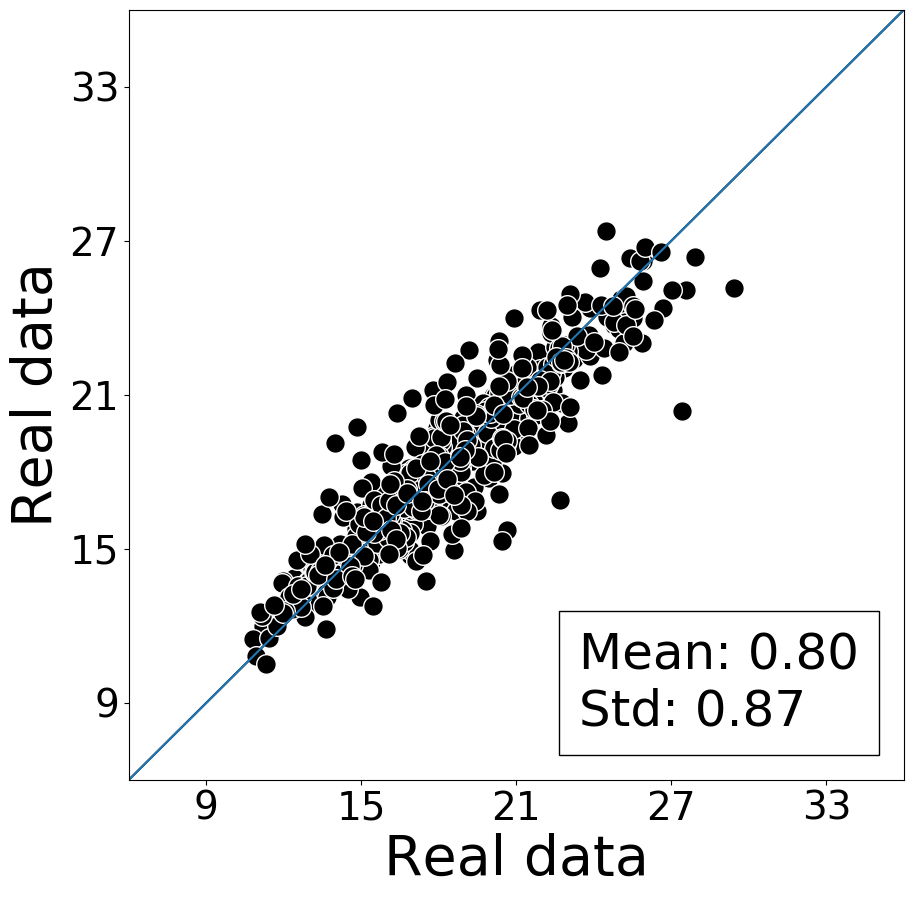}}
\vspace*{-2mm}

\subfigure{%
\includegraphics[width=2.5cm]{./bn_relu.pdf}}
\subfigure{%
\label{std_ccd3_1}%
\includegraphics[width=2.cm]{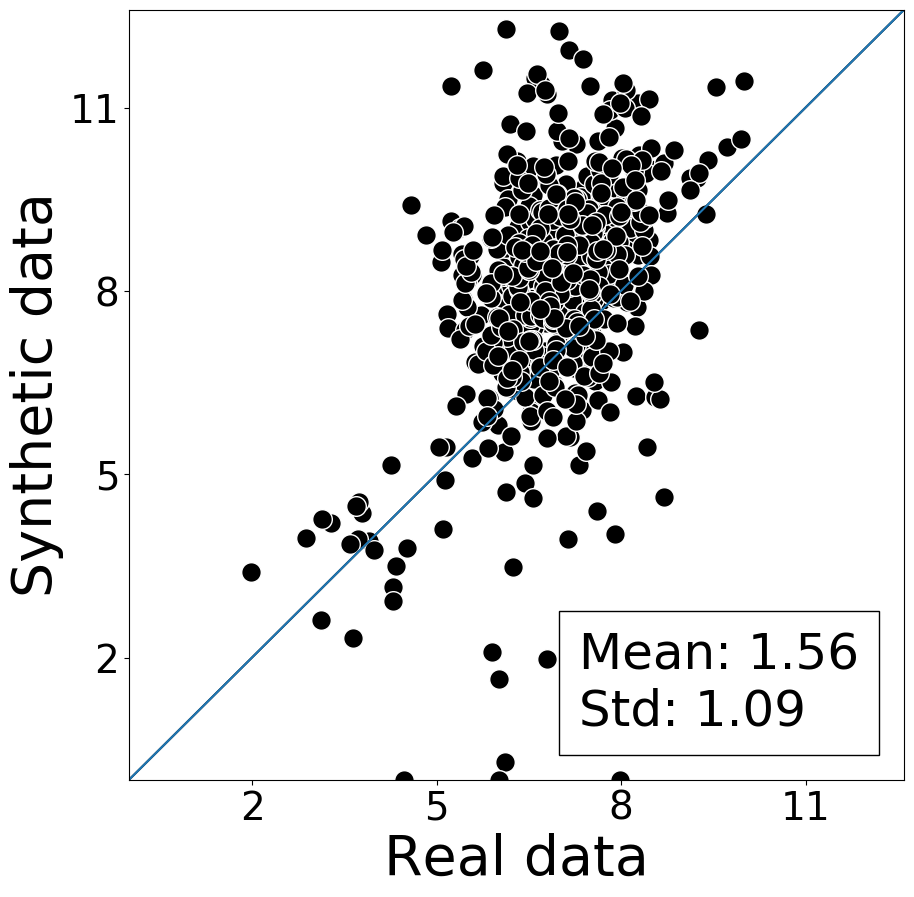}} 
\subfigure{%
\label{std_ccd3_3}%
\includegraphics[width=2.cm]{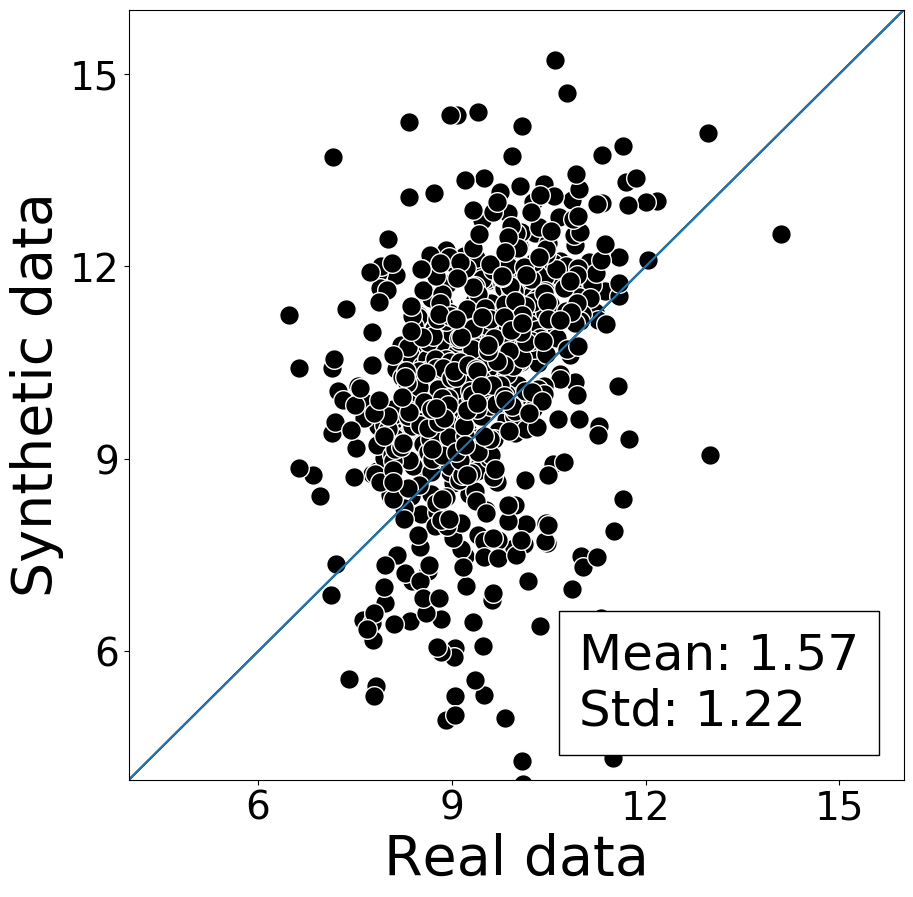}}
\subfigure{%
\label{std_ccd3_4}%
\includegraphics[width=2.cm]{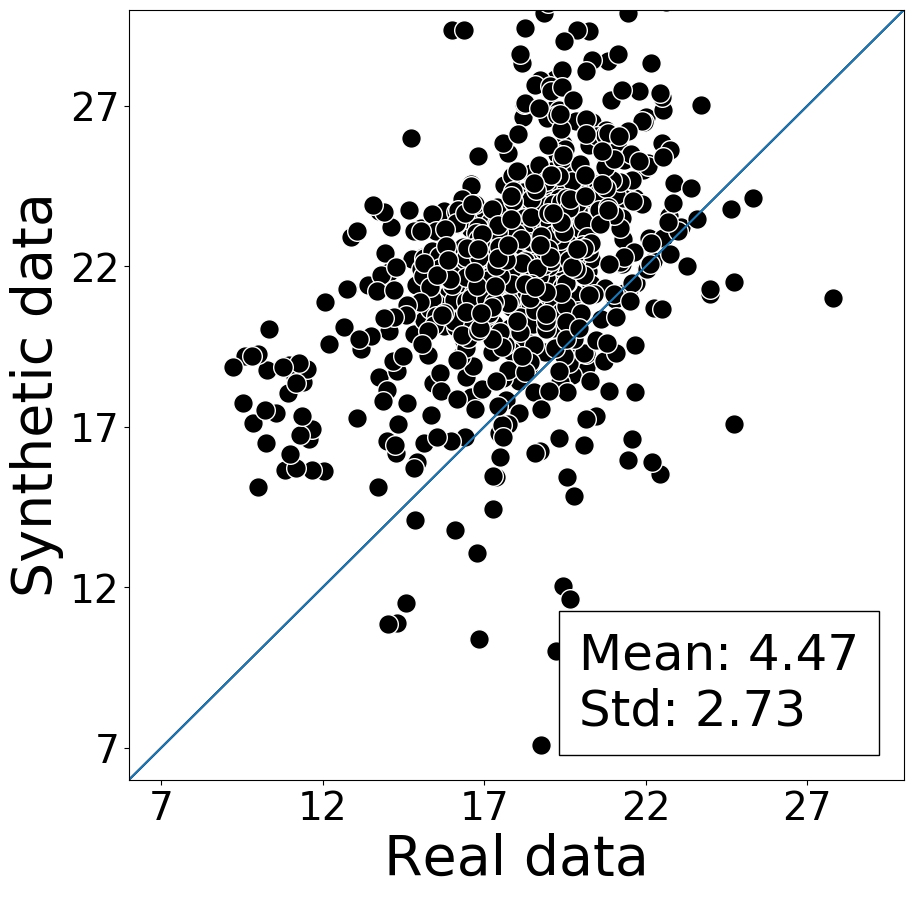}} 
\subfigure{%
\label{std_ccd3_6}%
\includegraphics[width=2.cm]{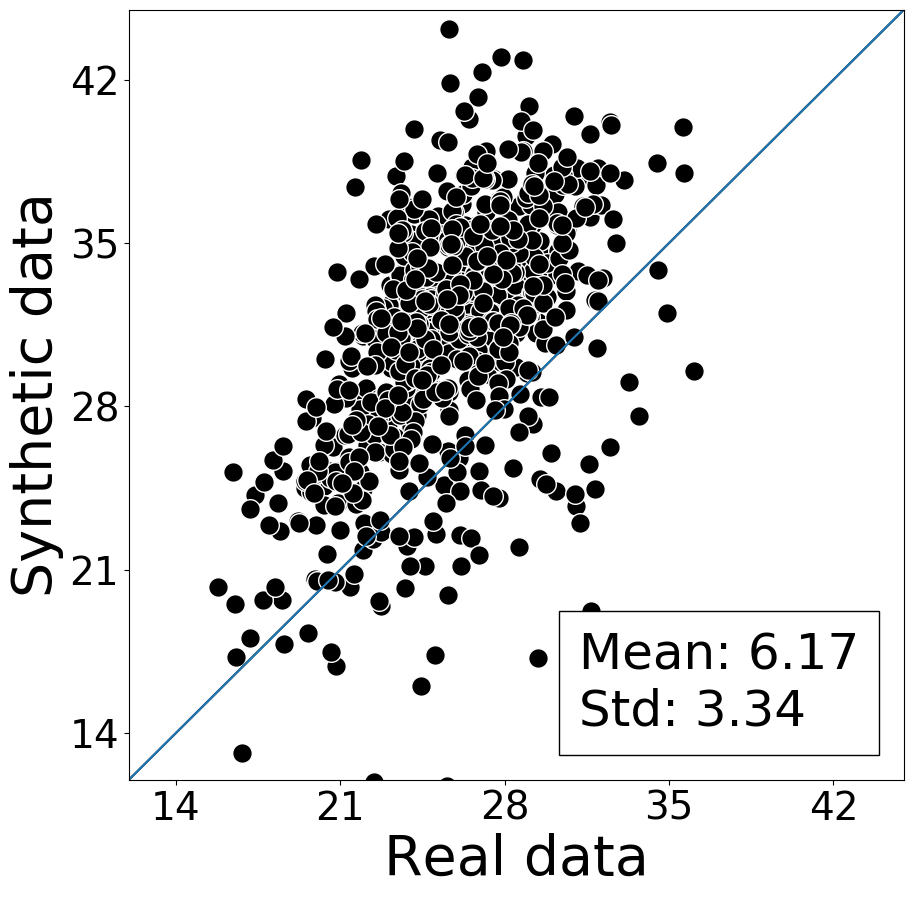}}
\subfigure{%
\label{std_ccd3_7}%
\includegraphics[width=2.cm]{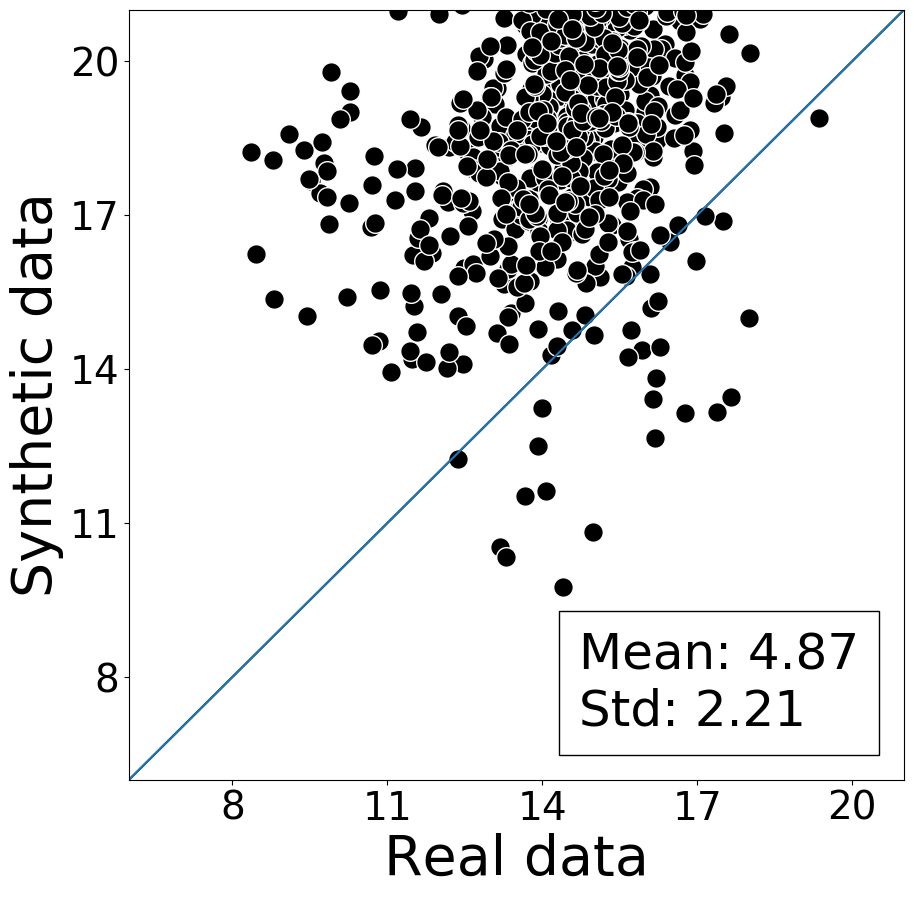}}
\subfigure{%
\label{std_ccd3_9}%
\includegraphics[width=2.cm]{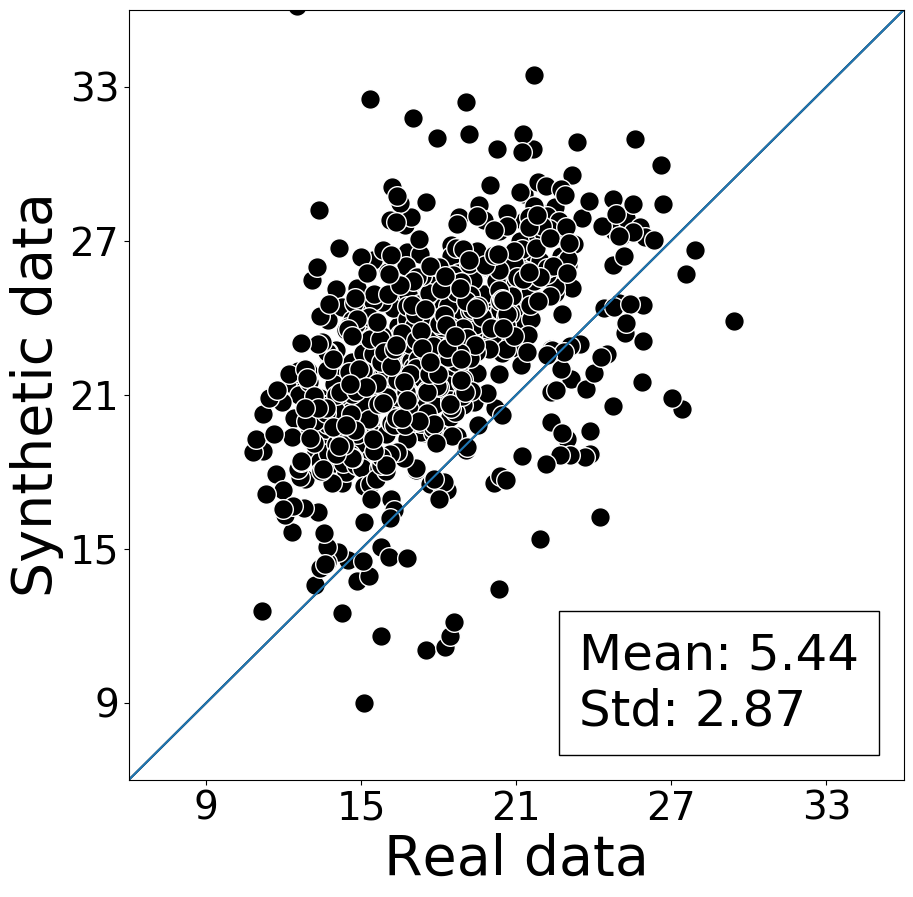}}
\vspace*{-2mm}

\subfigure{%
\includegraphics[width=2.5cm]{./relu_bn.pdf}}
\subfigure{%
\label{std_ccd2_1}%
\includegraphics[width=2.cm]{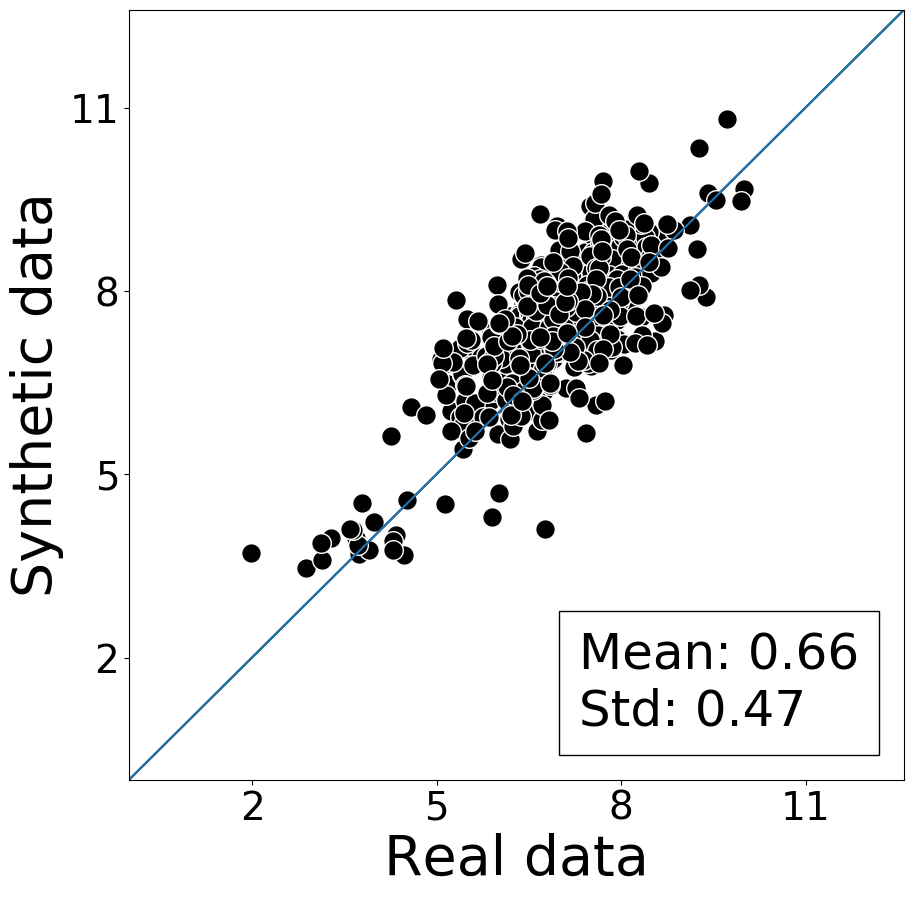}} 
\subfigure{%
\label{std_ccd2_3}%
\includegraphics[width=2.cm]{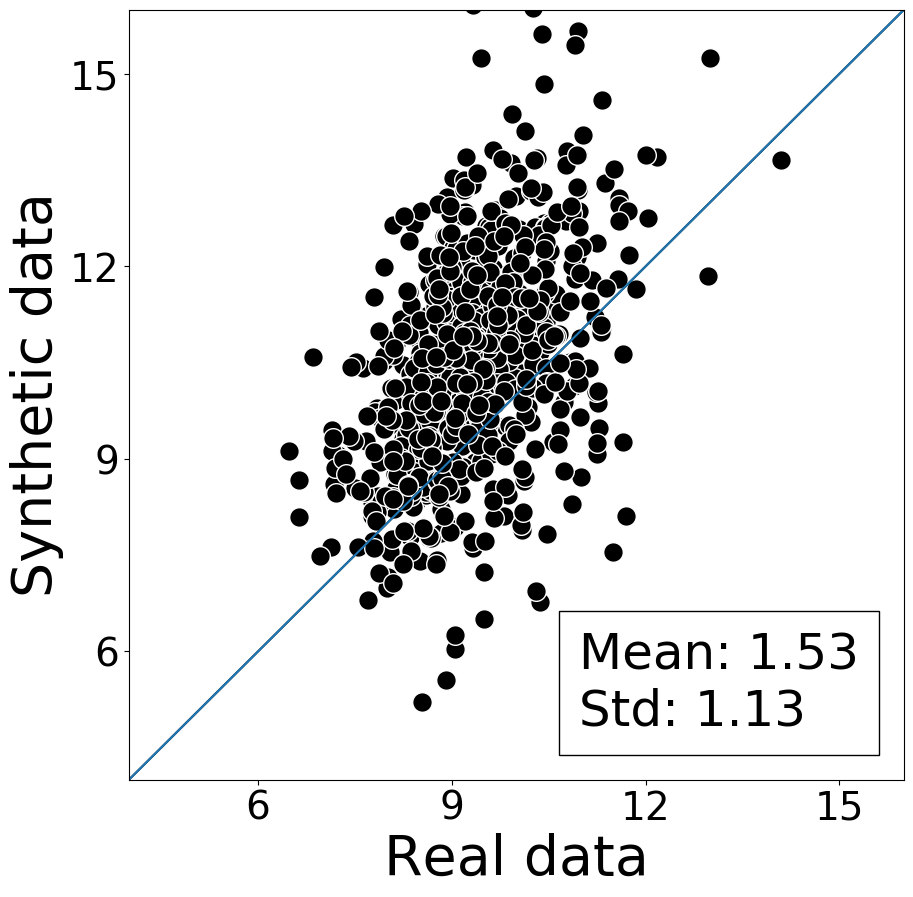}}
\subfigure{%
\label{std_ccd2_4}%
\includegraphics[width=2.cm]{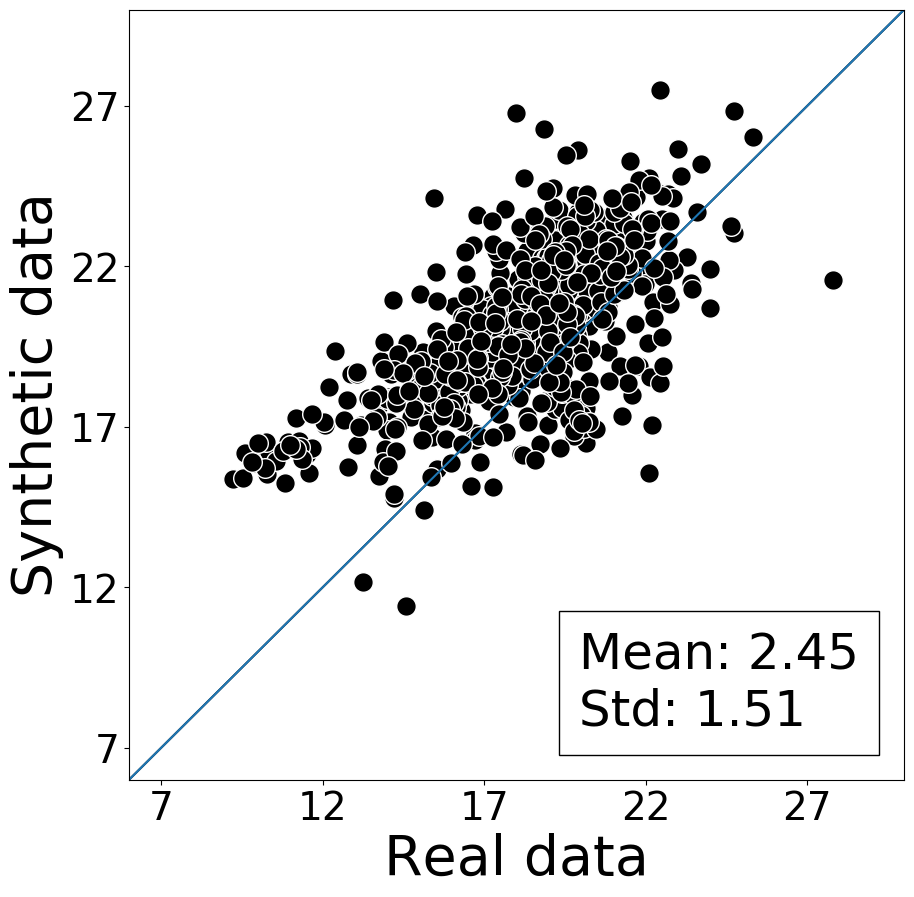}} 
\subfigure{%
\label{std_ccd2_6}%
\includegraphics[width=2.cm]{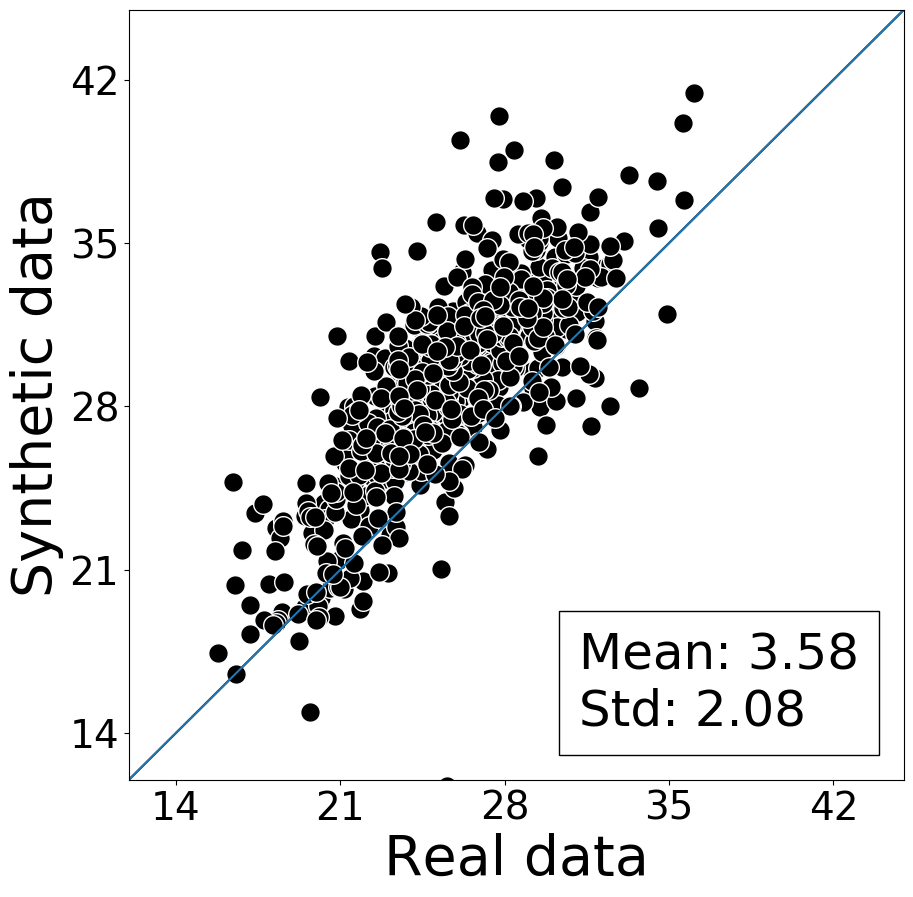}}
\subfigure{%
\label{std_ccd2_7}%
\includegraphics[width=2.cm]{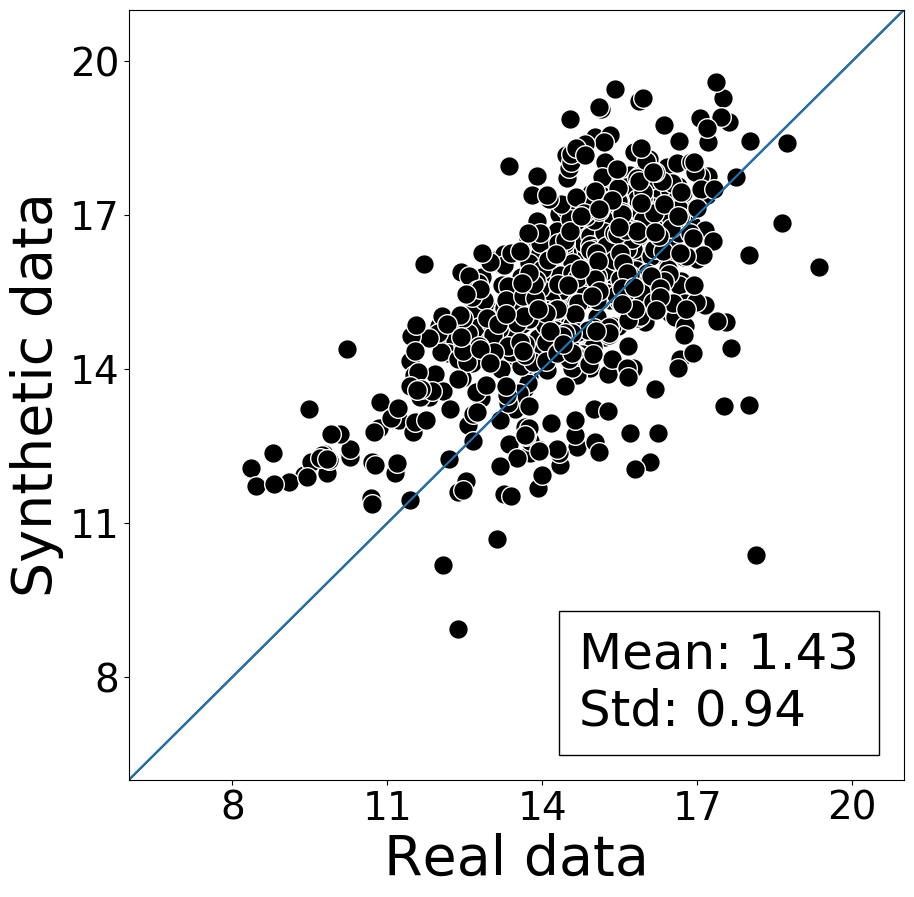}}
\subfigure{%
\label{std_ccd2_9}%
\includegraphics[width=2.cm]{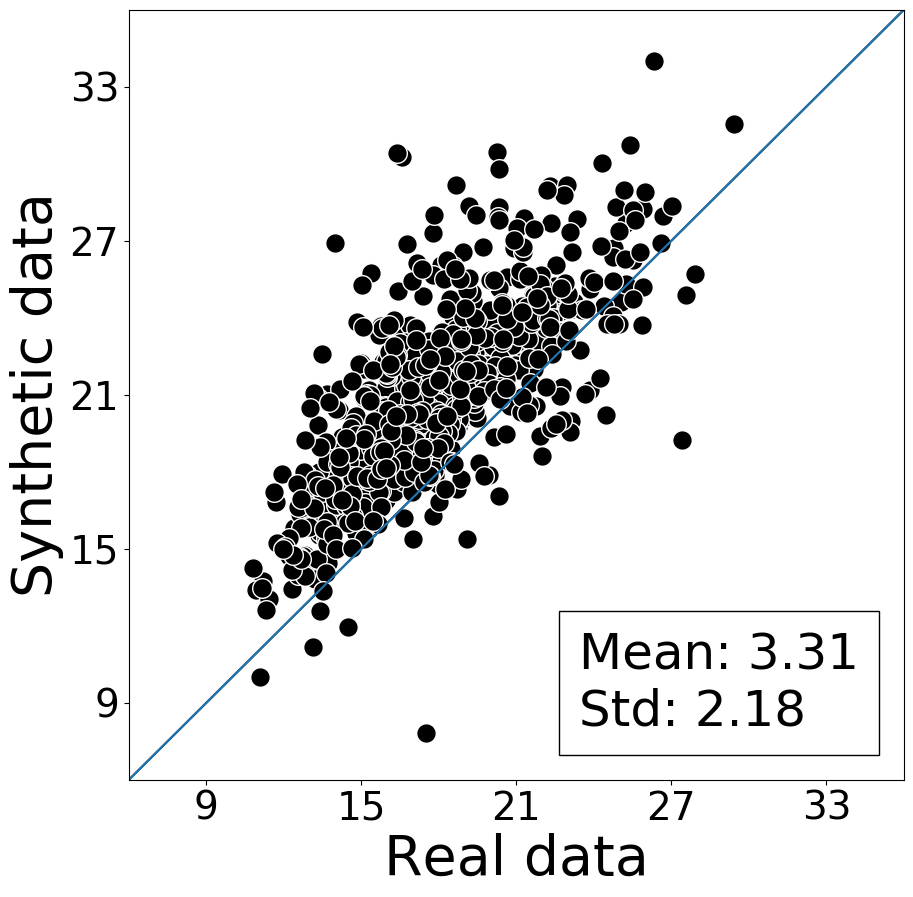}}
\vspace*{-1mm}

\subfigure{%
\includegraphics[width=2.5cm]{./d0.pdf}}
\subfigure{%
\label{std_ccd2_1}%
\includegraphics[width=2.cm]{./min_bmi.pdf}} 
\subfigure{%
\label{std_ccd2_3}%
\includegraphics[width=2.cm]{./max_bmi.pdf}}
\subfigure{%
\label{std_ccd2_4}%
\includegraphics[width=2.cm]{./min_systolic.pdf}} 
\subfigure{%
\label{std_ccd2_6}%
\includegraphics[width=2.cm]{./max_systolic.pdf}}
\subfigure{%
\label{std_ccd2_7}%
\includegraphics[width=2.cm]{./min_diastolic.pdf}}
\subfigure{%
\label{std_ccd2_9}%
\includegraphics[width=2.cm]{./max_diastolic.pdf}}
\vspace*{-20mm}
\caption{Cross-type conditional distribution on vital signs. The coordinates of each dot correspond to the standard deviations of the vital sign distributions on real and synthetic data set.}\label{cond_lab_std}
\end{figure}

\textbf{Cross-type Conditional Distribution (CCD).}
The results of CCD with respect to the demographic distribution are shown in Figure \ref{cond_age_sex}, where each dot denotes a (ICD or CPT) code. The $x$-coordinate corresponds to real data, and the $y$-coordinate of subfigures \ref{ccd_2}-\ref{ccd_3} and \ref{ccd_5}-\ref{ccd_6} correspond to the synthetic data. In each subfigure, the mean and standard deviation of the dot-to-diagonal shortest distance distribution is marked at the bottom right corner. Figures \ref{ccd_1} and \ref{ccd_4} illustrate that the original systems (the baselines of the real \emph{vs} real setting) are stable. When applying HGAN-U, both the age and gender distributions are poorly represented, as shown in Figures \ref{ccd_2} and \ref{ccd_5}. In particular, there are highly skewed gender ratios in multiple ICD and CPT codes. By comparing Figure \ref{ccd_2} with \ref{ccd_3} and Figure \ref{ccd_5} with \ref{ccd_6}, it is evident that HGAN outperforms HGAN-U. This is because it achieves smaller distances from the diagonal. 
For better representing the age distribution for ICD and CPT codes, Figure \ref{cond_age_std} shows the standard deviation of age on each code, which demonstrates the same patterns with results on mean of age. The results of CCD for the vitals are shown in Figures \ref{cond_lab} and \ref{cond_lab_std}, using the same presentation style with Figure \ref{cond_age_sex} and \ref{cond_age_std}, respectively. Both the means and standard deviations demonstrate similar patterns to the demographic results. Thus, our findings for CCD, with respect to three different data types, are consistent.

\emph{\textbf{Privacy Risk Analysis}}

We evaluated the privacy risks of the synthetic data generated by HGAN and EMR-CWGAN. Since EMR-CWGAN was designed to generate billing codes, for comparison we performed two sets of investigations on HGAN--one on ICD and CPT codes only (the same with EMR-CWGAN), the other on all data types. Though age and gender served as the conditions of simulation for both models, we combined these features with the corresponding synthetic instances for analysis. We follow the implementation details of membership and attribute inference in the previous study \cite{zhang2020ensuring}.

\begin{figure}[ht]%
\captionsetup[subfigure]{justification=centering}
\centering
\subfigure[EMR-CWGAN, Codes]{%
\label{mmi_1}%
\includegraphics[width=2.5cm]{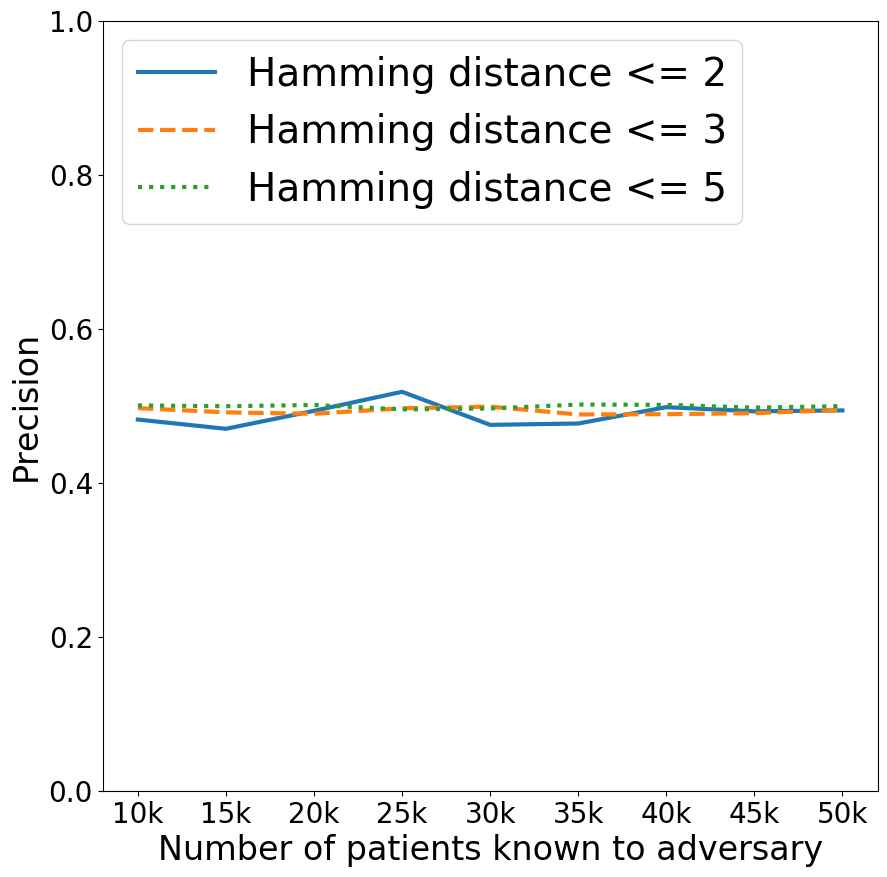}} 
\subfigure[HGAN, Codes]{%
\label{mmi_2}%
\includegraphics[width=2.5cm]{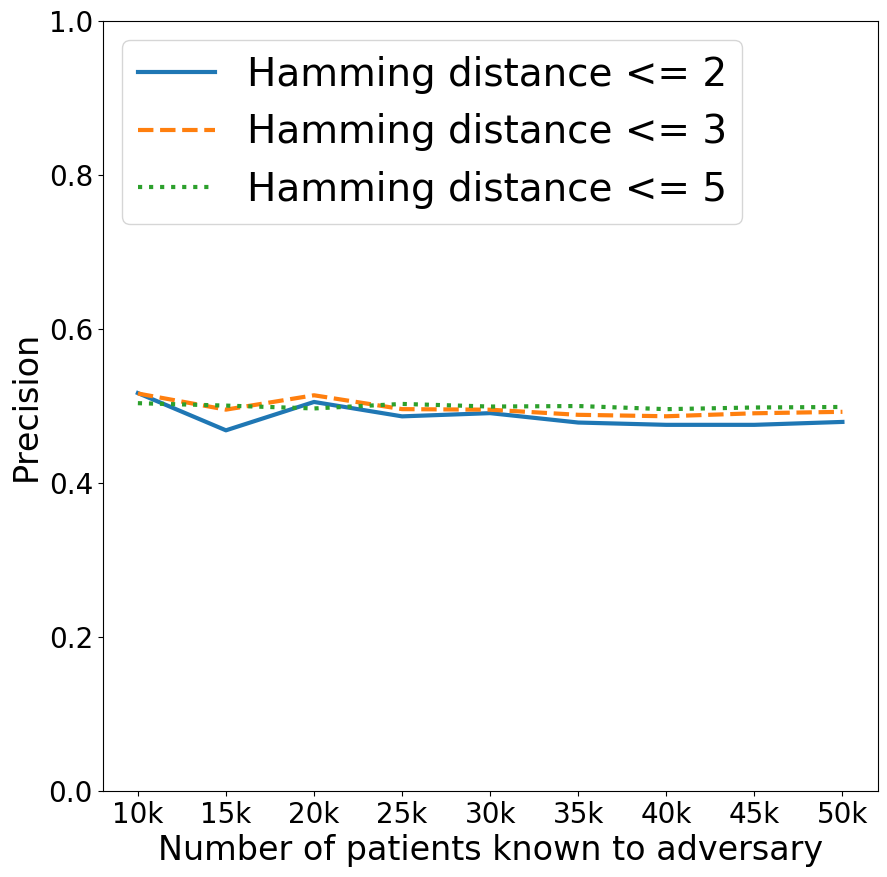}} 
\subfigure[HGAN, All]{%
\label{mmi_3}%
\includegraphics[width=2.5cm]{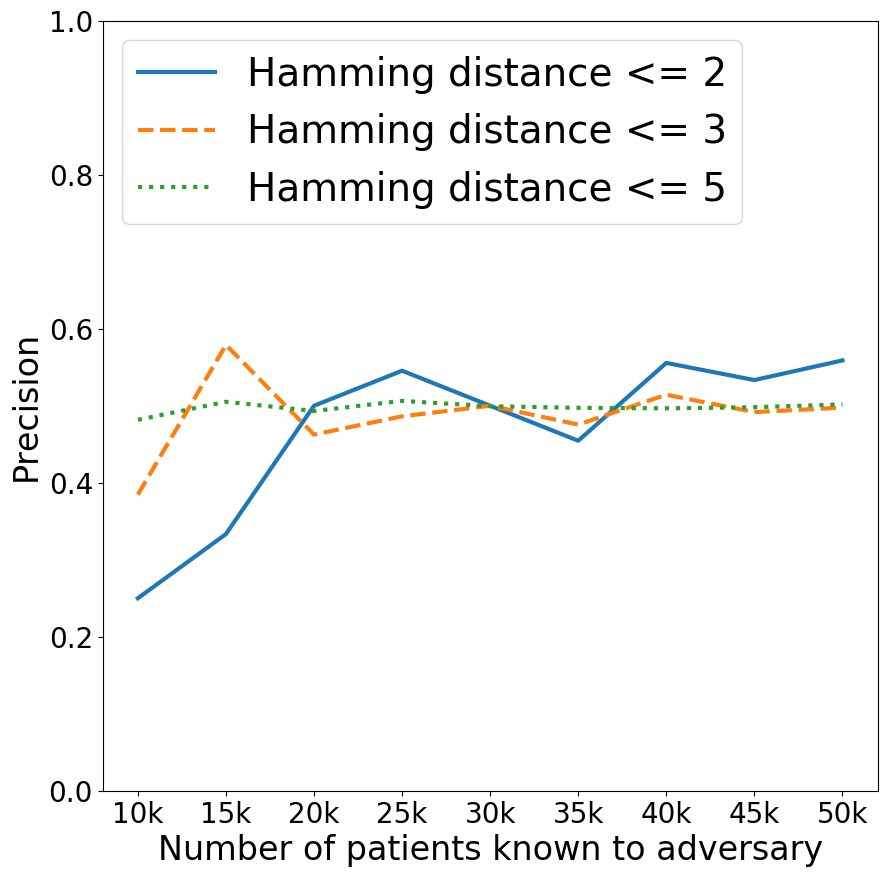}} 
\subfigure[EMR-CWGAN, Codes]{%
\label{mmi_4}%
\includegraphics[width=2.5cm]{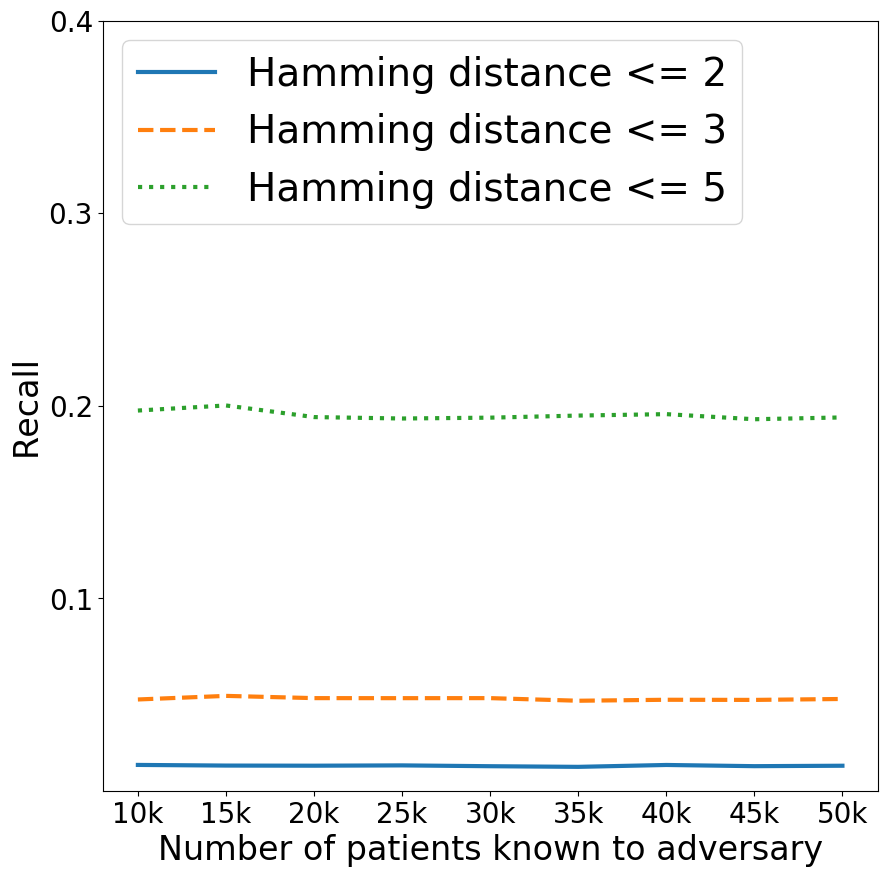}} 
\subfigure[HGAN, Codes]{%
\label{mmi_5}%
\includegraphics[width=2.5cm]{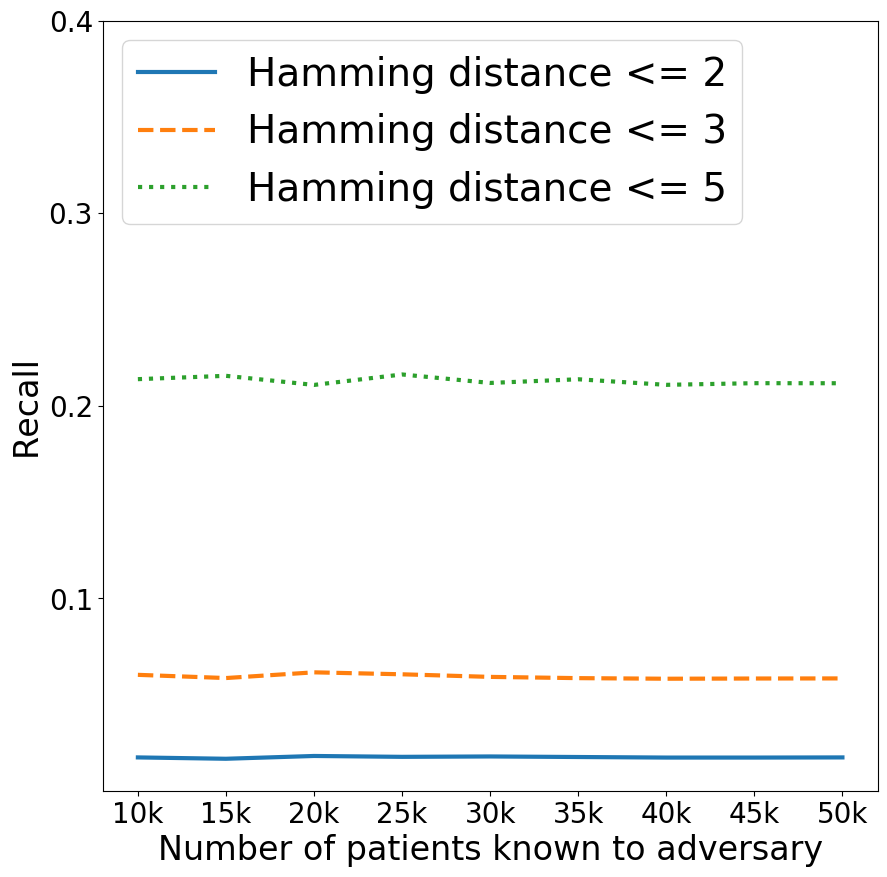}} 
\subfigure[HGAN, All]{%
\label{mmi_6}%
\includegraphics[width=2.5cm]{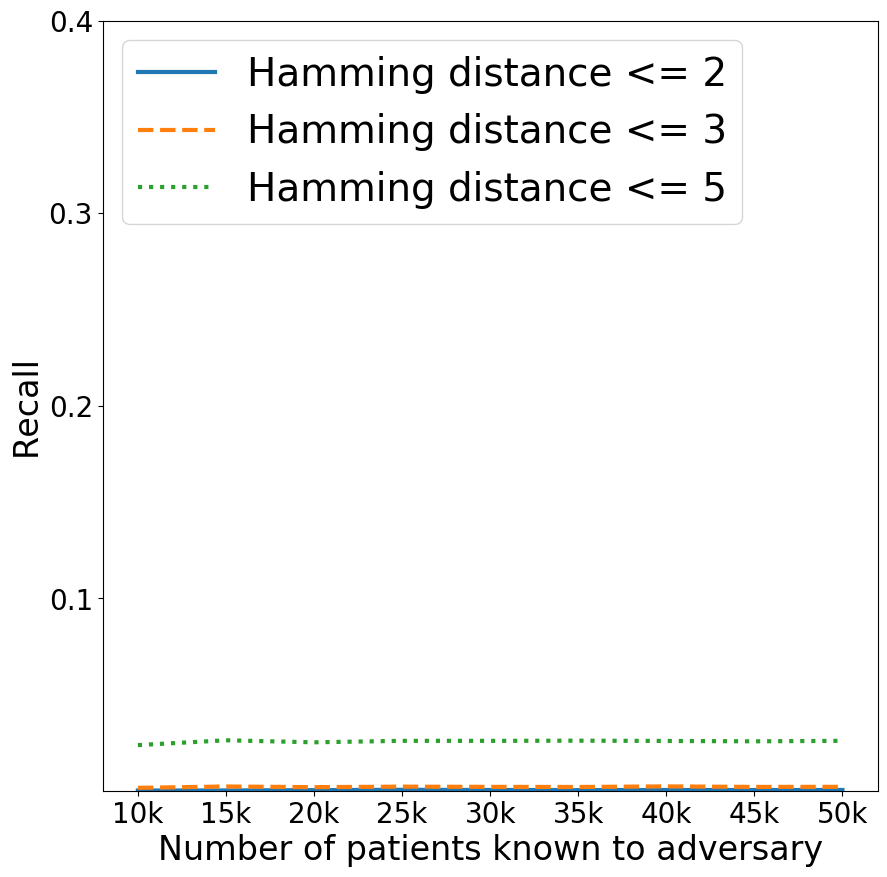}}
\vspace*{-10mm}
\caption{Precision (\ref{mmi_1}--\ref{mmi_3}) and recall (\ref{mmi_4}--\ref{mmi_6}) for membership inference. }\label{mmi}
\end{figure}

\textbf{Membership Inference.}
We evaluated both the precision and recall of success attacks by applying three different thresholds ($2$, $3$ and $5$) for Hamming distance and a range of numbers of known records to an attacker from $5,000$ to $50,000$. Figures \ref{mmi_1}-\ref{mmi_3} show the results of precision for three settings, and figures \ref{mmi_4}-\ref{mmi_6} show the results of recall on the same settings. By comparing Figure \ref{mmi_1} with \ref{mmi_2}, \ref{mmi_4} with \ref{mmi_5}, it can be seen that when investigating the risk on ICD and CPT codes, HGAN demonstrates very similar precision and recall with the state-of-the-art model, EMR-CWGAN, which indicates that in the space of binary features, HGAN suffers no more privacy risk on membership inference to achieve better data utilities as shown above. By comparing Figure \ref{mmi_3} with \ref{mmi_2}, \ref{mmi_6} with \ref{mmi_5}, we observed no obvious increase in precision when adding the continuous features (vital signs), whereas observed a large decrease on the recall of the success of inference. This implies that a larger feature space enables HGAN to generate records that are very similar to real records and, thus, helps lower down the risk of successful membership inference.

\textbf{Attribute Inference.} We randomly sampled $50,000$ real records as the compromised records by an attacher. For each compromised record, we randomly selected $n$ ($n=128, 256$) features under the condition that in each record at least $4\%$ of the selected features are positive as known attributes to the attacker. We found its $k$ ($k=1, 10$) nearest neighbors in the synthetic dataset and performed majority vote to predict the unknown features. The F1 scores of the predictions over HGAN and EMR-CWGAN are shown in Table \ref{attribute}. When considering only the ICD and CPT codes, very tiny difference can be observed between HGAN and EMR-CWGAN, indicating that their privacy risk level of attribute inference are roughly the same. By contrast, when adding the vital sign features, the attribute inference risks decrease about $13\%$. A similar implication with membership inference can be reached that generating more detailed data can decrease the risk of successful attribute inference.

Combining the results of membership and attribute inference, a conclusion can be reached that HGAN suffered no more privacy risks than EMR-CWGAN in billing codes space, but lower down the risks of privacy violation by generating more data types.


\begin{table}[ht]
       \fontsize{7.5}{7}\selectfont
	\centering
	\caption{F1 scores of Attribute inference ($n$: the number of compromised features; $k$: nearest neighbors).}\label{attribute}
	\begin{tabular}{@{}ccccc@{}}
	   \toprule
	 \multirow{2}{*}{\textbf{Model}}  & \multicolumn{4}{c}{\textbf{Known Features, Nearest Neighbors}}   \\[0.3em] \cmidrule(l){2-5} 
		  &   $n=128,k=1$ & $n=128,k=10$ & $n=256,k=1$ & $n=256,k=10$  \\[0.3em] 
		  \hline
		  \rule{0pt}{1.0\normalbaselineskip}
		 \textbf{EMR-CWGAN} & $0.304$ & $0.295$ & $0.328$ & $0.287$   \\[0.6em]
		 \textbf{HGAN with Codes} & $0.314$ & $0.296$ & $0.341$ & $0.294$   \\[0.6em]
		 \textbf{HGAN with All} & $0.266$ & $0.249$ & $0.275$ & $0.249$   \\[0.1em]
		\bottomrule  
	\end{tabular}
	\vspace{-2mm}
\end{table}

%% file: conclusion_final.tex
\section*{Discussions and Conclusions}
\begin{figure}[ht]%
\captionsetup[subfigure]{justification=centering}
\centering
\subfigure[Layer 1]{%
\label{sparsity_1}%
\includegraphics[width=2.5cm]{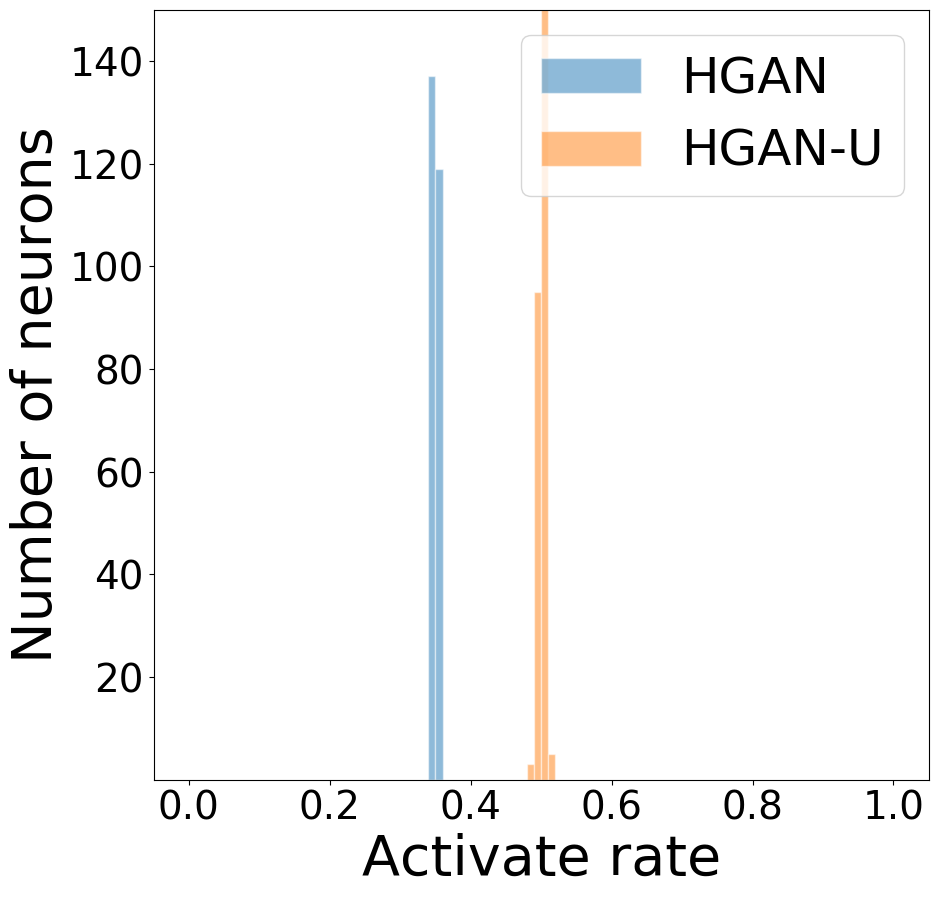}} 
\subfigure[Layer 2]{%
\label{sparsity_2}%
\includegraphics[width=2.5cm]{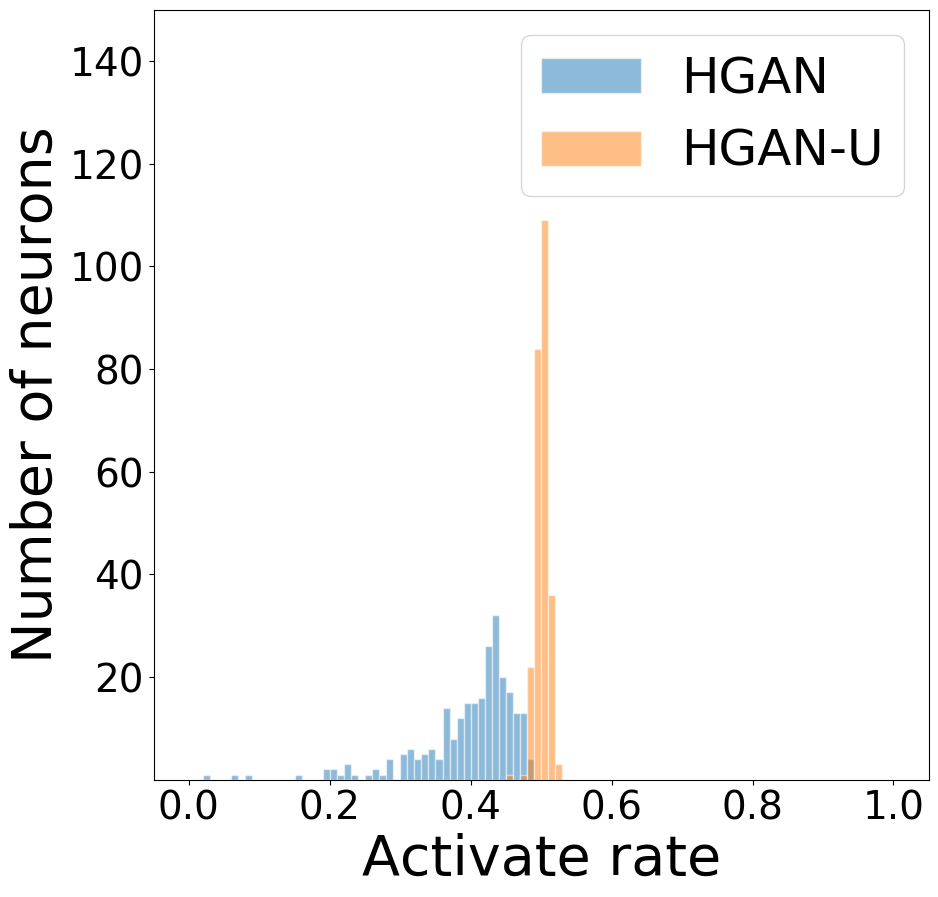}} 
\subfigure[Layer 3]{%
\label{sparsity_3}%
\includegraphics[width=2.5cm]{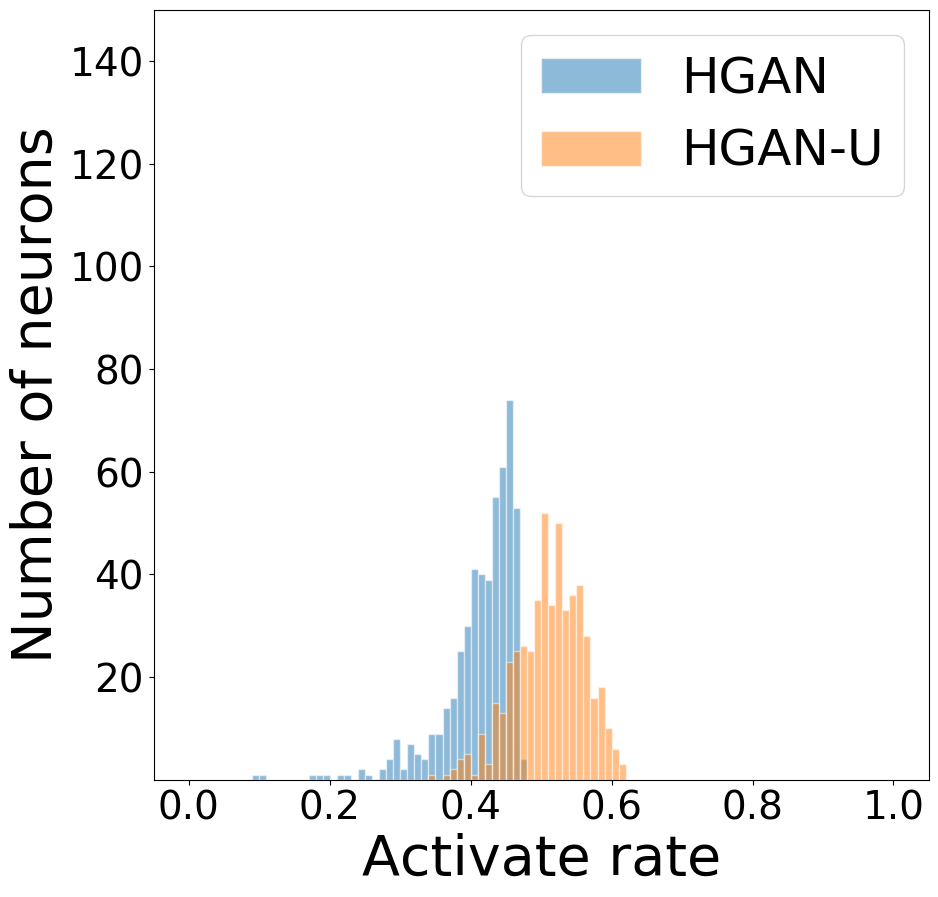}} 
\subfigure[Layer 4]{%
\label{sparsity_4}%
\includegraphics[width=2.5cm]{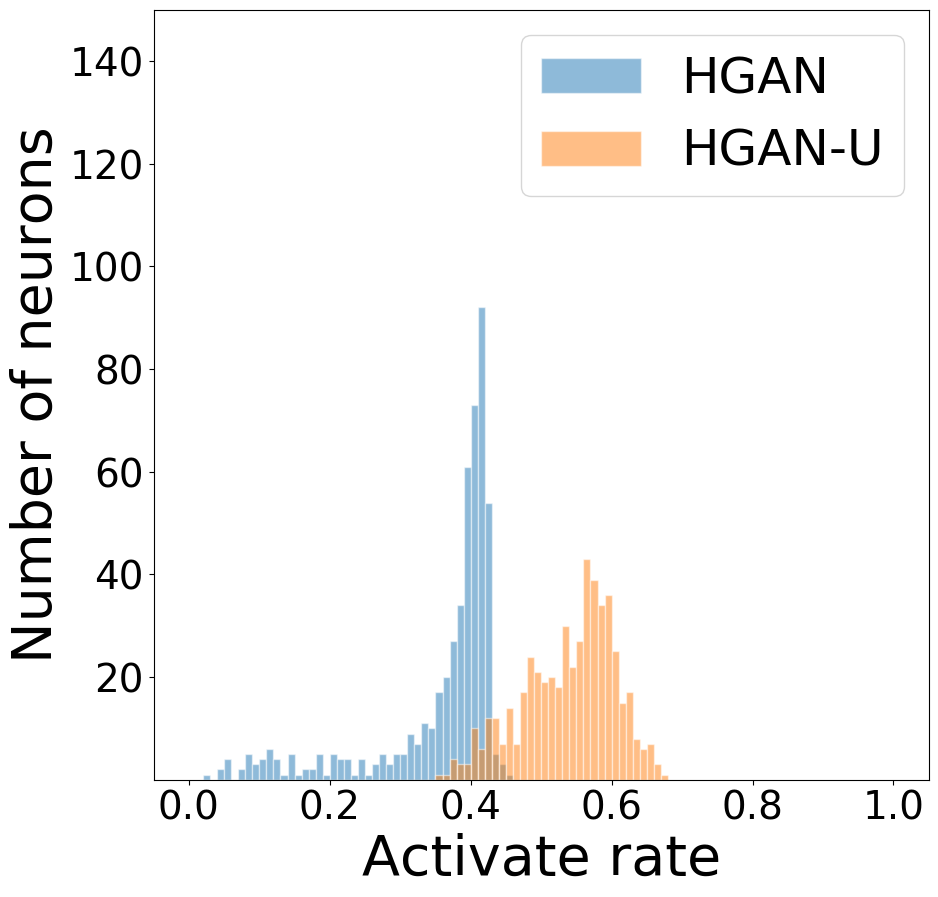}} 
\subfigure[Layer 5]{%
\label{sparsity_5}%
\includegraphics[width=2.5cm]{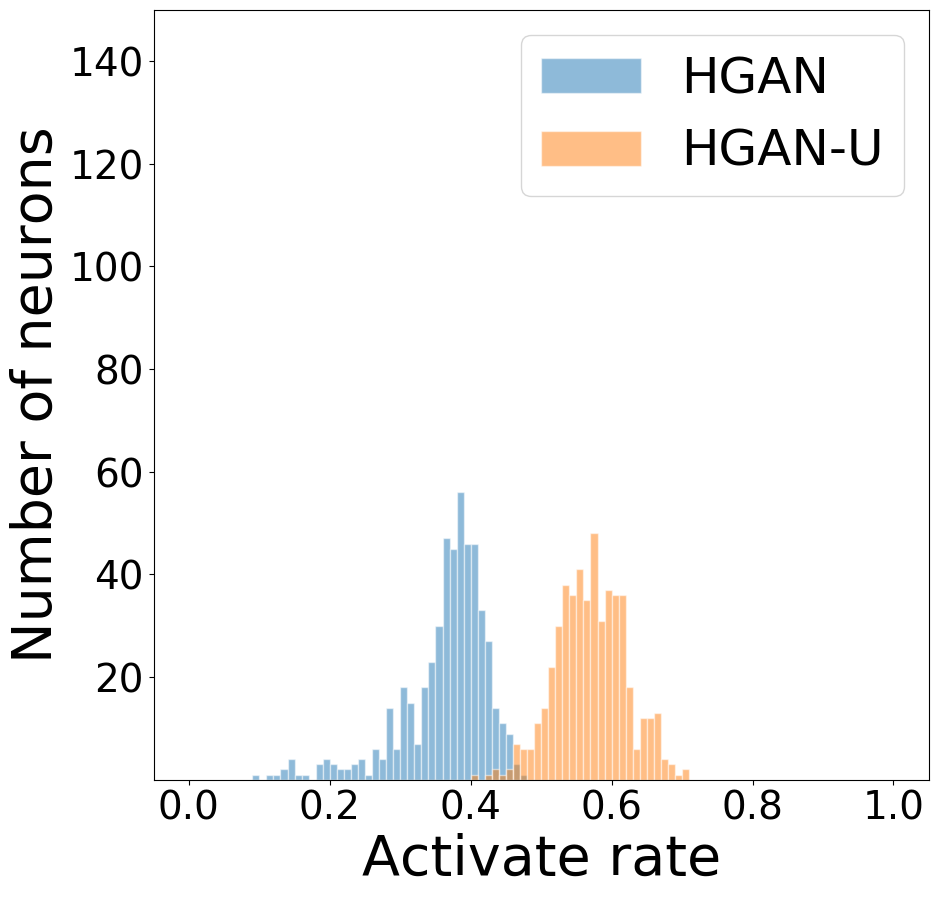}} 
\subfigure[Layer 6]{%
\label{sparsity_6}%
\includegraphics[width=2.5cm]{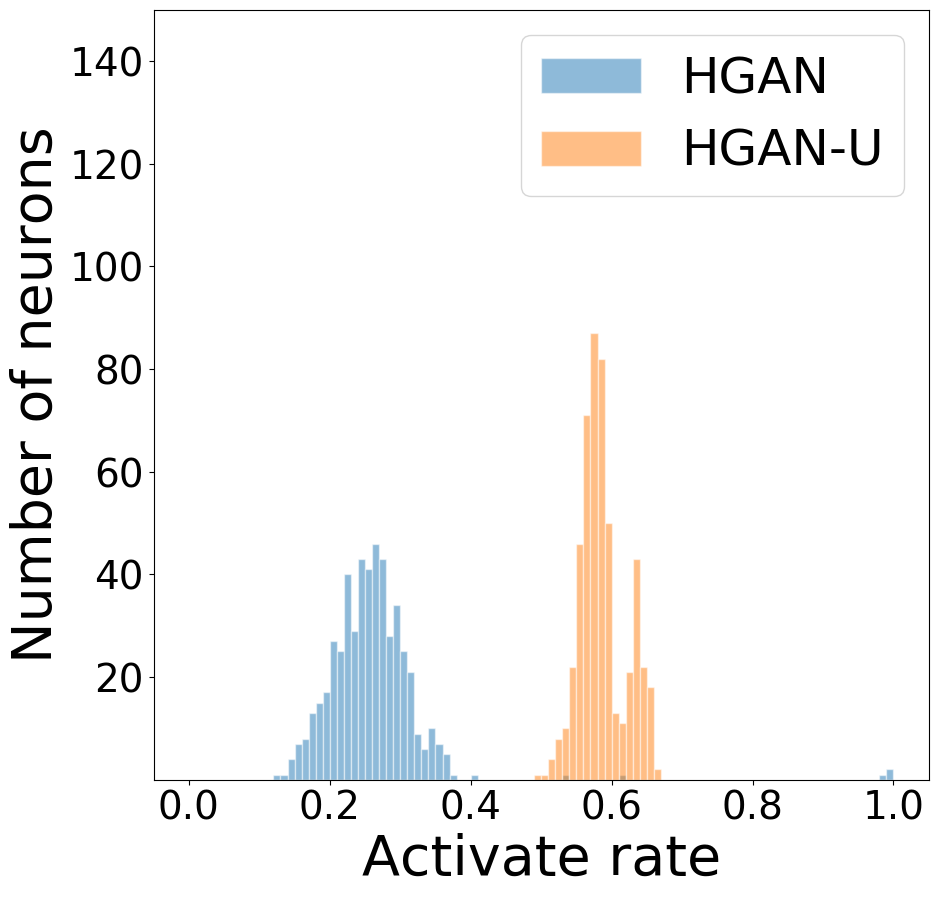}} 
\vspace*{-10mm}
\caption{The histograms of neurons over the activation rate in all layers of the well-trained HGAN and HGAN-U from the CSD dataset.}\label{sparsity}
\end{figure}

This investigation yields several notable implications for EHR data simulation. 
First, the refinement of the neural network design improved the data utility by making the signal sparser. For verifying this, Figure \ref{sparsity} shows the bin plot for the neurons of the well-trained generators (HGAN and HGAN-U) with respect to a range of activation rates among the CSD dataset. 
As can be seen, in all layers HGAN has lower activation rate, which contributes to the signal disentanglement and model robustness. \cite{glorot2011deep}
Second, it appears that a well-designed penalization can effectively prevent feature constraint violations. Importantly, based on the utility measures reported in this work, the incorporation of such penalization appears not bias learning the distribution of real data. Third, this investigation illustrates the importance of the new measures (i.e., CVT, FAR, and CCD) for evaluating the utilities of synthetic data in the simulation tasks with feature constraints and heterogenous data types.

In addition to investigating the adopted design of \emph{ReLU}  $\rightarrow$ \emph{conditional normalization} $\rightarrow$ \emph{ReLU} in HGAN, we also tested a more natural design, which is \emph{ReLU}  $\rightarrow$ \emph{conditional normalization}, as the filter between neighboring fully connected layers in both the generator and discriminator networks. We performed a series of data utility evaluation and privacy risk investigation. The main findings is that our model (the former design) outperformed the model using the latter one.

Despite the merits of this work, there are several limitations we wish to highlight. First, we have not assessed the scalability of our model on larger and sparser feature spaces (e.g., medications and laboratory tests). Second, the model needs to refined to ensure that all meaningful association rules can be retained. Third, we focused on static EHR data simulation, but it is necessary to incorporate temporal factors to simulate more complex phenotypes with trajectories that emerge over time.

\section*{Acknowledgements}
This research was sponsored in part by NIH grants RM1HG009034 and U2COD023196.